%% file: main.tex
\documentclass[10pt,twocolumn,letterpaper]{article}

\usepackage{3dv}
\usepackage{times}

\usepackage{amsmath}
\usepackage{graphicx}
\usepackage[colorinlistoftodos]{todonotes}
\usepackage{overpic}
\usepackage{subfigure}
\usepackage{amsfonts}
\usepackage{amsthm}
\usepackage{amsmath}
\usepackage{algorithm2e}
\usepackage{wrapfig}
\usepackage{pgfplots}
\usepackage{pgf,tikz}
\usepackage{cite}
\usepackage{xcolor}
\usepackage[inline]{enumitem}   
\usepackage{wrapfig}
\usepackage{caption}

\newtheorem{lemma}{Lemma}

\newtheorem{corollary}{Corollary}

\newcommand{\Rr}{\mathbb{R}}
\newcommand{\Xx}{\mathcal{X}}
\newcommand{\Yy}{\mathcal{Y}}

\newcommand{\Pn}{\mathcal{P}_{n}}
\newcommand{\Pnxny}{\mathcal{P}_{n_\Xx}^{n_\Yy}}
\newcommand{\Dn}{\mathcal{B}_{n}}
\newcommand{\PiM}{\mathbf{\Pi}}
\newcommand{\PM}{\mathbf{P}}

\newcommand{\bb}[1]{\mathbf{#1}}
\newcommand{\One}{\mathbf{1}}

\newcommand{\argmax}{\arg\!\!\max}
\newcommand{\argmin}{\arg\!\!\min}

\newcommand{\norm}[1]{\left\|#1\right\|}

\newcommand\blfootnote[1]{%
  \begingroup
  \renewcommand\thefootnote{}\footnote{#1}%
  \addtocounter{footnote}{-1}%
  \endgroup
}

\graphicspath {{figures/}}

\usepackage[pagebackref=true,breaklinks=true,letterpaper=true,colorlinks,bookmarks=false]{hyperref}

\threedvfinalcopy % *** Uncomment this line for the final submission

\def\threedvPaperID{140} % *** Enter the 3DV Paper ID here

\begin{document}

\title{Efficient Deformable Shape Correspondence via Kernel Matching}
\author{Matthias Vestner$^\star$\\
TU Munich
\and
Zorah L\"ahner$^\star$\\
TU Munich
\and
Amit Boyarski$^\star$\\
Technion
\and
Or Litany\\
TAU
\and
Ron Slossberg\\
Technion
\and
Tal Remez\\
TAU
\and
Emanuele Rodol\`a\\
Sapienza University of Rome / USI Lugano
\and
Alex Bronstein\\
Technion / TAU / Intel
\and
Michael Bronstein\\
USI Lugano / TAU / Intel
\and
Ron Kimmel\\
Technion / Intel
\and
Daniel Cremers\\
TU Munich
}

%\maketitle

\twocolumn[{%
\renewcommand\twocolumn[1][]{#1}%
\maketitle
\begin{center}{
\vspace{-5mm}
    \includegraphics[width=.14\linewidth]{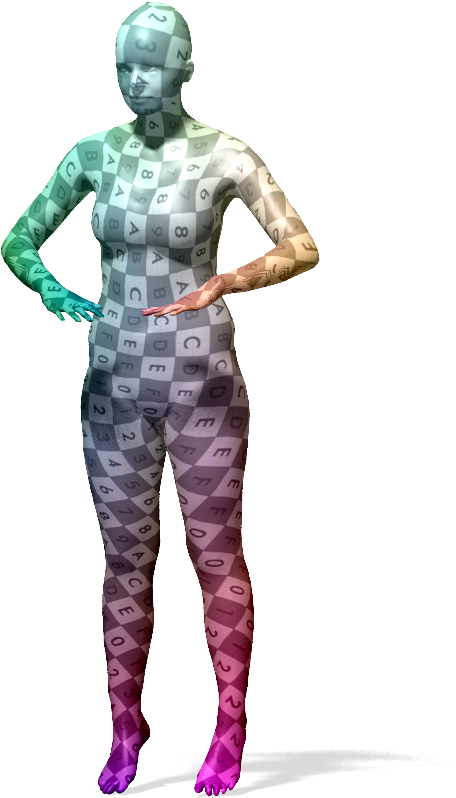}
    \includegraphics[width=.17\linewidth]{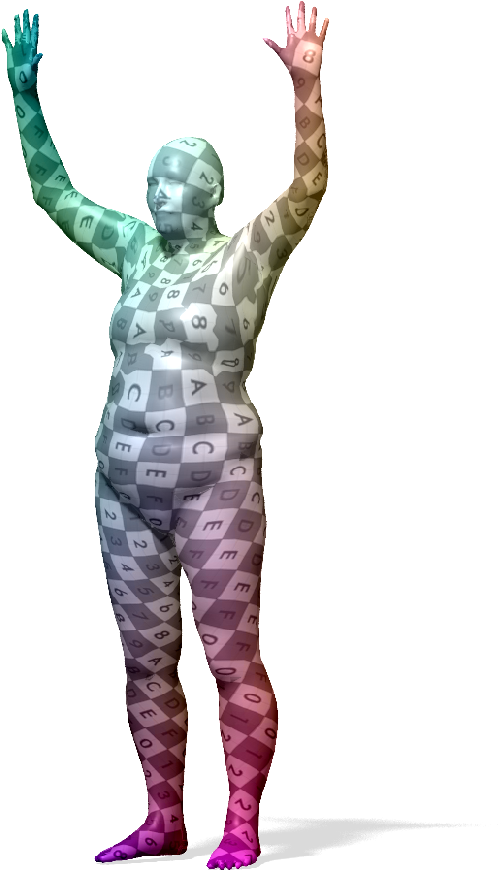}
	\includegraphics[width=.14\linewidth]{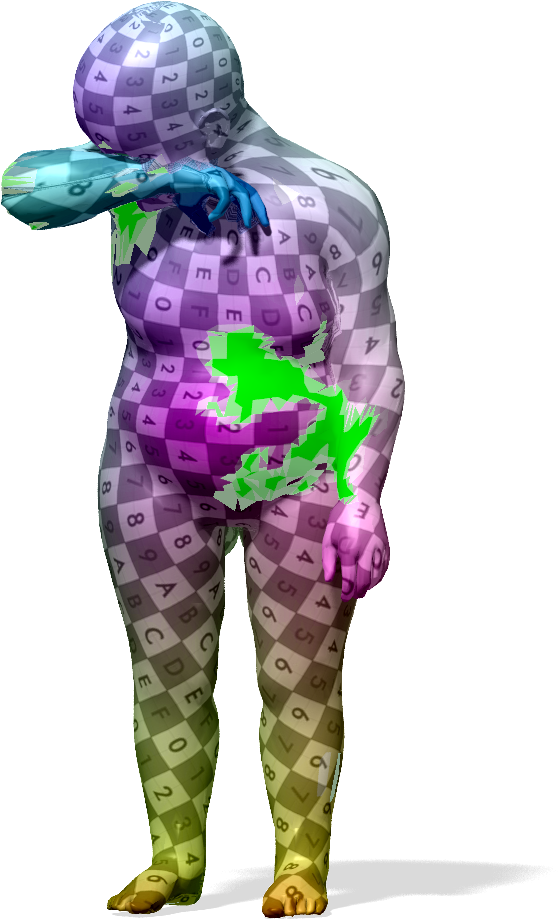}
    \includegraphics[width=.14\linewidth]{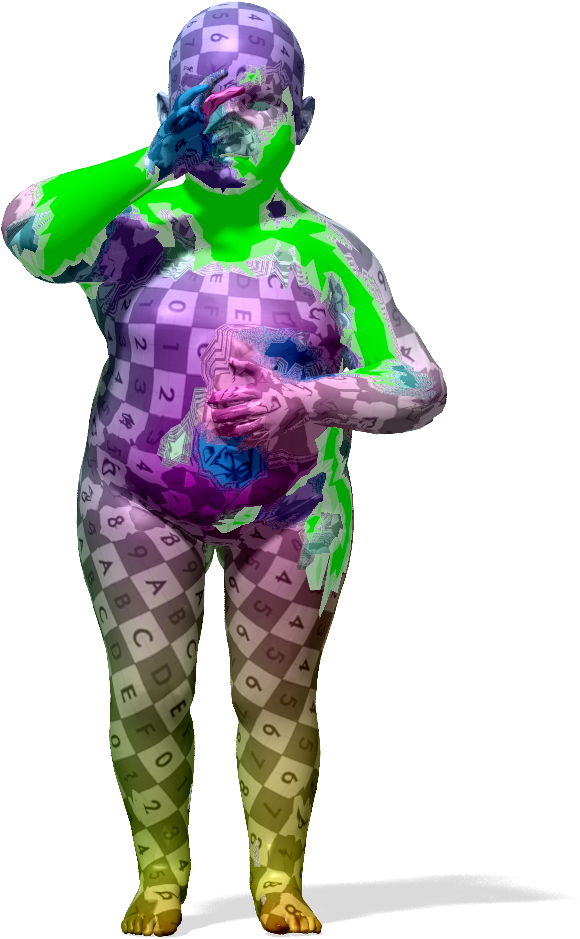}
    \includegraphics[width=.18\linewidth]{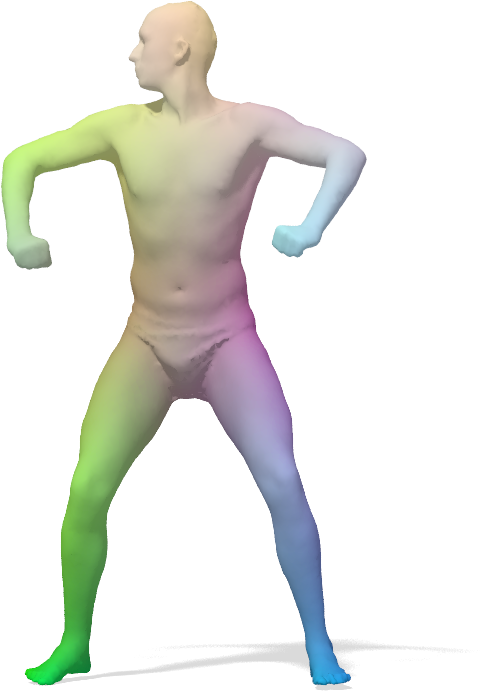}
    \includegraphics[width=.18\linewidth]{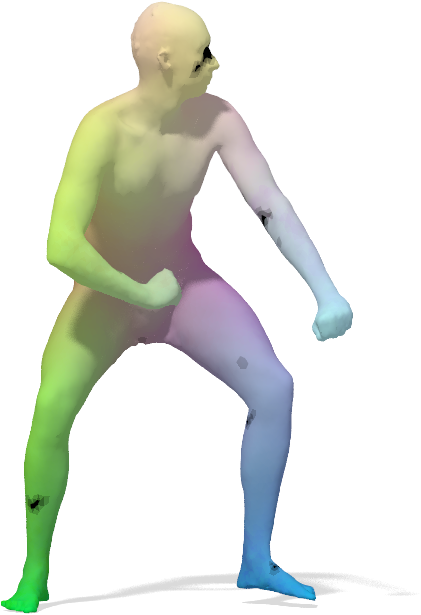}
    \captionof{figure}{Qualitative examples on FAUST models (left), SHREC'16 (middle) and SCAPE (right). In the SHREC experiment, the green parts mark where no correspondence was found. Notice how those areas are close to the parts that are hidden in the other model. The missing matches (marked in black) in the SCAPE experiment are an artifact due to the multiscale approach.\label{fig:renderings}}    
}
\end{center}%
}]

\begin{abstract}
We present a method to match three dimensional shapes under non-isometric deformations, topology changes and partiality. We formulate the problem as matching between a set of pair-wise and point-wise descriptors, imposing a continuity prior on the mapping, and propose a projected descent optimization procedure inspired by difference of convex functions (DC) programming.
\end{abstract}
%--------------------------------------------------------------------------------------------------------------------

\blfootnote{$\star$ equal contribution}

\input{01_introduction}

\input{02_background}
\input{03_method}
\input{04_interpretation}
\input{05_experiments}

\input{06_conclusion}

\clearpage % force a pagebreak and flush all deferred `table` and `figure` environments
{\small
\bibliographystyle{ieee}
\bibliography{references}
}
\clearpage
\appendix
\input{07_supplMaterial}

\end{document}

%% file: 01_introduction.tex
\vspace{-1cm}
\section{Introduction} \label{sec:introduction}

\noindent Finding correspondences between non-rigid shapes is a fundamental problem in  computer vision, graphics and pattern recognition, with applications including shape comparison, texture transfer, and shape interpolation just to name a few. Given two three-dimensional objects $\Xx$ and $\Yy$, modeled as compact two-dimensional Riemannian manifolds, our task is to find a meaningful correspondence $\varphi:\Xx \rightarrow \Yy$. While a rigorous definition of \emph{meaningful} is challenging, one can identify some desirable properties of $\varphi$:
%
% In this paper, we propose a family of pairwise descriptors and utilize them to obtain a continuity-promoting regularizer for the linear assignment objective arising from pointwise descriptor matching. This regularizing term leads to a convex quadratic assignment problem. We propose to optimize this objective by iteratively solving a series of linear assignment problems. In addition, we introduce a multiscale scheme to alleviate the computational burden arising when shapes are densely sampled. We show that our method produces a smooth matching for non-isometric shapes and outperforms previous methods on the challenging task of matching shapes with different topology.
%
%
\begin{enumerate}[topsep=0pt,itemsep=-1ex,partopsep=1ex,parsep=1ex]
 \item Bijective.
\item Continuous in both directions, in the sense that nearby points on $\Xx$ should be mapped to nearby points on $\Yy$ (and vice versa).
\item Similar points should be put into correspondence.
\end{enumerate}

\noindent For the simplicity of the introduction, we assume the two shapes $\Xx$ and $\Yy$ to be sampled at $n$ points each, and defer the case of different number of samples to the algorithmic part of this paper detailed in Section \ref{sec:algo}. Assuming a consistent sampling (e.g. via farthest point sampling with a sufficiently large number $n$ of points), the discrete counterpart to the correspondence $\varphi$ is a mapping $\pi:\{x_1,\ldots x_n\} \rightarrow \{y_1,\ldots y_n\}$, which admits a representation as a permutation matrix $\mathbf{\Pi} \in \{0,1\}^{n \times n}$ satisfying $\mathbf \Pi^\top\One = \mathbf \Pi \One =\One$ with $\One$ being a column vector of ones. 
We henceforth denote the space of $n \times n$ permutation matrices by $\Pn$.

\noindent The vast majority of shape matching approaches phrase the correspondence problem as an energy minimization problem 
\begin{align}\label{energyMinimization}
 {\PiM}^* &=\argmin_{\PiM\in \Pn} E({\PiM})\,,
\end{align}
where \noindent $E(\PiM)$ is usually a weighed aggregate of two terms
\begin{align}
E(\PiM) = \alpha g(\PiM)+ h(\PiM)\,.
\end{align}
The first term $g(\PiM)$ is a fidelity term trying to align a set of \textit{pointwise descriptors} encoding the similarity between points, while the second term $h(\PiM)$ is a regularization term promoting the continuity of the correspondence by aligning a set of \textit{pairwise descriptors} encoding global/local relations between pairs of points. The parameter $\alpha$ governs the tradeoff between the two terms.

\noindent  While the constraint $\PiM \in \Pn$ guarantees bijectivity of the correspondence, the two terms $h$ and $g$ correspond, respectively, to the second and the third desirable qualities of a meaningful correspondence, and provide a trade-off between complexity, fidelity and regularity. We stress in our work that despite their seemingly unrelated nature, those properties are in fact tightly connected, i.e., choosing a particular set of pairwise descriptors might have a profound effect not only on the regularity of the final solution, but also on the complexity of the resulting optimization problem. We will elaborate on these aspects in more detail.% in Section \ref{sec:background}.

 %---------------------------------------------------------

\vspace{1ex}\noindent\textbf{Related work.} 
Finding correspondences between shapes is a well-studied problem. Traditionally, the solution involves minimization of a distortion criterion which fits into one of the two categories: pointwise descriptor similarity \cite{rustamov2007laplace,sun2009concise,bronstein2010scale,aubry2011wave,shot}, and pairwise relations \cite{memoli2005theoretical,chen2015robust, coifman2005geometric,torresani2008feature}. In the former case, matches are obtained via nearest neighbor search or, when injectivity is required, by solving a linear assignment problem (LAP). Pairwise methods usually come at a high computational cost, with the most classical formulation taking the form of an NP-hard {\em quadratic assignment problem} (QAP)~\cite{pardalos1994quadratic}. Several heuristics have been proposed to address this issue by using subsampling \cite{tevs2011intrinsic} or coarse-to-fine techniques \cite{wang2011discrete,sahillioglu2011coarse}. Various relaxations have been used to make the QAP problem tractable \cite{bronstein2006generalized,leordeanu2005spectral,rodola2012game,aflalo2015convex,chen2015robust,kezurer2015tight}, however they result in approximate solutions. In addition, pairwise geodesics are computationally expensive, and sensitive to noise. In \cite{hu2013spectral} the use of heat kernels was proposed as a noise-tolerant approximation of matching adjacency matrices. In \cite{PMF} dense bijective correspondences were derived from sparse and possibly noisy input using an iterative filtering scheme, making use of geodesic Gaussian kernels. %Their method could be interpreted as matching between geodesic Gaussian kernels.

\noindent A different family of methods look for pointwise matches in a lower-dimensional ``canonical'' embedding space. %, in which the richness of the deformations of the original shape is replaced by a simpler group of transformations.
Such embedding can be carried out by multidimensional scaling \cite{elad2003bending,bronstein2006efficient} or via the eigenfunctions of the Laplace-Beltrami operator (LBO) \cite{mateus2008articulated,shtern2013matching}. The correspondence is then calculated in the embedding space using a simple rigid alignment technique such as ICP \cite{besl1992method}. Functional maps \cite{ovsjanikov2012functional,kovnatsky2015functional} can be seen as a sophisticated way to initialize ICP when using this spectral embedding. Other bases can be used within the functional map framework \cite{quasiHarmonic}. In  particular, the eigenspaces arising from the spectral decomposition of the geodesic distance matrices have been shown to outperform the LBO basis for the case of isometric shapes \cite{ShamaiK16}. 
In \cite{windheuser2011geometrically} the matching problem is phrased as an integer linear program, enforcing continuity of the correspondence via a linear constraint. This additional constraint however makes the problem computationally intractable even for modestly-sized shapes, requiring the use of relaxation and post-processing heuristics.

\noindent Most recent works attempt to formulate the correspondence problem as a learning problem \cite{rodola2014dense} and design intrinsic deep learning architectures on manifolds and point clouds  \cite{masci2015geodesic,boscaini2015learning,boscaini2016anisotropic,boscaini2016learning,monti2016geometric,litany2017fmnet}. As of today, these methods hold the record of performance on deformable correspondence benchmarks; however, supervised learning requires a significant annotated training set that is often hard to obtain.% In addition, some of the reported state-of-the-art results are obtained by an additional post process

%\MV{Discuss PMF.}
%Additional previous works are discussed in Section 2.4 as part of the discussion of pairwise descriptors.
%--------------------------------------------------------------------------------------------------------------------

\vspace{1ex}\noindent\textbf{Contribution.}
%--------------------------------------------------------------------------------------------------------------------
The main contribution of this paper is a simple method that works out-of-the-box for finding high quality continuous (regular) correspondence between two not necessarily isometric shapes. The method can be seen as an improved version of \cite{PMF}, and is accompanied by theoretical insights that shed light on its effectiveness. In particular, we contrast the method with other shape matching approaches and elaborate on the computational benefits of using {\em kernels} rather than distances as pairwise descriptors. The key insight is the realization that high quality regular correspondence can be obtained from a rough irregular one by a sequence of smoothing and projection operations. Remarkably, this process admits an appealing interpretation as an alternating diffusion process~\cite{lederman2015learning}.
We report drastic runtime and scalability improvements compared to~\cite{PMF}, and present an extension to the setting of {\em partial} shape correspondence and an effective multi-scale approach.

%that we use to design an effective multi-scale approach.

%% file: 02_background.tex
\section{Background}\label{sec:background}

\subsection{Pointwise descriptors}
\noindent Similarity of points is often measured with the help of {\em pointwise descriptors} $f_\Xx:\Xx\rightarrow\Rr^q$, $f_\Yy:\Yy\rightarrow\Rr^q$ that are constructed in a way such that similar points on the two shapes are assigned closeby (in the Euclidean sense) descriptors, while dissimilar points are assigned distant descriptors. In the discrete case, the descriptors $f_\Xx,f_\Yy$ can be encoded as matrices $\mathbf F_\Xx,\mathbf F_\Yy\in \Rr^{n\times q}$ giving rise to the optimization problem\footnote{Throughout this paper we use the Frobenius norm $\|\mathbf{A}\|=\sqrt{\langle\mathbf{A},\mathbf{A}\rangle}$, where $\langle\mathbf{A},\mathbf{B}\rangle=\mathrm{tr}(\mathbf{A}^\top\mathbf{B})$ is the Euclidean inner product.}
\begin{align}\label{eq:LAP}
 \argmin_{\mathbf \Pi\in \Pn} \norm{ {\mathbf \Pi} \mathbf F_\Xx-\mathbf F_\Yy}^2 = \argmax_{\mathbf \Pi\in \Pn} \langle  {\mathbf \Pi},\mathbf F_\Yy \mathbf F_\Xx^\top \rangle \,.
\end{align}
Problem \eqref{eq:LAP} is linear in ${\mathbf \Pi}$ and is therefore one of the rare examples of combinatorial optimization problems that can be globally optimized in polynomial time; the best known complexity $O(n^2\log n)$ is achieved by the auction algorithm \cite{auctionAlgo}. 

\noindent Over the last years, \emph{intrinsic} features have extensively been used due to their invariance to isometry. However, they come with two main drawbacks: First, the implicit assumption that the shapes at hand are isometric is not always met in practice. Today's best performing approaches partially tackle this problem using deep learning  \cite{masci2015geodesic,boscaini2015learning,boscaini2016anisotropic,boscaini2016learning,monti2016geometric}. Secondly, many natural shapes come with at least one intrinsic (e.g., bilateral) symmetry that is impossible to capture by purely intrinsic features, be these handcrafted or learned. Correspondences obtained by \eqref{eq:LAP} may suffer from severe discontinuities due to some points being mapped to the desired destination, and others to the symmetric counterpart.
%
% Windheuser et al. \cite{windheuser2011geometrically} introduced an additional linear constraint on ${\mathbf \Pi}$ to enforce continuity of the correspondence. Unfortunately, this additional constraint makes the optimization problem intractable and heuristic binarization techniques have to be applied after relaxing the integer constraints $\pi_{ij} \in \{0,1\}$. More recently, Litany et al. \cite{litany2017fmnet} used a deep learning architecture with a functional map layer to enforce smooth correspondence.

%-------------------------------------------------------------------------------------------

%\vspace{1ex}\noindent\textbf{Pairwise descriptors.}

\subsection{Pairwise descriptors}
\noindent Another family of methods consider {\em pairwise descriptors} of the form $d_\Xx: \Xx\times \Xx \rightarrow \Rr$, $d_\Yy: \Yy\times \Yy \rightarrow \Rr$ encoded in the discrete setting as symmetric matrices $\mathbf D_\Xx,\mathbf D_\Yy \in \Rr^{n\times n}$. These methods aim at solving optimization problems of the form
\begin{align}
 {\mathbf \Pi}^* 
 %&= \argmin_{\mathbf \Pi\in \Pn} \norm{{\mathbf \Pi}\mathbf  D_\Xx {\mathbf \Pi}^\top-\mathbf D_\Yy}^2 \label{qap1}\\
 &= \argmin_{\mathbf \Pi\in \Pn} \norm{{\mathbf \Pi} \mathbf D_\Xx - \mathbf D_\Yy {\mathbf \Pi}}^2\label{qap2}\\
 &= \argmax_{\mathbf \Pi\in \Pn} \langle {\mathbf \Pi},\mathbf D_\Yy {\mathbf \Pi}\mathbf  D_\Xx \rangle \label{qap3}\,,
 %\\ &= \argmin_{\mathbf \Pi} - \trace(vec(\mathbf \Pi)^T \mathbf D_\Xx \otimes\mathbf  D_\Yy vec(\mathbf \Pi)) \label{qap4}
\end{align}
known under the names of {\em graph matching} (GM) or {\em quadratic assignment problem} (QAP), and are in general not solvable in polynomial time.
A typical way to circumvent the complexity issue is to relax the integer constraint $\pi_{ij} \in \{0,1\}$ and optimize the objectives \eqref{qap2}-\eqref{qap3} over the convex set of {\em bi-stochastic matrices} $\Dn = \{ {\mathbf P} \ge \bb{0} : {\mathbf P}^\top \One = {\mathbf P} \One = \One  \}$.
Note that when viewed as functions over this convex set, the objectives \eqref{qap2}-\eqref{qap3} are no longer equivalent. In particular, \eqref{qap2} will always be convex, while the convexity of \eqref{qap3} depends on the eigenvalues of the matrices $\mathbf D_\Xx$ and $\mathbf D_\Yy$, as shown in the following lemma.

\begin{lemma}
 Let $\mathbf D_\Xx,\mathbf D_\Yy$ be symmetric. The function
 %\\ $h:\Dn\rightarrow \Rr$ defined as
$h(\mathbf P) = \langle {\mathbf P},\mathbf D_\Yy {\mathbf P}\mathbf  D_\Xx \rangle$
over the set of bi-stochastic matrices $\Dn$ is (strictly) convex iff all eigenvalues of $\mathbf D_\Xx$ and $\mathbf D_\Yy$ are (strictly) positive.
\end{lemma}

\begin{corollary}\label{cor:cor1}
If all eigenvalues of $\mathbf D_\Xx$ and $\mathbf D_\Yy$ are strictly positive, the optimum of the relaxed problem coincides with that of the original  combinatorial problem:
\begin{align}
 \argmax_{\mathbf P\in \Dn} h(\mathbf P) &= \argmax_{\mathbf \Pi\in \Pn} h(\mathbf \Pi)\,.
\end{align}

\end{corollary} 

\noindent Notice that we can add a linear term (weighted by a scalar factor $\alpha$), such as the one in \eqref{eq:LAP}, while still keeping this property:
$E(\mathbf P) = \alpha \langle  {\mathbf P},\mathbf F_\Yy \mathbf F_\Xx^\top \rangle + \langle {\mathbf P},\mathbf D_\Yy {\mathbf P}\mathbf  D_\Xx \rangle\,$.

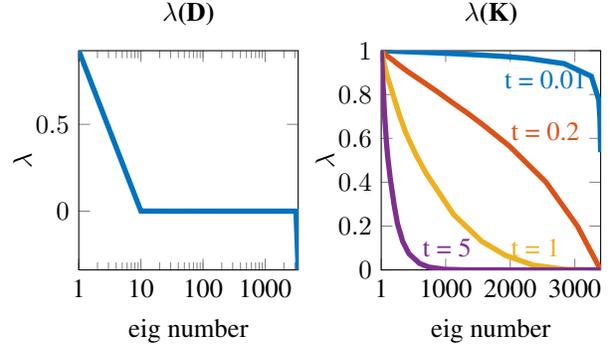
\begin{figure}[t]
\input{figures/eigs1.tikz}
\input{figures/eigs2.tikz}
 \caption{Spectrum of distance matrix (left) vs. spectrum of heat-kernel matrix (right) for several values of $t\in[0.01,5]$ computed on the cat shape from TOSCA.}
 \label{fig:spectrum}
\end{figure}

\noindent \noindent Popular pairwise descriptors include a variety of pairwise distances  \cite{bronstein2006generalized,bronstein2010gromov,raviv2011affine} and kernels \cite{liu2008kernel,shtern2014iterative,PMF} tailored for the specific class of deformations. In what follows, we advocate the superiority of using kernels over distances.
 
% \paragraph{Choice of pairwise descriptors}\label{subsec:choicePairwise}

\vspace{1ex}\noindent\textbf{Pairwise distances.}
A common choice for pairwise descriptors are geodesic distances $d_\Xx(x_i,x_j)$, a choice motivated by the fact that, for isometric shapes, these are preserved by the optimal ${\mathbf \Pi}$. Geodesic distances have major drawbacks, both from the modeling and computational point of view.
%
%which give rise to a family of methods looking for minimum isometric distortion correspondence \cite{}. In particular for isometric shapes, the optimal ${\mathbf \Pi}$ will preserve all distances. While accompanied with some nice theory from metric geometry \cite{bronstein2008numerical}, and providing some versatility as to the types of transformations they describe via the choice of the metric, using pairwise distances as descriptors comes with drawbacks, both from the modeling and computational point of view.
On the modeling side, they introduce a bias towards far away points and are sensitive to topological noise.
On the computational side, they are slow to compute and give rise to highly non-convex (and non-differentiable) optimization problems.
Note that, although one may employ more robust definitions of distance \cite{bronstein2010gromov,chen2015robust}, these do not solve the optimization issues.
%To cope with these problems, different choices of distances can be used, like diffusion distances \cite{bronstein2010gromov}, which are less sensitive to topological noise, and different robust norms which are less sensitive to outliers \cite{chen2015robust}. While diffusion distances relax the computational burden of computing geodesic distances, the issue of optimization still exists. This issue, as mentioned above, is usually coped with by means of relaxation. 

\vspace{1ex}\noindent\textbf{Heat kernels. }
Heat kernels are fundamental solutions to the {\em heat diffusion equation} on manifold $\mathcal{X}$,
\begin{align}
 \frac{\partial u(t,x)}{\partial t} &= \Delta_{\mathcal{X}}  u(t,x)\,,
\end{align}
with the initial condition $u(0,x) = u_0(x)$ and additional boundary conditions if applicable. Here $u: [0,\infty) \times \mathcal{X} \rightarrow \mathbb{R}$ represents the amount of heat at point $x$ at time $t$. % and $u_0$ is the initial condition. 
The solution is linear in the initial distribution and is given by
\begin{align}
 u(t,x) &= \int_\Xx k(t,x,x') u_0(x') dx'\,,
\end{align}
where $k:\Rr^+\times \mathcal{X}\times \mathcal{X} \rightarrow \mathbb{R}$ is the \emph{heat kernel} and its values can be interpreted as the amount of heat transported from $x'$ to $x$ in time $t$.
In the Euclidean case, the heat kernel is an isotropic Gaussian kernel with the variance proportional to the diffusion time $t$. 

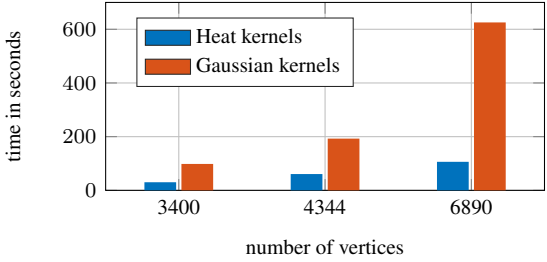
\begin{figure}[t]
\input{figures/run_times.tikz}
 \caption{Runtime comparison of matching shapes with varying number of vertices using our algorithm with heat kernels compared to Gaussian kernels\cite{PMF}. For more info see supp. material.}
 \label{fig:runtimes}
\end{figure}

\noindent For a compact manifold $\Xx$, the heat kernel can be expressed as the exponent 
of the intrinsic self-adjoint negative semi-definite Laplacian operator $\Delta_{\mathcal{X}}$,
\begin{align}
 k(t,x,x') &= \sum_i e^{\lambda_i t} \phi_i(x)\phi_i(x'),
\end{align}
where 
%\begin{align}
 $\Delta_{\mathcal{X}} \phi_i(x) = \lambda_i \phi_i(x)$ 
is the eigendecomposition of the Laplacian with eigenvectors $\phi_1, \phi_2, \hdots$ and corresponding non-positive eigenvalues $0=\lambda_1 \geq \lambda_2 \geq \hdots$. The null  eigenvalue is associated with a constant eigenvector.

\noindent In the discrete setting, the heat kernel is given by the positive-definite matrix $\mathbf{K}_\Xx = e^{t \boldsymbol{\Delta}_{\mathcal{X}}} = \boldsymbol{\Phi} e^{t\boldsymbol{\Lambda}_{\mathcal{X}} } \boldsymbol{\Phi}^\top$. 
The constant eigenvector corresponds to the unit eigenvalue, $\mathbf{K}_\Xx \One = \One$. 
%
%Furthermore, the elements of $\mathbf{K}_\Xx$ are strictly positive for any $t>0$ assuming a full basis is used.

\noindent An issue that is often overlooked is the relation between the original and relaxed solution of \eqref{qap3}, which is tightly connected to the choice of pairwise descriptors. Corollary \ref{cor:cor1} asserts the sufficient condition under which this relaxation is exact.
Whereas heat kernels, being (strictly) positive definite, satisfy this condition, distance matrices never do. A distance matrix, having non-negative entries and trace zero, will always, by the Perron-Frobenius theorem, have one large positive eigenvalue and several low magnitude negative eigenvalues\footnote{In the Euclidean case, a distance matrix has exactly one positive eigenvalue and all the rest are negative with small magnitude.} \cite{bogomolny2003spectral}. This distribution of eigenvalues is illustrated in Figure \ref{fig:spectrum}.   

%--------------------------------------------------------------------------------------------------------------------

%\vspace{1ex}\noindent\textbf{Bijective maps %and functional maps.}

\subsection{Bijective maps and functional maps}
\noindent The requirement of bijectivity is what makes a problem \eqref{energyMinimization} computationally hard. A variety of relaxation techniques can be applied to alleviate this complexity. Amongst the most popular are relaxing the column or row sum constraints, relaxing the integer constraints, or restricting the matrix to a sphere of constant norm \cite{leordeanu2005spectral}. A bijective mapping can then be recovered by a post processing step, such as projection onto the set of permutation matrices
\begin{align}\label{eq:proj}
\PiM^* = \argmin_{\PiM\in\Pn}\|\PiM-\PM\|^2 =\argmax_{\PiM\in\Pn}\left<\PiM,\PM\right>\,.
\end{align}
One popular technique in recent years replaces the combinatorially hard point-wise map recovery problem with the simpler problem of finding a linear map between functions\cite{ovsjanikov2012functional}. A \textit{functional map} is a map between functional spaces $T:L^2(\mathcal{X}) \rightarrow L^2(\mathcal{Y})$, which can be disretized (under the previous assumptions of $n$ vertices in each shape) as an $n\times n$ matrix $\mathbf{T}$. 
Providing a pair of orthonormal bases $\boldsymbol{\Phi} = (\boldsymbol{\phi}_1, \hdots, \boldsymbol{\phi}_n)$ and $\boldsymbol{\Psi} = (\boldsymbol{\psi}_1, \hdots, \boldsymbol{\psi}_n)$ for $L^2(\mathcal{X})$ and $L^2(\mathcal{Y})$, respectively, one can express $\mathbf{T} = \boldsymbol{\Psi} \mathbf{C} \boldsymbol{\Phi}^\top$, where $\mathbf{C}$ acts as a basis transformation matrix. Two common choices for basis are the Dirac (or hat) basis, in which the functional map attains the form of  a permutation matrix, and the Laplacian eigenbasis, which is especially suited when the map is smooth, so it can be approximated using a truncated basis of $k$ first basis functions corresponding to the lowest frequencies. The computation of the functional map thus boils down to solving a linear system $\mathbf{C}\boldsymbol{\Phi}^\top \mathbf{F}_\mathcal{X} = \boldsymbol{\Psi}^\top \mathbf{F}_\mathcal{Y}$. The recovery of the point-wise map from the functional map can be obtained by ICP-like procedures~\cite{ovsjanikov2012functional,rodola-vmv15}, with the possible introduction of bijectivity constraints~\cite{PMF}. The fact that the map is band-limited is often erroneously referred to as ``smoothness'' in the literature; however, the bijective map recovered from such a band-limited map is not guaranteed to be continuous let alone smooth (i.e., continuously differentiable).%, as demonstrated in Figure~\ref{fig:}.

%% file: figures/eigs1.tikz
\definecolor{mycolor1}{rgb}{0.00000,0.44700,0.74100}%
\begin{tikzpicture}

\begin{axis}[%
width=0.35\linewidth,
height=0.35\linewidth,
at={(1.162in,0.516in)},
scale only axis,
xmode=log,
xmin=1,
xmax=3400,
xminorticks=true,
xlabel={eig number},
ymin=-0.338378322790773,
ymax=0.924115092115306,
ylabel={$\lambda$},
axis background/.style={fill=white},
title style={font=\bfseries},
title={$\lambda\text{(D)}$},
x tick label style={/pgf/number format/fixed},
y label style={at={(axis description cs:0.25,0.5)},rotate=0,anchor=south},
xticklabels={1, 10, 100, 1000}
]
\addplot [color=mycolor1,solid,line width=2.0pt,forget plot]
  table[row sep=crcr]{%
1	0.924115092115306\\
% 2	0.0042707441691132\\
% 3	0.00363862186991063\\
% 4	0.00135079974539419\\
% 5	0.00121147131777001\\
% 6	0.000984849863553833\\
% 7	0.000765650042362319\\
% 8	0.000643005292104687\\
% 9	0.000574303937575585\\
10	0.000521770717480804\\
% 11	0.000484275342713903\\
% 12	0.000426817266331015\\
% 13	0.000423905628217071\\
% 14	0.000341020111067001\\
% 15	0.000315832111018067\\
% 16	0.000303698689435182\\
% 17	0.000300741243849016\\
% 18	0.000286357133063073\\
% 19	0.000269681179485972\\
20	0.000251924898016778\\
% 21	0.000238039266987723\\
% 22	0.000234709329994858\\
% 23	0.000228132256435507\\
% 24	0.000215887841519486\\
% 25	0.000204288996100387\\
% 26	0.000192166563389289\\
% 27	0.000181080346134718\\
% 28	0.000170857910583566\\
% 29	0.000166501415620392\\
30	0.000165448057511327\\
% 31	0.00015245353102065\\
% 32	0.000144971390102662\\
% 33	0.000142411975779587\\
% 34	0.000141440330770228\\
% 35	0.000135951111918304\\
% 36	0.000131536851952801\\
% 37	0.000125752586175789\\
% 38	0.000125617604868674\\
% 39	0.00012182827187199\\
40	0.000114369005190506\\
% 41	0.000110468871837371\\
% 42	0.000108179854136797\\
% 43	0.000104592601255253\\
% 44	0.000101060831613394\\
% 45	9.86025889177808e-05\\
% 46	9.65701245702896e-05\\
% 47	9.25449023034931e-05\\
% 48	9.14967903077432e-05\\
% 49	9.07357117996418e-05\\
50	9.00977587225976e-05\\
% 51	8.53062299296028e-05\\
% 52	8.4186951030659e-05\\
% 53	8.26244364156269e-05\\
% 54	7.98421379791511e-05\\
% 55	7.94630656286865e-05\\
% 56	7.77341564946721e-05\\
% 57	7.70280456780484e-05\\
% 58	7.67402914621866e-05\\
% 59	7.32769686606794e-05\\
60	7.25133386902543e-05\\
% 61	7.14541255865788e-05\\
% 62	7.11634471136412e-05\\
% 63	6.9591698314807e-05\\
% 64	6.77892750258607e-05\\
% 65	6.64184475031553e-05\\
% 66	6.57159380971116e-05\\
% 67	6.48659132372127e-05\\
% 68	6.28251899762683e-05\\
% 69	6.22209976400076e-05\\
70	6.11132976718284e-05\\
% 71	5.94741372024819e-05\\
% 72	5.74391867624412e-05\\
% 73	5.72874916770841e-05\\
% 74	5.67241678767571e-05\\
% 75	5.65629871104588e-05\\
% 76	5.56142103340462e-05\\
% 77	5.31728317201981e-05\\
% 78	5.2788637274599e-05\\
% 79	5.20826284186899e-05\\
80	5.10899568717895e-05\\
% 81	4.97199363096728e-05\\
% 82	4.89735113536589e-05\\
% 83	4.8786982810142e-05\\
% 84	4.79867120495307e-05\\
% 85	4.7505038901752e-05\\
% 86	4.71494182034665e-05\\
% 87	4.62336910832058e-05\\
% 88	4.54342718649695e-05\\
% 89	4.48208452708369e-05\\
90	4.32640701196754e-05\\
% 91	4.25486435064361e-05\\
% 92	4.18801688696824e-05\\
% 93	4.09419323132571e-05\\
% 94	4.0253321620785e-05\\
% 95	4.00563810181904e-05\\
% 96	3.97974420880604e-05\\
% 97	3.82906156757371e-05\\
% 98	3.80954853310542e-05\\
% 99	3.7417818013734e-05\\
100	3.70729014898921e-05\\
% 101	3.65397055773664e-05\\
% 102	3.61917106968961e-05\\
% 103	3.57628846009248e-05\\
% 104	3.54407443956042e-05\\
% 105	3.5145493783086e-05\\
% 106	3.47749178050033e-05\\
% 107	3.35137544139326e-05\\
% 108	3.31071367137181e-05\\
% 109	3.25786619899402e-05\\
110	3.25300420107646e-05\\
% 111	3.20207728347732e-05\\
% 112	3.08760224521877e-05\\
% 113	3.06657192999244e-05\\
% 114	3.03354598592035e-05\\
% 115	2.99497438363972e-05\\
% 116	2.97005139041921e-05\\
% 117	2.95513492279881e-05\\
% 118	2.92883318299198e-05\\
% 119	2.90626185420277e-05\\
120	2.84318588453966e-05\\
% 121	2.80635500117625e-05\\
% 122	2.80146609020636e-05\\
% 123	2.7676308524737e-05\\
% 124	2.74658292932191e-05\\
% 125	2.74127238404069e-05\\
% 126	2.66054794394097e-05\\
% 127	2.58971033351089e-05\\
% 128	2.55464831228513e-05\\
% 129	2.55021286844708e-05\\
130	2.53278305369108e-05\\
% 131	2.52255909884857e-05\\
% 132	2.47804788115979e-05\\
% 133	2.46615755045725e-05\\
% 134	2.41987754653578e-05\\
% 135	2.40310542592266e-05\\
% 136	2.37657956904283e-05\\
% 137	2.36300120472273e-05\\
% 138	2.34796526683197e-05\\
% 139	2.32317396394162e-05\\
140	2.29208995889126e-05\\
% 141	2.2660944677936e-05\\
% 142	2.25824945551095e-05\\
% 143	2.23297500640388e-05\\
% 144	2.2075290727221e-05\\
% 145	2.18067673319316e-05\\
% 146	2.15531128003592e-05\\
% 147	2.1292113651754e-05\\
% 148	2.10363172965091e-05\\
% 149	2.09856911498708e-05\\
150	2.05688715173263e-05\\
% 151	2.03954961133776e-05\\
% 152	2.01881251026898e-05\\
% 153	2.01440751602755e-05\\
% 154	2.00494693662382e-05\\
% 155	1.96926329218853e-05\\
% 156	1.95620806421929e-05\\
% 157	1.95154018816929e-05\\
% 158	1.92222755540466e-05\\
% 159	1.90273524457602e-05\\
160	1.89418324143535e-05\\
% 161	1.86220921336063e-05\\
% 162	1.84101918434964e-05\\
% 163	1.8205949181992e-05\\
% 164	1.81373217824214e-05\\
% 165	1.78570530643108e-05\\
% 166	1.75585210759121e-05\\
% 167	1.75399339513145e-05\\
% 168	1.74496641917417e-05\\
% 169	1.7258913596763e-05\\
170	1.70766742452546e-05\\
% 171	1.70602791634367e-05\\
% 172	1.69711697400346e-05\\
% 173	1.68792936882045e-05\\
% 174	1.67303238042438e-05\\
% 175	1.63925278420531e-05\\
% 176	1.63417295192406e-05\\
% 177	1.61310693591555e-05\\
% 178	1.60016423530609e-05\\
% 179	1.59931064670836e-05\\
180	1.57856954932558e-05\\
% 181	1.563778767627e-05\\
% 182	1.55102436014427e-05\\
% 183	1.53576735229861e-05\\
% 184	1.52652152870118e-05\\
% 185	1.51902607248963e-05\\
% 186	1.50858048280364e-05\\
% 187	1.49445703107317e-05\\
% 188	1.49234426029347e-05\\
% 189	1.4737731493158e-05\\
190	1.4625175333166e-05\\
% 191	1.44866307360045e-05\\
% 192	1.44026831064621e-05\\
% 193	1.43257603271922e-05\\
% 194	1.41546087377328e-05\\
% 195	1.40279925990497e-05\\
% 196	1.3997911039831e-05\\
% 197	1.39397820874069e-05\\
% 198	1.38035848640345e-05\\
% 199	1.36617939036181e-05\\
200	1.35399965160056e-05\\
% 201	1.34740625375395e-05\\
% 202	1.33550439664989e-05\\
% 203	1.32896597196723e-05\\
% 204	1.3226352429959e-05\\
% 205	1.31385613461742e-05\\
% 206	1.29634317811204e-05\\
% 207	1.28968301725207e-05\\
% 208	1.26984787085158e-05\\
% 209	1.25977764922149e-05\\
210	1.25173429260325e-05\\
% 211	1.24419040397375e-05\\
% 212	1.23324015226457e-05\\
% 213	1.22815633834358e-05\\
% 214	1.21724920379664e-05\\
% 215	1.21063053975887e-05\\
% 216	1.2050394175462e-05\\
% 217	1.19486904639195e-05\\
% 218	1.19015082669238e-05\\
% 219	1.18141725821096e-05\\
220	1.1744230080774e-05\\
% 221	1.16723940835045e-05\\
% 222	1.15687056361732e-05\\
% 223	1.14555339526198e-05\\
% 224	1.14011479112366e-05\\
% 225	1.13274337461905e-05\\
% 226	1.11782722906848e-05\\
% 227	1.11183418280952e-05\\
% 228	1.09495744914991e-05\\
% 229	1.09148003426519e-05\\
230	1.08568850660018e-05\\
% 231	1.08009762571801e-05\\
% 232	1.07121677074718e-05\\
% 233	1.06753565861751e-05\\
% 234	1.0601483992069e-05\\
% 235	1.05063067196311e-05\\
% 236	1.04393832209468e-05\\
% 237	1.03771590220113e-05\\
% 238	1.02823821987932e-05\\
% 239	1.024398058917e-05\\
240	1.01478865808324e-05\\
% 241	1.00782412755177e-05\\
% 242	1.00553154323223e-05\\
% 243	9.9796780494e-06\\
% 244	9.95194804733083e-06\\
% 245	9.90353431333583e-06\\
% 246	9.83729138133226e-06\\
% 247	9.73620086163969e-06\\
% 248	9.54550483463391e-06\\
% 249	9.47494329981277e-06\\
250	9.41911698837614e-06\\
% 251	9.35909598334536e-06\\
% 252	9.31404508812809e-06\\
% 253	9.27708536978406e-06\\
% 254	9.19285801544772e-06\\
% 255	9.12770973084801e-06\\
% 256	9.0661485181688e-06\\
% 257	9.00420016510713e-06\\
% 258	8.98141771068813e-06\\
% 259	8.91439780037141e-06\\
260	8.85645057205551e-06\\
% 261	8.81500637159217e-06\\
% 262	8.72355288782923e-06\\
% 263	8.70281962378676e-06\\
% 264	8.64775971067642e-06\\
% 265	8.62027441037521e-06\\
% 266	8.56087583643568e-06\\
% 267	8.48639258753623e-06\\
% 268	8.40026962714638e-06\\
% 269	8.38007221331534e-06\\
270	8.24715283279013e-06\\
% 271	8.21987013804177e-06\\
% 272	8.17107752429935e-06\\
% 273	8.11264353475569e-06\\
% 274	8.07260356149628e-06\\
% 275	8.0176719907835e-06\\
% 276	7.96997571208996e-06\\
% 277	7.91220831626747e-06\\
% 278	7.87650661065925e-06\\
% 279	7.79153635028397e-06\\
280	7.76468950390542e-06\\
% 281	7.72603679283225e-06\\
% 282	7.63594356432254e-06\\
% 283	7.56966362613227e-06\\
% 284	7.54059888397724e-06\\
% 285	7.46265204204907e-06\\
% 286	7.43798453831143e-06\\
% 287	7.39136794347249e-06\\
% 288	7.35309568309443e-06\\
% 289	7.3140306342857e-06\\
290	7.24324261986963e-06\\
% 291	7.22793921710243e-06\\
% 292	7.16547896031859e-06\\
% 293	7.14986835581097e-06\\
% 294	7.10907097701776e-06\\
% 295	7.07404618394397e-06\\
% 296	7.03758697807182e-06\\
% 297	7.0049409669902e-06\\
% 298	6.93408424744326e-06\\
% 299	6.89524438855324e-06\\
300	6.86793601391135e-06\\
% 301	6.81227992400175e-06\\
% 302	6.77935518471088e-06\\
% 303	6.73536884386571e-06\\
% 304	6.70893879131757e-06\\
% 305	6.68107537056536e-06\\
% 306	6.63733392161966e-06\\
% 307	6.56011409053133e-06\\
% 308	6.55298999891581e-06\\
% 309	6.49810775044107e-06\\
310	6.43406403472048e-06\\
% 311	6.358790672794e-06\\
% 312	6.34101239266212e-06\\
% 313	6.30883448079368e-06\\
% 314	6.2856620869416e-06\\
% 315	6.2089131470254e-06\\
% 316	6.20247327642772e-06\\
% 317	6.18168759032856e-06\\
% 318	6.13432233789717e-06\\
% 319	6.09624388818812e-06\\
320	6.08285297628925e-06\\
% 321	6.0384902313383e-06\\
% 322	6.01009855361535e-06\\
% 323	5.91780540460617e-06\\
% 324	5.88810279023317e-06\\
% 325	5.88158867209029e-06\\
% 326	5.8725627989382e-06\\
% 327	5.83928919639583e-06\\
% 328	5.7918770501459e-06\\
% 329	5.78365515126081e-06\\
330	5.71963047352542e-06\\
% 331	5.68221823271722e-06\\
% 332	5.65814488424007e-06\\
% 333	5.60966159697513e-06\\
% 334	5.56367050800394e-06\\
% 335	5.55329675337989e-06\\
% 336	5.52215071539128e-06\\
% 337	5.50443386618665e-06\\
% 338	5.42477976437681e-06\\
% 339	5.39153230924762e-06\\
340	5.36325137211549e-06\\
% 341	5.32308221097776e-06\\
% 342	5.30709244946356e-06\\
% 343	5.28974926190811e-06\\
% 344	5.20648962976534e-06\\
% 345	5.18900500924517e-06\\
% 346	5.17428431652424e-06\\
% 347	5.13380082300993e-06\\
% 348	5.10156191055861e-06\\
% 349	5.05992510892389e-06\\
350	5.03552572416909e-06\\
% 351	4.99865474043285e-06\\
% 352	4.98349934240711e-06\\
% 353	4.97007422728617e-06\\
% 354	4.91873590526701e-06\\
% 355	4.90699511445555e-06\\
% 356	4.87689317954513e-06\\
% 357	4.83898655200147e-06\\
% 358	4.79404926904079e-06\\
% 359	4.78031015930441e-06\\
360	4.75473921940252e-06\\
% 361	4.73352332341685e-06\\
% 362	4.69892699905165e-06\\
% 363	4.64514062985274e-06\\
% 364	4.61317054586773e-06\\
% 365	4.59756428317992e-06\\
% 366	4.5758868339789e-06\\
% 367	4.54301428649775e-06\\
% 368	4.51791636142718e-06\\
% 369	4.482017788831e-06\\
370	4.47241160211664e-06\\
% 371	4.45163537613158e-06\\
% 372	4.40953232570136e-06\\
% 373	4.36772751064006e-06\\
% 374	4.34137139680835e-06\\
% 375	4.32830532464293e-06\\
% 376	4.30126491783061e-06\\
% 377	4.26248345068826e-06\\
% 378	4.24774539340518e-06\\
% 379	4.21362327885937e-06\\
380	4.19161437172412e-06\\
% 381	4.14714703832368e-06\\
% 382	4.11129621610961e-06\\
% 383	4.09188309399519e-06\\
% 384	4.086232711666e-06\\
% 385	4.05323336805541e-06\\
% 386	4.02022584630543e-06\\
% 387	4.00539752411415e-06\\
% 388	3.98105810423091e-06\\
% 389	3.9721661846924e-06\\
390	3.95105186819561e-06\\
% 391	3.91540797316642e-06\\
% 392	3.88221874654423e-06\\
% 393	3.86561846532594e-06\\
% 394	3.85077673986438e-06\\
% 395	3.83142373801745e-06\\
% 396	3.77419497561265e-06\\
% 397	3.77183447539561e-06\\
% 398	3.73347046861064e-06\\
% 399	3.71659341659622e-06\\
400	3.7001449232173e-06\\
% 401	3.69166365075876e-06\\
% 402	3.67014890578718e-06\\
% 403	3.66615659785158e-06\\
% 404	3.61930281922584e-06\\
% 405	3.59917294533394e-06\\
% 406	3.56692052526508e-06\\
% 407	3.56197715107084e-06\\
% 408	3.53112708898201e-06\\
% 409	3.50225095233398e-06\\
410	3.48944635568389e-06\\
% 411	3.47954603389513e-06\\
% 412	3.46615745352452e-06\\
% 413	3.42079390339452e-06\\
% 414	3.38643381927301e-06\\
% 415	3.38043387867544e-06\\
% 416	3.36980571060666e-06\\
% 417	3.34263387632265e-06\\
% 418	3.32426467982676e-06\\
% 419	3.29991843540794e-06\\
420	3.28086618447371e-06\\
% 421	3.26290999121505e-06\\
% 422	3.21796558854442e-06\\
% 423	3.20738828379017e-06\\
% 424	3.18792579111925e-06\\
% 425	3.18201035097961e-06\\
% 426	3.16858006801229e-06\\
% 427	3.1345389276547e-06\\
% 428	3.12886768992279e-06\\
% 429	3.11817226473535e-06\\
430	3.10404570148953e-06\\
% 431	3.07153617622237e-06\\
% 432	3.04654373081957e-06\\
% 433	3.02103627656164e-06\\
% 434	3.00114747616222e-06\\
% 435	2.98385308043595e-06\\
% 436	2.96318308431318e-06\\
% 437	2.95457268797174e-06\\
% 438	2.93092855651931e-06\\
% 439	2.92289780971328e-06\\
440	2.89787805428059e-06\\
% 441	2.88888569508608e-06\\
% 442	2.86236742124935e-06\\
% 443	2.85028830330978e-06\\
% 444	2.83408237216445e-06\\
% 445	2.82308988751727e-06\\
% 446	2.78921656550881e-06\\
% 447	2.77184194632907e-06\\
% 448	2.75182966025484e-06\\
% 449	2.72120217347956e-06\\
450	2.7139691030351e-06\\
% 451	2.71207733237072e-06\\
% 452	2.69989773068223e-06\\
% 453	2.67242311898821e-06\\
% 454	2.63332687700022e-06\\
% 455	2.62161729311699e-06\\
% 456	2.60834447314518e-06\\
% 457	2.59707986860428e-06\\
% 458	2.5864052447998e-06\\
% 459	2.57092038255229e-06\\
460	2.55138709985919e-06\\
% 461	2.54116542309848e-06\\
% 462	2.51259534687147e-06\\
% 463	2.50655932177903e-06\\
% 464	2.49816199620747e-06\\
% 465	2.49125911301336e-06\\
% 466	2.47003807508904e-06\\
% 467	2.44331244501168e-06\\
% 468	2.43455998239607e-06\\
% 469	2.410158214402e-06\\
470	2.3892579414203e-06\\
% 471	2.37917982106749e-06\\
% 472	2.37089724357719e-06\\
% 473	2.34994243896926e-06\\
% 474	2.34123098974349e-06\\
% 475	2.32659119583584e-06\\
% 476	2.30239578772166e-06\\
% 477	2.28271928692621e-06\\
% 478	2.27616339081701e-06\\
% 479	2.26681748846357e-06\\
480	2.25381156331057e-06\\
% 481	2.24400379891754e-06\\
% 482	2.2230613814104e-06\\
% 483	2.2082791168879e-06\\
% 484	2.19780454386042e-06\\
% 485	2.18625071110204e-06\\
% 486	2.17592859505184e-06\\
% 487	2.16557552844477e-06\\
% 488	2.14811508058192e-06\\
% 489	2.12933398361784e-06\\
490	2.11731609921128e-06\\
% 491	2.11402382749685e-06\\
% 492	2.1050288401542e-06\\
% 493	2.08176540636832e-06\\
% 494	2.06783157782188e-06\\
% 495	2.05883683723512e-06\\
% 496	2.04541202863872e-06\\
% 497	2.02010659372215e-06\\
% 498	2.00823350892384e-06\\
% 499	1.99345795039684e-06\\
500	1.98481019979177e-06\\
% 501	1.97415002836967e-06\\
% 502	1.96454036163156e-06\\
% 503	1.95427263461711e-06\\
% 504	1.9366395119154e-06\\
% 505	1.92579819658771e-06\\
% 506	1.9162357394876e-06\\
% 507	1.89017673900343e-06\\
% 508	1.8705489506494e-06\\
% 509	1.85969651601949e-06\\
510	1.84510359149276e-06\\
% 511	1.82182397790917e-06\\
% 512	1.81693688752557e-06\\
% 513	1.79633912822926e-06\\
% 514	1.77936035311156e-06\\
% 515	1.77658613573041e-06\\
% 516	1.76534339894718e-06\\
% 517	1.75976857595298e-06\\
% 518	1.75074830222398e-06\\
% 519	1.72272097056196e-06\\
520	1.71317071610021e-06\\
% 521	1.69993357737104e-06\\
% 522	1.6861126714503e-06\\
% 523	1.67394642950577e-06\\
% 524	1.6618442745574e-06\\
% 525	1.65014227554516e-06\\
% 526	1.63708918409731e-06\\
% 527	1.62913052213605e-06\\
% 528	1.61123805842578e-06\\
% 529	1.60662201016508e-06\\
530	1.60227277911807e-06\\
613	7.34298674884874e-07\\
688	1.29598112562524e-07\\
% 689	1.18443761511375e-07\\
% 690	1.16198868595325e-07\\
% 691	1.04935918095649e-07\\
% 692	9.6621076125045e-08\\
% 693	9.21692579883711e-08\\
% 694	8.55210408642009e-08\\
% 695	7.99889918145263e-08\\
% 696	7.71801243012577e-08\\
% 697	7.281510050523e-08\\
% 698	6.99441331608059e-08\\
% 699	5.77023048565829e-08\\
% 700	5.05405517594998e-08\\
% 701	4.46028593155008e-08\\
% 702	3.77291882141693e-08\\
% 703	3.26238350468791e-08\\
% 704	2.27887182263948e-08\\
% 705	1.74759688256955e-08\\
% 706	1.31531562377017e-08\\
% 707	3.75508682524429e-09\\
% 708	-1.50269470547857e-09\\
% 709	-9.32837136587285e-09\\
% 710	-1.26087457105721e-08\\
% 711	-2.23350563312029e-08\\
% 712	-2.75099555920359e-08\\
% 713	-3.63574326014854e-08\\
% 714	-4.36582152886624e-08\\
% 715	-5.28405537972431e-08\\
% 716	-5.64985808424801e-08\\
% 717	-7.00861054186276e-08\\
% 718	-7.4232697115908e-08\\
% 719	-8.21579269524028e-08\\
% 720	-8.53416770231407e-08\\
% 721	-9.45981220229199e-08\\
% 722	-1.08807826331961e-07\\
% 723	-1.12054794173027e-07\\
% 724	-1.2029565203968e-07\\
% 725	-1.22362402625364e-07\\
% 726	-1.29998620753437e-07\\
% 727	-1.30498153053924e-07\\
% 728	-1.44927379954864e-07\\
% 729	-1.48716973117046e-07\\
% 730	-1.51483023082172e-07\\
% 731	-1.61035193975954e-07\\
% 732	-1.70426684316358e-07\\
% 733	-1.71590357696476e-07\\
% 734	-1.7961266491965e-07\\
% 735	-1.90276914418849e-07\\
% 736	-1.90348535996269e-07\\
% 737	-1.98118452098899e-07\\
% 738	-2.0237089056544e-07\\
% 739	-2.12104375895458e-07\\
% 740	-2.15565161100926e-07\\
% 741	-2.20997257362512e-07\\
% 742	-2.25843020549264e-07\\
% 743	-2.27371427004565e-07\\
% 744	-2.39951813019612e-07\\
745	-2.43022842456958e-07\\
853	-8.16035100284933e-07\\
939	-1.1900611261926e-06\\
% 940	-1.19488451950644e-06\\
% 941	-1.19626015738333e-06\\
% 942	-1.19940058885962e-06\\
% 943	-1.20271877359466e-06\\
% 944	-1.20854496071717e-06\\
% 945	-1.21242832391713e-06\\
% 946	-1.21773317745342e-06\\
% 947	-1.22114412811475e-06\\
% 948	-1.22439323654118e-06\\
% 949	-1.22523689365877e-06\\
% 950	-1.23091669568578e-06\\
% 951	-1.23688124265662e-06\\
% 952	-1.2422960432851e-06\\
953	-1.24260322898333e-06\\
% 954	-1.24486027476129e-06\\
% 955	-1.25099792904872e-06\\
% 956	-1.25210323440242e-06\\
% 957	-1.26168360401958e-06\\
% 958	-1.26860175374969e-06\\
% 959	-1.27148422961948e-06\\
% 960	-1.27466219621506e-06\\
% 961	-1.27579943673965e-06\\
% 962	-1.2830814265617e-06\\
% 963	-1.28610763815463e-06\\
% 964	-1.29085475919435e-06\\
965	-1.29197922179442e-06\\
% 966	-1.29912802499204e-06\\
% 967	-1.30106238191663e-06\\
% 968	-1.30454485675569e-06\\
% 969	-1.31184170849449e-06\\
% 970	-1.31290979307551e-06\\
% 971	-1.31492345250543e-06\\
% 972	-1.31789971057906e-06\\
% 973	-1.32228250749112e-06\\
% 974	-1.32732381800552e-06\\
% 975	-1.32895008763576e-06\\
% 976	-1.3348229114309e-06\\
977	-1.33999416267735e-06\\
% 978	-1.34093991031791e-06\\
% 979	-1.34511687015779e-06\\
% 980	-1.35128131570842e-06\\
% 981	-1.35259974992094e-06\\
% 982	-1.35553122738269e-06\\
% 983	-1.36087090076531e-06\\
% 984	-1.36279879970544e-06\\
% 985	-1.37062492407876e-06\\
% 986	-1.37597860771174e-06\\
% 987	-1.37843107089241e-06\\
% 988	-1.38087641991091e-06\\
989	-1.38500289551598e-06\\
% 990	-1.38803979822287e-06\\
% 991	-1.39246021347458e-06\\
% 992	-1.39370347668648e-06\\
% 993	-1.39880676482308e-06\\
% 994	-1.40125139291279e-06\\
% 995	-1.40607851811556e-06\\
% 996	-1.4085776930051e-06\\
% 997	-1.41022637641853e-06\\
% 998	-1.41664799431774e-06\\
999	-1.41831259147177e-06\\
1000	-1.42322582648698e-06\\
% 1001	-1.42575779957536e-06\\
% 1002	-1.43021279318399e-06\\
% 1003	-1.43266547693565e-06\\
% 1004	-1.43730654475101e-06\\
% 1005	-1.4443248397306e-06\\
% 1006	-1.45431798848427e-06\\
% 1007	-1.4556093865463e-06\\
% 1008	-1.46206093682385e-06\\
% 1009	-1.4670829698629e-06\\
% 1010	-1.4681246841519e-06\\
1011	-1.47114569656208e-06\\
% 1012	-1.47357087934433e-06\\
% 1013	-1.47941963640123e-06\\
% 1014	-1.48060495181354e-06\\
% 1015	-1.48251925859076e-06\\
% 1016	-1.48412297860328e-06\\
% 1017	-1.48711234399575e-06\\
% 1018	-1.49120418319073e-06\\
% 1019	-1.50091231731501e-06\\
% 1020	-1.50148090681256e-06\\
% 1021	-1.5058573346621e-06\\
% 1022	-1.50614103691088e-06\\
% 1023	-1.5100450646546e-06\\
% 1024	-1.51351672356664e-06\\
% 1025	-1.51792677869804e-06\\
% 1026	-1.52259224360194e-06\\
% 1027	-1.52747154984256e-06\\
% 1028	-1.53074775585162e-06\\
% 1029	-1.53252833190381e-06\\
% 1030	-1.53948651181634e-06\\
% 1031	-1.5416165182659e-06\\
% 1032	-1.54344836654024e-06\\
% 1033	-1.54701013887212e-06\\
% 1034	-1.55004056080078e-06\\
% 1035	-1.55438351357249e-06\\
% 1036	-1.55700374139775e-06\\
% 1037	-1.55835271262107e-06\\
% 1038	-1.56372003153282e-06\\
% 1039	-1.57280521811126e-06\\
% 1040	-1.57505467253469e-06\\
% 1041	-1.57950643029592e-06\\
% 1042	-1.58210744047716e-06\\
% 1043	-1.58501517962164e-06\\
% 1044	-1.58988679644905e-06\\
% 1045	-1.5970402043727e-06\\
% 1046	-1.599163595746e-06\\
% 1047	-1.60243040987404e-06\\
% 1048	-1.60671665230206e-06\\
% 1049	-1.61072653207445e-06\\
1050	-1.61164897186283e-06\\
% 1051	-1.61916534583955e-06\\
% 1052	-1.62287994457632e-06\\
% 1053	-1.6263597950491e-06\\
% 1054	-1.62776958703723e-06\\
% 1055	-1.63303604665761e-06\\
% 1056	-1.63788853264597e-06\\
% 1057	-1.63915840479347e-06\\
% 1058	-1.64480575333445e-06\\
% 1059	-1.64869546125447e-06\\
% 1060	-1.64951847052394e-06\\
% 1061	-1.65567549661203e-06\\
% 1062	-1.65787308078732e-06\\
% 1063	-1.65967715571257e-06\\
% 1064	-1.66366377843976e-06\\
% 1065	-1.67078143720157e-06\\
% 1066	-1.67242988785463e-06\\
% 1067	-1.6794615882473e-06\\
% 1068	-1.68382730577799e-06\\
% 1069	-1.68629501432557e-06\\
% 1070	-1.69072325364007e-06\\
% 1071	-1.69363814910159e-06\\
% 1072	-1.69686243225505e-06\\
1073	-1.6979498042106e-06\\
% 1074	-1.70199264210208e-06\\
% 1075	-1.70270183898246e-06\\
% 1076	-1.70824566404922e-06\\
% 1077	-1.71296758296808e-06\\
% 1078	-1.71574953041728e-06\\
% 1079	-1.7189876378087e-06\\
% 1080	-1.72561519348193e-06\\
% 1081	-1.72664154133344e-06\\
% 1082	-1.7334212196061e-06\\
% 1083	-1.73687545719498e-06\\
% 1084	-1.73852963222228e-06\\
% 1085	-1.74044529043667e-06\\
% 1086	-1.74628261008769e-06\\
% 1087	-1.74941030275551e-06\\
% 1088	-1.75377506379089e-06\\
% 1089	-1.75746369403088e-06\\
% 1090	-1.76315953138238e-06\\
% 1091	-1.76651908961764e-06\\
% 1092	-1.76935359027528e-06\\
% 1093	-1.77152784430536e-06\\
% 1094	-1.77890559642482e-06\\
% 1095	-1.77955447371514e-06\\
% 1096	-1.78173461136713e-06\\
% 1097	-1.78620368310504e-06\\
% 1098	-1.79171742276273e-06\\
% 1099	-1.79400147086454e-06\\
% 1100	-1.79921302986887e-06\\
% 1101	-1.80140908474634e-06\\
% 1102	-1.80498131191569e-06\\
% 1103	-1.80997405041208e-06\\
% 1104	-1.813095050759e-06\\
% 1105	-1.81787849124877e-06\\
% 1106	-1.81904690921949e-06\\
% 1107	-1.82333935134982e-06\\
% 1108	-1.8255698740061e-06\\
1109	-1.82951198946674e-06\\
% 1110	-1.83323176476848e-06\\
% 1111	-1.83716757824076e-06\\
% 1112	-1.84455401201599e-06\\
% 1113	-1.84690010781078e-06\\
% 1114	-1.85100875547987e-06\\
% 1115	-1.85243464456092e-06\\
% 1116	-1.85810911735532e-06\\
% 1117	-1.86245727551551e-06\\
% 1118	-1.8694908619857e-06\\
% 1119	-1.87279317489457e-06\\
% 1120	-1.87524198940843e-06\\
% 1121	-1.88056894925439e-06\\
% 1122	-1.88176812140411e-06\\
% 1123	-1.88418213457234e-06\\
% 1124	-1.88676760441242e-06\\
% 1125	-1.88798768735639e-06\\
% 1126	-1.89360588334156e-06\\
% 1127	-1.90235086927864e-06\\
% 1128	-1.90657575327543e-06\\
% 1129	-1.90767106331983e-06\\
% 1130	-1.90997392234249e-06\\
% 1131	-1.91640755420342e-06\\
% 1132	-1.92062171668185e-06\\
% 1133	-1.9215460855465e-06\\
% 1134	-1.92643185979176e-06\\
% 1135	-1.92897555049399e-06\\
% 1136	-1.93578430660715e-06\\
% 1137	-1.93706566402537e-06\\
% 1138	-1.94259373382369e-06\\
% 1139	-1.94480798200713e-06\\
% 1140	-1.95049110769435e-06\\
% 1141	-1.95652665967063e-06\\
% 1142	-1.95940981256724e-06\\
% 1143	-1.96599252179176e-06\\
% 1144	-1.968922662563e-06\\
% 1145	-1.97291000736527e-06\\
% 1146	-1.97618372528944e-06\\
% 1147	-1.97991306485526e-06\\
% 1148	-1.98370727158039e-06\\
% 1149	-1.98796216315043e-06\\
% 1150	-1.98915307577818e-06\\
% 1151	-1.99284910110932e-06\\
% 1152	-1.99903386644703e-06\\
% 1153	-2.00001208884294e-06\\
% 1154	-2.00261789977925e-06\\
% 1155	-2.00747530736704e-06\\
% 1156	-2.0143313993301e-06\\
% 1157	-2.01686080680598e-06\\
% 1158	-2.0185402316751e-06\\
% 1159	-2.02145199707163e-06\\
% 1160	-2.02577891382062e-06\\
% 1161	-2.0306208677904e-06\\
% 1162	-2.03257095852076e-06\\
1163	-2.03635140099854e-06\\
1268	-2.4517256361517e-06\\
1353	-2.80514656283411e-06\\
1479	-3.36583302450084e-06\\
1649	-4.25354368882488e-06\\
1836	-5.39996104534697e-06\\
2299	-1.00147887394299e-05\\
2740	-2.14296384042979e-05\\
3133	-7.34756747305788e-05\\
3400	-0.338378322790773\\
};
\end{axis}
\end{tikzpicture}%

%% file: figures/eigs2.tikz
\definecolor{mycolor1}{rgb}{0.00000,0.44700,0.74100}%
\definecolor{mycolor2}{rgb}{0.85000,0.32500,0.09800}%
\definecolor{mycolor3}{rgb}{0.92900,0.69400,0.12500}%
\definecolor{mycolor4}{rgb}{0.49400,0.18400,0.55600}%
\begin{tikzpicture}

\begin{axis}[%
width=0.35\linewidth,
height=0.35\linewidth,
at={(1.162in,0.516in)},
scale only axis,
xmin=1,
xmax=3400,
xlabel={eig number},
ymin=0,
ymax=1,
ylabel={$\lambda$},
axis background/.style={fill=white},
title style={font=\bfseries},
title={$\lambda\text{(K)}$},
y label style={at={(axis description cs:0.25,0.5)},rotate=0,anchor=south},
xticklabels={1, 1, 1000, 2000, 3000}
]
\addplot [color=mycolor1,solid,line width=2.0pt,forget plot]
  table[row sep=crcr]{%
1	1\\
% 2	0.999993431936178\\
% 3	0.999988434625623\\
% 4	0.999981106081743\\
% 5	0.999971938591314\\
% 6	0.999970833169205\\
% 7	0.9999666982696\\
% 8	0.999949516309625\\
% 9	0.999928646237768\\
10	0.999924497484139\\
% 11	0.999916741712662\\
% 12	0.99989352700742\\
% 13	0.999890120989805\\
% 14	0.99988569755708\\
% 15	0.999865459861497\\
% 16	0.99983881834893\\
% 17	0.999832840703215\\
% 18	0.999821967067234\\
% 19	0.999821340177982\\
20	0.999784697548632\\
% 21	0.99977658587653\\
% 22	0.999773499573687\\
% 23	0.99976507864436\\
% 24	0.999741973006607\\
% 25	0.99972612543366\\
% 26	0.999714187752701\\
% 27	0.999707073369322\\
% 28	0.999686265918425\\
% 29	0.999652071201398\\
30	0.999648396588062\\
% 31	0.999637785840054\\
% 32	0.999634483210134\\
% 33	0.999612009969173\\
% 34	0.999604714513178\\
% 35	0.999580310768558\\
% 36	0.999561632103793\\
% 37	0.999545359327998\\
% 38	0.999530312688222\\
% 39	0.999514354561653\\
% 40	0.99950986240765\\
% 41	0.999490177513625\\
% 42	0.999488426013796\\
% 43	0.999457050421\\
% 44	0.999449953151679\\
% 45	0.999435561827648\\
% 46	0.999412543289451\\
% 47	0.999411229173536\\
% 48	0.999394702933987\\
% 49	0.999393045493379\\
50	0.999352124898842\\
% 51	0.999348367165053\\
% 52	0.99933084072156\\
% 53	0.99931837659247\\
% 54	0.999309367476227\\
% 55	0.999292633901753\\
% 56	0.999272956028575\\
% 57	0.999265009896349\\
% 58	0.999235544272386\\
% 59	0.999229781343636\\
60	0.999220530664189\\
% 61	0.99919532230236\\
% 62	0.999190397646043\\
% 63	0.999160230370083\\
% 64	0.999154966392054\\
% 65	0.999139187299888\\
% 66	0.999135216212495\\
% 67	0.999103888807746\\
% 68	0.999101495816937\\
% 69	0.99909912477216\\
70	0.999090337632951\\
% 71	0.999078790611528\\
% 72	0.99906340797791\\
% 73	0.99903260844334\\
% 74	0.99902724651408\\
% 75	0.999001287560576\\
% 76	0.998995468483161\\
% 77	0.998972291526698\\
% 78	0.998954390632932\\
% 79	0.998951523198886\\
80	0.998939497350323\\
% 81	0.998924039115616\\
% 82	0.998910431874585\\
% 83	0.998907799119226\\
% 84	0.998897602401865\\
% 85	0.998892172043533\\
% 86	0.998882288946783\\
% 87	0.998854458997027\\
% 88	0.998846975454047\\
% 89	0.99883631416121\\
90	0.998819394990295\\
% 91	0.998814490192519\\
% 92	0.99880784277616\\
% 93	0.99879576198223\\
% 94	0.998755050555506\\
% 95	0.998753191220131\\
% 96	0.998748559053739\\
% 97	0.998743120405115\\
% 98	0.998724589731835\\
% 99	0.998706882761516\\
100	0.998694793297162\\
% 101	0.998669843879853\\
% 102	0.998660054769847\\
% 103	0.998649567990119\\
% 104	0.998640820474704\\
% 105	0.998636644792015\\
% 106	0.998630838779362\\
% 107	0.998609281959871\\
% 108	0.998599383258719\\
% 109	0.998567629134593\\
% 110	0.998562738189308\\
% 111	0.99855495489124\\
% 112	0.998548060654575\\
% 113	0.99854149404917\\
% 114	0.998523429109717\\
% 115	0.998516768540995\\
% 116	0.998504331518944\\
% 117	0.998499049781255\\
% 118	0.998488336009111\\
% 119	0.998471065538464\\
% 120	0.998448179263694\\
% 121	0.998440286186334\\
% 122	0.99842475751363\\
% 123	0.998423785767307\\
% 124	0.998418726673387\\
% 125	0.998416321983976\\
% 126	0.998413389705679\\
% 127	0.998400137654993\\
% 128	0.998396872050307\\
% 129	0.998392711439093\\
% 130	0.998375066946001\\
% 131	0.998372381507985\\
% 132	0.998357944911027\\
% 133	0.99835523745233\\
% 134	0.998341789353882\\
% 135	0.998340166120592\\
% 136	0.99832313501097\\
% 137	0.998306769768531\\
% 138	0.998298419600695\\
% 139	0.998279921600708\\
% 140	0.99826347810554\\
% 141	0.998254540335328\\
% 142	0.998237006512388\\
% 143	0.998224096369799\\
144	0.998195554444274\\
218	0.997338293660006\\
339	0.995876204343922\\
478	0.994294482206801\\
651	0.992327864937129\\
751	0.991154348679646\\
866	0.989709413417078\\
956	0.988578924945253\\
% 957	0.988572876387934\\
% 958	0.988557040968636\\
% 959	0.98855151647574\\
% 960	0.988546510217227\\
% 961	0.988543841105727\\
% 962	0.988526305822273\\
% 963	0.988521739813992\\
% 964	0.988486716535369\\
% 965	0.988464644508106\\
% 966	0.988459389396841\\
% 967	0.988450076468875\\
% 968	0.988425279595366\\
% 969	0.988417556603115\\
% 970	0.988388052288918\\
% 971	0.988386111055044\\
% 972	0.98837910263519\\
% 973	0.988358622719911\\
% 974	0.988347918140643\\
% 975	0.988344389738441\\
% 976	0.988333955805854\\
% 977	0.988325928287482\\
% 978	0.9883207438147\\
% 979	0.988318424416938\\
% 980	0.988297070543062\\
% 981	0.988267341772527\\
% 982	0.988264504747366\\
% 983	0.988235398722816\\
% 984	0.988232170251145\\
% 985	0.988230524426973\\
% 986	0.988224450505449\\
% 987	0.988214879366392\\
% 988	0.98821249701256\\
% 989	0.988205119832156\\
% 990	0.988194414162769\\
% 991	0.98818002525166\\
% 992	0.988171105597547\\
% 993	0.988153870719335\\
% 994	0.988132463554238\\
% 995	0.988125081312784\\
% 996	0.988110646428951\\
% 997	0.988098019555916\\
% 998	0.98809396827907\\
% 999	0.98808179097546\\
% 1000	0.988065563939412\\
% 1001	0.988024867980401\\
% 1002	0.98801734103714\\
% 1003	0.987997322305944\\
% 1004	0.987987173573014\\
% 1005	0.987984643370031\\
% 1006	0.987962037186242\\
% 1007	0.987943562932897\\
% 1008	0.987935722045013\\
% 1009	0.987927309912144\\
% 1010	0.987920226404871\\
% 1011	0.987889232387429\\
% 1012	0.987868097469422\\
% 1013	0.987863761488746\\
% 1014	0.987860649098448\\
% 1015	0.987858872565066\\
1016	0.987852273712738\\
1186	0.98555451578257\\
1440	0.981787617005051\\
1612	0.979020677250132\\
1750	0.976567227152383\\
1939	0.97290668964368\\
2262	0.965262265915809\\
2836	0.940696001461333\\
3255	0.883562574612618\\
3371	0.77517560846163\\
% 3372	0.774567627054867\\
% 3373	0.774567627054867\\
% 3374	0.765390807113328\\
% 3375	0.765390807113328\\
% 3376	0.763002642158665\\
% 3377	0.757990911066467\\
% 3378	0.757990840199263\\
% 3379	0.748305329975197\\
% 3380	0.748191138676628\\
% 3381	0.748191138676628\\
% 3382	0.743836665700528\\
% 3383	0.743833981240195\\
% 3384	0.719518106675302\\
% 3385	0.719516446556612\\
% 3386	0.713670205291908\\
% 3387	0.713670193516162\\
% 3388	0.695998351113086\\
% 3389	0.686848730650843\\
3390	0.686848456973418\\
% 3391	0.681570539123949\\
% 3392	0.681570536510568\\
% 3393	0.642629351344138\\
% 3394	0.642629351344138\\
% 3395	0.632214578670345\\
% 3396	0.632214557231657\\
% 3397	0.603139561014564\\
% 3398	0.593523705757986\\
% 3399	0.536875014172309\\
3400	0.53687501360588\\
};
\node[right, align=left, text=mycolor1]
at (axis cs:1770,0.87) {t = 0.01};
\addplot [color=mycolor2,solid,line width=2.0pt,forget plot]
  table[row sep=crcr]{%
1	1\\
% 2	0.999868646919733\\
% 3	0.999768717924701\\
% 4	0.999622189453396\\
% 5	0.999438921415189\\
% 6	0.999416824989584\\
% 7	0.999334176060893\\
% 8	0.998990810280424\\
% 9	0.998573891699629\\
10	0.998491032311952\\
% 11	0.998336150664596\\
% 12	0.997872692707662\\
% 13	0.997804712229882\\
% 14	0.997716431799203\\
% 15	0.997312633654545\\
% 16	0.996781298317893\\
% 17	0.996662117767161\\
% 18	0.99644535710443\\
% 19	0.996432861736568\\
20	0.995702747083066\\
% 21	0.995541188465394\\
% 22	0.995479725703849\\
% 23	0.995312043850248\\
% 24	0.994852090376227\\
% 25	0.994536736664722\\
% 26	0.994299249311782\\
% 27	0.994157741910413\\
% 28	0.993743984735607\\
% 29	0.993064376429271\\
30	0.992991371025145\\
% 31	0.992780590537671\\
% 32	0.99271499309783\\
% 33	0.9922687347998\\
% 34	0.99212390758824\\
% 35	0.991639597667982\\
% 36	0.991269057839756\\
% 37	0.990946352283724\\
% 38	0.990648051050198\\
% 39	0.990331772709758\\
% 40	0.990242758822592\\
41	0.989852784138845\\
% 42	0.989818092489856\\
% 43	0.989196837280333\\
% 44	0.989056358552051\\
% 45	0.988771564229355\\
% 46	0.988316205271151\\
% 47	0.988290215086419\\
% 48	0.987963419571614\\
% 49	0.987930650438717\\
% 50	0.987121939824407\\
% 51	0.987047707551825\\
% 52	0.986701550906406\\
53	0.986455447856329\\
% 54	0.986277600016994\\
% 55	0.985947345442309\\
% 56	0.985559116466262\\
% 57	0.985402386687295\\
% 58	0.98482141239846\\
% 59	0.984707822670444\\
% 60	0.984525513946609\\
% 61	0.984028880272229\\
% 62	0.98393188668068\\
% 63	0.983337925153907\\
% 64	0.983234317944303\\
% 65	0.982923811205172\\
% 66	0.982845681370424\\
% 67	0.98222953183758\\
% 68	0.982182481419965\\
% 69	0.982135864611763\\
% 70	0.981963120120017\\
% 71	0.98173616357989\\
% 72	0.98143389554779\\
% 73	0.980828951841403\\
% 74	0.980723672648931\\
% 75	0.98021413145601\\
76	0.980099944890686\\
% 77	0.979645273602153\\
% 78	0.979294242024243\\
% 79	0.979238023540106\\
% 80	0.979002279938628\\
% 81	0.978699330210749\\
% 82	0.978432729870306\\
% 83	0.978381155486192\\
% 84	0.978181431172588\\
% 85	0.978075081906199\\
% 86	0.977881557471682\\
% 87	0.977336804741538\\
% 88	0.977190368564175\\
% 89	0.976981786937316\\
% 90	0.97665086059998\\
% 91	0.976554946333004\\
% 92	0.976424969105506\\
% 93	0.976188794878183\\
% 94	0.975393303852681\\
% 95	0.975356987616793\\
% 96	0.975266518483131\\
% 97	0.975160308417004\\
% 98	0.974798509833997\\
% 99	0.97445291262882\\
100	0.974217022413194\\
% 101	0.973730379656874\\
% 102	0.973539504438319\\
% 103	0.973335064981753\\
% 104	0.973164563630914\\
% 105	0.973083183720884\\
% 106	0.972970041042426\\
% 107	0.972550069254863\\
% 108	0.972357279619145\\
% 109	0.97173907318531\\
% 110	0.971643886813173\\
% 111	0.971492428447457\\
% 112	0.971358289421787\\
% 113	0.971230541376972\\
% 114	0.97087918480538\\
% 115	0.970749669611697\\
% 116	0.970507874842875\\
% 117	0.970405207077139\\
% 118	0.970196981729868\\
% 119	0.969861414354563\\
% 120	0.969416901080204\\
% 121	0.969263641088668\\
% 122	0.968962187824453\\
% 123	0.968943326578676\\
% 124	0.968845137024278\\
% 125	0.96879846886244\\
% 126	0.968741564595087\\
% 127	0.968484432753728\\
% 128	0.96842107961683\\
% 129	0.968340368945431\\
% 130	0.967998158781679\\
% 131	0.967946085513483\\
% 132	0.967666191391142\\
133	0.967613708235904\\
% 134	0.967353061546829\\
% 135	0.967321605076616\\
% 136	0.966991619542901\\
% 137	0.966674636243063\\
% 138	0.966512937367918\\
% 139	0.966154819812485\\
% 140	0.965836582893489\\
% 141	0.965663648764884\\
% 142	0.96532447774644\\
% 143	0.965074818687262\\
% 144	0.964523086608411\\
% 145	0.964482943385459\\
% 146	0.964173765614567\\
% 147	0.963986929308211\\
% 148	0.963880399180092\\
% 149	0.963750136177362\\
% 150	0.963741445722053\\
% 151	0.963541758083507\\
% 152	0.963105231909387\\
% 153	0.962808224738367\\
% 154	0.962744822307767\\
% 155	0.962604477613192\\
% 156	0.962150848708859\\
% 157	0.96205431078147\\
% 158	0.961942763842858\\
% 159	0.961598138348663\\
% 160	0.96118234960359\\
% 161	0.961053029329937\\
% 162	0.960855104543263\\
% 163	0.960709280870407\\
% 164	0.960361427433014\\
% 165	0.959992118452611\\
% 166	0.959951755086097\\
% 167	0.95991723919163\\
% 168	0.959773345890205\\
% 169	0.959345697632122\\
% 170	0.959288011657497\\
% 171	0.958967316456483\\
% 172	0.958879960829808\\
% 173	0.958505236950207\\
% 174	0.95834789801749\\
% 175	0.958049066102521\\
% 176	0.95783225436024\\
% 177	0.957650856631057\\
% 178	0.957563245680604\\
% 179	0.957068215577503\\
% 180	0.956959082609149\\
% 181	0.956684984789436\\
% 182	0.95652482477064\\
% 183	0.956159564425533\\
% 184	0.956146996613538\\
% 185	0.955767466818674\\
186	0.95560941983463\\
% 187	0.955514592428396\\
% 188	0.95534876328048\\
% 189	0.955051027750805\\
% 190	0.954791209999392\\
% 191	0.95462003632969\\
% 192	0.954392425013067\\
% 193	0.954310287184607\\
% 194	0.954090496807186\\
% 195	0.953924031226743\\
% 196	0.953526620870343\\
% 197	0.953155680137088\\
% 198	0.953006565500517\\
% 199	0.952946712526535\\
% 200	0.952761597645327\\
% 201	0.952455386549859\\
% 202	0.952273184600103\\
% 203	0.95167621055238\\
% 204	0.951519351054134\\
% 205	0.951414594810718\\
% 206	0.951249806391501\\
% 207	0.950876385883891\\
% 208	0.950676406341927\\
% 209	0.950383829365009\\
% 210	0.950089592453839\\
% 211	0.949889271575858\\
% 212	0.949734562770016\\
% 213	0.949452728459469\\
% 214	0.949150160879944\\
% 215	0.948992064664095\\
% 216	0.948779740693873\\
% 217	0.948526956051362\\
% 218	0.948090706280929\\
% 219	0.947931828095749\\
% 220	0.947879143949444\\
% 221	0.947552256054186\\
% 222	0.947454970463347\\
% 223	0.947214637423265\\
% 224	0.946893782646473\\
% 225	0.946491198401861\\
% 226	0.946254323406843\\
% 227	0.945722580513911\\
% 228	0.945638764818316\\
% 229	0.94529661474475\\
% 230	0.945165982944725\\
% 231	0.9451182302895\\
% 232	0.944490355818762\\
% 233	0.944364511614282\\
% 234	0.944247250899279\\
% 235	0.944062411586102\\
% 236	0.943524542252108\\
% 237	0.943285047385673\\
% 238	0.942770014161296\\
% 239	0.942715373326587\\
% 240	0.942216366046328\\
% 241	0.942058692343692\\
% 242	0.942001478332475\\
% 243	0.941721835613978\\
244	0.941584940776628\\
339	0.92067660502476\\
% 340	0.920328700850751\\
% 341	0.919958650054166\\
% 342	0.919764538425085\\
% 343	0.919727240053331\\
% 344	0.919596682592001\\
% 345	0.919479493230373\\
% 346	0.91927282882497\\
% 347	0.919043018597524\\
% 348	0.918791054794\\
% 349	0.918624260236379\\
% 350	0.918508771678538\\
% 351	0.918231029557142\\
% 352	0.91816941094813\\
% 353	0.917828397585223\\
% 354	0.916480694498693\\
% 355	0.916225912654435\\
% 356	0.916083878420087\\
% 357	0.915963088866693\\
% 358	0.915805250534741\\
% 359	0.915474782153737\\
% 360	0.915391037543005\\
% 361	0.915192648805151\\
% 362	0.914742242791457\\
% 363	0.914715230656545\\
% 364	0.914366311048651\\
% 365	0.914289332698661\\
% 366	0.913890785922866\\
% 367	0.913854628150184\\
% 368	0.913652559215655\\
% 369	0.913377181996977\\
% 370	0.913133790232008\\
% 371	0.912757741916614\\
% 372	0.91260503772976\\
% 373	0.912428056470752\\
% 374	0.912223849238953\\
% 375	0.912189800540324\\
% 376	0.91157172658276\\
% 377	0.911495227387868\\
% 378	0.91122458812685\\
% 379	0.911131844109492\\
% 380	0.910971059336048\\
% 381	0.910579279906959\\
% 382	0.910417655815399\\
% 383	0.910346666426619\\
% 384	0.910158347323806\\
% 385	0.910139782532293\\
% 386	0.909909341453196\\
% 387	0.90981354261095\\
% 388	0.909756655727785\\
% 389	0.909561276709927\\
% 390	0.90945905744364\\
% 391	0.909281320407105\\
% 392	0.908892176087327\\
% 393	0.908685545405618\\
% 394	0.908616806011784\\
% 395	0.908210626442605\\
% 396	0.908119145812374\\
% 397	0.907866073069434\\
% 398	0.907453214555139\\
% 399	0.907428641319222\\
% 400	0.90726234540883\\
% 401	0.907098518336838\\
% 402	0.906882362005239\\
403	0.906490350213883\\
527	0.88159563146584\\
663	0.854657497936666\\
882	0.810337037577421\\
1109	0.761895864012148\\
1335	0.715360364535949\\
1582	0.660685650204177\\
1985	0.565296464029775\\
2552	0.400979229841649\\
3042	0.199376845815878\\
3400	3.95789112786604e-06\\
};
\node[right, align=left, text=mycolor2]
at (axis cs:1870,0.633) {t = 0.2};
\addplot [color=mycolor3,solid,line width=2.0pt,forget plot]
  table[row sep=crcr]{%
1	0.999999999999999\\
% 2	0.99934340711232\\
% 3	0.998844124413788\\
% 4	0.998112374135886\\
% 5	0.997197753401894\\
% 6	0.997087523896089\\
% 7	0.996675310568887\\
% 8	0.994964225767973\\
% 9	0.992889767363688\\
10	0.992477897061564\\
% 11	0.991708391245483\\
% 12	0.98940862173389\\
% 13	0.989071648352175\\
% 14	0.988634186888334\\
% 15	0.986635193832188\\
% 16	0.984009759072155\\
% 17	0.983421632145315\\
% 18	0.9823526920346\\
19	0.982291100346099\\
% 20	0.978697607398058\\
% 21	0.977903867848157\\
% 22	0.977602035779784\\
% 23	0.976778960723548\\
% 24	0.974524100877809\\
% 25	0.972980529602076\\
% 26	0.971819384755969\\
% 27	0.971128041087735\\
% 28	0.969108860134968\\
% 29	0.965799586208597\\
30	0.965444633273952\\
% 31	0.964420402233953\\
% 32	0.964101826557373\\
% 33	0.961936795283551\\
% 34	0.961234999704892\\
% 35	0.958891132383914\\
% 36	0.957100956172126\\
% 37	0.955544059179124\\
% 38	0.954106703800114\\
% 39	0.952584616002578\\
40	0.952156587637215\\
% 41	0.950283185312707\\
% 42	0.950116672774101\\
% 43	0.947138729415931\\
% 44	0.946466390736084\\
% 45	0.945104521583456\\
% 46	0.942930280235084\\
% 47	0.94280630350002\\
% 48	0.941248556729857\\
% 49	0.941092468698569\\
50	0.93724692308258\\
% 51	0.936894567902195\\
% 52	0.935252879835697\\
% 53	0.934087107888399\\
% 54	0.933245379639883\\
% 55	0.931683941876023\\
% 56	0.929851075427423\\
% 57	0.92911195685521\\
% 58	0.926376251923706\\
% 59	0.925842131959638\\
60	0.924985397575914\\
% 61	0.92265475369649\\
% 62	0.922200122943938\\
% 63	0.919419999202877\\
% 64	0.918935738061092\\
% 65	0.917485648431567\\
% 66	0.917121064694628\\
% 67	0.914249934138618\\
% 68	0.914030984702923\\
% 69	0.913814094443601\\
70	0.913010739036632\\
% 71	0.911956127149331\\
% 72	0.910553074563931\\
% 73	0.907750263702215\\
% 74	0.907263192531631\\
% 75	0.904908768637692\\
% 76	0.904381820749697\\
% 77	0.902286039006557\\
% 78	0.900670638021454\\
% 79	0.900412143063596\\
80	0.899328830232103\\
% 81	0.897938216219436\\
% 82	0.896715878362612\\
% 83	0.89647956833094\\
% 84	0.895564916207808\\
% 85	0.895078186644028\\
% 86	0.894193024710257\\
% 87	0.891705138153392\\
% 88	0.891037309173956\\
% 89	0.890086753922689\\
90	0.888580309908242\\
% 91	0.888144070130954\\
% 92	0.887553177733267\\
% 93	0.886480305704272\\
% 94	0.882874247322359\\
% 95	0.882709901914988\\
% 96	0.882300599536229\\
% 97	0.881820275466232\\
% 98	0.880185648455166\\
% 99	0.878626484773627\\
100	0.877563534126559\\
% 101	0.87537391164641\\
% 102	0.874516273348399\\
% 103	0.873598434071829\\
% 104	0.87283355081699\\
% 105	0.872468662700056\\
% 106	0.871961560684531\\
% 107	0.87008132159916\\
% 108	0.869219277700824\\
% 109	0.866459622826153\\
% 110	0.866035337159829\\
% 111	0.865360566226033\\
% 112	0.864763306952191\\
% 113	0.86419481039771\\
% 114	0.862632766682656\\
% 115	0.862057544514841\\
% 116	0.860984470786824\\
% 117	0.860529159429023\\
% 118	0.859606312451262\\
% 119	0.858120756585391\\
% 120	0.856156060557\\
% 121	0.855479504431891\\
% 122	0.854150006884797\\
% 123	0.85406687822283\\
% 124	0.853634224169011\\
% 125	0.85342865105454\\
% 126	0.853178041515759\\
% 127	0.852046352628666\\
% 128	0.851767707222969\\
% 129	0.851412823965717\\
% 130	0.84990944627922\\
% 131	0.849680867326151\\
% 132	0.848453096554075\\
% 133	0.848223034441965\\
% 134	0.847081217995475\\
135	0.846943499652238\\
% 136	0.845499882104996\\
% 137	0.844115000964444\\
% 138	0.843409247520722\\
% 139	0.841847882418952\\
% 140	0.840462335093385\\
% 141	0.839710175850298\\
% 142	0.838236550115208\\
% 143	0.837153157257311\\
% 144	0.83476289472344\\
% 145	0.834589196010503\\
% 146	0.833252360309871\\
% 147	0.832445340481685\\
% 148	0.831985474750676\\
% 149	0.831423435948156\\
% 150	0.831385950516849\\
% 151	0.830524989756174\\
% 152	0.828645374555095\\
% 153	0.827368453499887\\
% 154	0.827096071828818\\
% 155	0.82649339547579\\
% 156	0.82454779837905\\
% 157	0.82413422412636\\
% 158	0.823656557048291\\
% 159	0.822182198358673\\
% 160	0.820406203856921\\
% 161	0.81985445314088\\
% 162	0.819010573099301\\
% 163	0.818389278150626\\
% 164	0.816908739498571\\
% 165	0.815339227323771\\
% 166	0.815167834933524\\
% 167	0.815021295153992\\
% 168	0.814410612557716\\
% 169	0.812597835349905\\
% 170	0.812353554996812\\
% 171	0.81099659151672\\
% 172	0.810627276481714\\
% 173	0.809044575375907\\
% 174	0.808380768746609\\
% 175	0.807121208580708\\
% 176	0.806208342133315\\
% 177	0.805445218007118\\
% 178	0.805076853533191\\
% 179	0.802998006887399\\
% 180	0.802540288349181\\
% 181	0.801391605137119\\
% 182	0.800721019056236\\
% 183	0.799193362136404\\
184	0.79914084031756\\
277	0.711440316678438\\
376	0.629439847384643\\
464	0.57248741218043\\
550	0.519383188003502\\
691	0.443037575080567\\
1119	0.253656604550855\\
1557	0.130703678355676\\
1927	0.066143295657849\\
2357	0.0217387786306896\\
2876	0.00163619416878737\\
3365	1.69931561663495e-11\\
% 3366	1.69920727362856e-11\\
% 3367	1.31712606594957e-11\\
% 3368	1.31665933163837e-11\\
% 3369	9.1254965016088e-12\\
% 3370	8.70983409110889e-12\\
% 3371	8.70983409110886e-12\\
% 3372	8.05256381588531e-12\\
% 3373	8.05256381588525e-12\\
% 3374	2.44525510066221e-12\\
% 3375	2.44525510066218e-12\\
% 3376	1.78897464392017e-12\\
% 3377	9.25549843203669e-13\\
% 3378	9.25541189956645e-13\\
% 3379	2.55790175295134e-13\\
% 3380	2.51916158770001e-13\\
% 3381	2.51916158769993e-13\\
% 3382	1.40526493240848e-13\\
% 3383	1.40475787160943e-13\\
% 3384	5.06025896782619e-15\\
% 3385	5.05909156549812e-15\\
% 3386	2.23746678168953e-15\\
% 3387	2.23746308981349e-15\\
% 3388	1.82314442034673e-16\\
% 3389	4.85411428877412e-17\\
% 3390	4.8539208786042e-17\\
% 3391	2.24430036041115e-17\\
% 3392	2.24429949986771e-17\\
% 3393	6.25252959591887e-20\\
% 3394	6.25252959591887e-20\\
% 3395	1.22025509983653e-20\\
% 3396	1.22025096190198e-20\\
% 3397	1.10098539156547e-22\\
% 3398	2.20702354677296e-23\\
% 3399	9.71223682935152e-28\\
3400	9.71223580466415e-28\\
};
\node[right, align=left, text=mycolor3]
at (axis cs:1870,0.101) {t = 1};
\addplot [color=mycolor4,solid,line width=2.0pt,forget plot]
  table[row sep=crcr]{%
1	0.999999999999995\\
% 2	0.996721343874066\\
% 3	0.994233967118514\\
% 4	0.990597434798322\\
% 5	0.986067073128757\\
% 6	0.985522197959278\\
% 7	0.983486721556536\\
% 8	0.975073445250088\\
% 9	0.964950809053807\\
10	0.962951065458798\\
% 11	0.959223787033611\\
% 12	0.948153063269162\\
% 13	0.946539550011805\\
% 14	0.944448152216989\\
% 15	0.934938436740253\\
% 16	0.922565114149934\\
% 17	0.919811395608602\\
% 18	0.914823259614859\\
% 19	0.914536506982013\\
20	0.89793031300616\\
% 21	0.894295034474119\\
% 22	0.892915756033421\\
% 23	0.889163205576378\\
% 24	0.878947470116486\\
% 25	0.87200856016877\\
% 26	0.86681773457857\\
% 27	0.863738886086605\\
% 28	0.854796668280858\\
% 29	0.840301376507527\\
30	0.838758363420275\\
% 31	0.834318639641885\\
% 32	0.832941553031129\\
% 33	0.823631004685979\\
% 34	0.820630922933866\\
% 35	0.810674499167904\\
% 36	0.803135366119617\\
% 37	0.796624362018305\\
% 38	0.790650839912589\\
% 39	0.784364297682853\\
40	0.78260367406824\\
% 41	0.774934906303159\\
% 42	0.774256207743366\\
% 43	0.762198304711725\\
% 44	0.759496861110278\\
% 45	0.754048368541622\\
% 46	0.745414629745501\\
% 47	0.744924721976386\\
% 48	0.738791034374585\\
% 49	0.738178665847886\\
50	0.723219482144515\\
% 51	0.721861042913698\\
% 52	0.715558707227756\\
% 53	0.711110170865996\\
% 54	0.707911948082412\\
% 55	0.702009599994193\\
% 56	0.695131531036705\\
% 57	0.692373194273827\\
% 58	0.682239822916407\\
% 59	0.680275297433546\\
60	0.677133630020157\\
% 61	0.66864579581118\\
% 62	0.667000069100132\\
% 63	0.657006598623451\\
% 64	0.655278184036596\\
% 65	0.650124298403654\\
% 66	0.648833616379846\\
% 67	0.638740845186005\\
% 68	0.637976366166357\\
% 69	0.637219798777515\\
70	0.634423744888394\\
% 71	0.630768109191295\\
% 72	0.62593080494343\\
% 73	0.616356412569925\\
% 74	0.614704595667874\\
% 75	0.606769836123175\\
% 76	0.605005216677202\\
% 77	0.598027545205833\\
% 78	0.592693309189981\\
% 79	0.591843274180603\\
80	0.588291509046333\\
% 81	0.583757236839017\\
% 82	0.579794778550027\\
% 83	0.579031219366529\\
% 84	0.576083396800126\\
% 85	0.574519622941543\\
% 86	0.57168446236943\\
% 87	0.563775690601694\\
% 88	0.561667694394621\\
% 89	0.558678154739104\\
90	0.553966404149577\\
% 91	0.552607916845764\\
% 92	0.550772079473142\\
% 93	0.5474512581244\\
% 94	0.536406763325884\\
% 95	0.535907693530477\\
% 96	0.534666374219083\\
% 97	0.533212597222007\\
% 98	0.528288814307028\\
% 99	0.523626299478352\\
100	0.520466573223949\\
% 101	0.514005773276144\\
% 102	0.51149274382934\\
% 103	0.508814213266895\\
% 104	0.506590636907548\\
% 105	0.505532620658105\\
% 106	0.504065182385658\\
% 107	0.498653909001208\\
% 108	0.496188561158997\\
% 109	0.488361755941087\\
% 110	0.487167227887933\\
% 111	0.485272302725129\\
% 112	0.483599973334928\\
% 113	0.482012465620872\\
% 114	0.477671965039457\\
% 115	0.476081477660087\\
% 116	0.473125756974078\\
% 117	0.471876072342788\\
% 118	0.469351251463036\\
% 119	0.465309623829465\\
% 120	0.460007250064436\\
% 121	0.458192573648905\\
% 122	0.45464324584565\\
% 123	0.454422052093599\\
% 124	0.453272209771047\\
% 125	0.452726685070069\\
% 126	0.452062358817121\\
% 127	0.449072135643739\\
% 128	0.448338313652511\\
% 129	0.447405105967871\\
% 130	0.443469014811701\\
% 131	0.442872991473255\\
% 132	0.439682514905862\\
% 133	0.439086728018037\\
% 134	0.436139339987866\\
% 135	0.435784917892887\\
% 136	0.432083574305375\\
% 137	0.428556504851244\\
% 138	0.426767945864668\\
% 139	0.422832264519243\\
140	0.419364127033061\\
% 141	0.417490964370118\\
% 142	0.413840479429265\\
% 143	0.411173008313959\\
% 144	0.405336470938959\\
% 145	0.40491493130596\\
% 146	0.40168236250383\\
% 147	0.399740943949214\\
% 148	0.398638023659622\\
% 149	0.397293363569371\\
% 150	0.397203810102362\\
% 151	0.395151397842045\\
% 152	0.390700127078852\\
% 153	0.387699096491578\\
% 154	0.387061335777373\\
% 155	0.385653198232926\\
% 156	0.381135306971258\\
% 157	0.380180421733359\\
% 158	0.37907993786794\\
% 159	0.375699266066385\\
% 160	0.371659021430208\\
% 161	0.370410935469912\\
% 162	0.368508527235288\\
% 163	0.367112907970188\\
% 164	0.363804202383987\\
% 165	0.360322753212764\\
% 166	0.35994419532157\\
% 167	0.359620782275765\\
% 168	0.35827550888253\\
% 169	0.354305835605041\\
% 170	0.353773604716611\\
% 171	0.350828724690282\\
% 172	0.350030642596556\\
% 173	0.346626890705625\\
% 174	0.345207216377223\\
% 175	0.342526200148891\\
% 176	0.34059356488671\\
% 177	0.338984653342051\\
% 178	0.338210201298259\\
% 179	0.333866109676515\\
% 180	0.332915655311001\\
% 181	0.330539939795472\\
% 182	0.329159311153812\\
% 183	0.326031333700976\\
% 184	0.325924216269148\\
% 185	0.322705293407281\\
% 186	0.321373862854818\\
% 187	0.320577544100969\\
% 188	0.319189534552698\\
% 189	0.316711918345495\\
% 190	0.314564930798057\\
% 191	0.313158090421337\\
% 192	0.311296754102905\\
% 193	0.310627667431532\\
% 194	0.308844059525722\\
% 195	0.307499731913878\\
% 196	0.304313036952892\\
% 197	0.301367217024112\\
% 198	0.300190756512584\\
% 199	0.299719779302114\\
200	0.298267613886635\\
% 201	0.295880304345337\\
% 202	0.29446852179445\\
% 203	0.289888061534808\\
% 204	0.288695905243555\\
% 205	0.287902364422708\\
% 206	0.286658309016313\\
% 207	0.28385827286928\\
% 208	0.28236957209465\\
% 209	0.280205048742631\\
% 210	0.278044314115206\\
% 211	0.276582415529785\\
% 212	0.27545843690721\\
% 213	0.273422136038019\\
% 214	0.271252120630375\\
% 215	0.270124840437143\\
% 216	0.268617971767874\\
% 217	0.266834474459871\\
% 218	0.263783262253333\\
% 219	0.262680381556866\\
% 220	0.262315643677293\\
% 221	0.260063407955132\\
% 222	0.259396709221665\\
% 223	0.257756731437379\\
% 224	0.255582800208866\\
% 225	0.252880006113412\\
% 226	0.251302564067477\\
% 227	0.247795813347707\\
% 228	0.247247367492748\\
% 229	0.245020581087777\\
% 230	0.244175490314986\\
% 231	0.243867265046863\\
% 232	0.239849154909692\\
% 233	0.239051491517567\\
% 234	0.238310527071581\\
% 235	0.237147012003538\\
% 236	0.23379220647618\\
% 237	0.232313129550273\\
% 238	0.22916274799935\\
% 239	0.228830934863626\\
% 240	0.225821916409507\\
% 241	0.224879065884007\\
% 242	0.224537875334678\\
% 243	0.222877389337817\\
% 244	0.222068827850987\\
% 245	0.220026469167522\\
% 246	0.218659333688008\\
% 247	0.218050168232656\\
% 248	0.217338409154983\\
% 249	0.216276708401782\\
% 250	0.215615073029674\\
% 251	0.213457600978313\\
% 252	0.212350670646879\\
253	0.210307081033341\\
335	0.129417301113019\\
431	0.0736395426579758\\
593	0.0299399205393801\\
735	0.0130006488761223\\
987	0.00266498240884976\\
1314	0.000272696339478743\\
1692	1.18523886565873e-05\\
2147	9.39465738436037e-08\\
3400	8.64163841970062e-136\\
};
\node[right, align=left, text=mycolor4]
at (axis cs:500,0.1) {t = 5};
\end{axis}
\end{tikzpicture}%

%% file: figures/run_times.tikz
\definecolor{mycolor1}{rgb}{0.00000,0.44700,0.74100}%
\definecolor{mycolor2}{rgb}{0.85000,0.32500,0.09800}%
\begin{tikzpicture}

\begin{axis}[%
width=.7\linewidth,
height=.3\linewidth,
at={(1.011in,0.642in)},
scale only axis,
xmin=0.5,
xmax=3.5,
grid,
xtick={1,2,3},
xticklabels={{3400},{4344},{6890}},
xlabel={\footnotesize{number of vertices}},
ymin=0,
ymax=700,
grid,
ylabel={\footnotesize{time in seconds}},
axis background/.style={fill=white},
legend style={at={(0.07,0.55)},anchor=south west,legend cell align=left,align=left,draw=white!15!black, font=\footnotesize},
every x tick label/.append style={font=\color{black}, font=\footnotesize},
every y tick label/.append style={font=\color{black}, font=\footnotesize}
]
\addplot[ybar,bar width=12,bar shift=-7,draw=none,fill=mycolor1,area legend] plot table[row sep=crcr] {%
1	30.18666667\\
2	60.7\\
3	106.2694038\\
};
\addlegendentry{Heat kernels};

\addplot[ybar,bar width=12,bar shift=7,draw=none,fill=mycolor2,area legend] plot table[row sep=crcr] {%
1	98.37\\
2	192.72\\
3	625.502\\
};
\addlegendentry{Gaussian kernels};

\end{axis}
\end{tikzpicture}%

%% file: 03_method.tex
\section{Method}
\label{sec:algo}

\subsection{Optimization}

\noindent We aim at maximizing $E(\PiM)$ over $\Pn$, which by Corollary~\ref{cor:cor1} is equivalent to the relaxed problem
\begin{align}
 \argmax_{\mathbf P\in \Dn} E(\mathbf P)=\argmax_{\mathbf P \in \Dn}  \langle  {\mathbf P},\alpha\mathbf F_\Yy \mathbf F_\Xx^\top+\mathbf K_\Yy {\mathbf P}\mathbf  K_\Xx \rangle
\end{align}
where $\mathbf F_\Xx, \mathbf F_\Yy $ are matrices of pointwise descriptors and $\mathbf K_\Xx,\mathbf K_\Yy$ are the positive-definite heat kernel matrices on $\Xx$ and $\Yy$, respectively.
This maximization problem can be seen as the minimization of the difference of convex functions:
\begin{align}
 \argmin_{\mathbf P\in \Rr^{n\times n}} B(\mathbf P)-E(\mathbf P).
\end{align}
where $B$ is the (convex) indicator function on the set of bistochastic matrices $\Dn$.

\noindent A renowned way to optimize this type of energy is the difference of convex functions (DC) algorithm that starts with some initial $\mathbf P^0$ and then iterates the following two steps until convergence:
\begin{itemize}
\item Select $\mathbf Q^k \in \partial E(\mathbf P^k)$.
\item Select $\mathbf P^{k+1} \in \partial B^*(\mathbf Q^k)$.
\end{itemize}
Here $B^*$ denotes the convex conjugate of $B$ and $\partial E,\partial B^*$ denote the subdifferentials (set of supporting hyperplanes) of $E$ and $B^*$, respectively.

\noindent For a differentiable $E$, the step of the DC algorithm assumes the form
\begin{align}
\mathbf P^{k+1} = \argmax_{\mathbf P\in \Dn} \langle \mathbf P, \nabla  E(\mathbf P^k) \rangle\,.
\end{align}
Moreover, the value of the objective is an increasing sequence, $E(\mathbf P^{k+1})>  E(\mathbf P^{k})$, and each iterate $\mathbf P^k$ is a permutation matrix. 
We provide the proof in the supplementary material. Figure \ref{fig:FrankWolfe} illustrates this iterative process.

\noindent Since $\mathbf P^k$ is guaranteed to be a permutation matrix, we henceforth use $\PiM^k$ to denote the iterates. For our choice of $E$, the gradient is given by
\begin{align}
 \nabla E  &= \alpha \mathbf F_\Yy \mathbf F_\Xx^\top +  \mathbf K_\Yy { \PiM}\mathbf  K_\Xx
\end{align}
yielding the step
\begin{align}\label{eq:PMF}
\PiM^{k+1} &= \argmax_{\PiM\in \Dn} \langle \PiM, \alpha \mathbf F_\Yy \mathbf F_\Xx^\top +  \mathbf K_\Yy { \PiM}^k\mathbf  K_\Xx \rangle\,.
\end{align}
In the experiments presented in this paper, we use the data fidelity term  $\langle \PiM,\mathbf F_\Yy \mathbf F_\Xx^\top \rangle$ mainly to initialize the process:
\begin{align}
 \PiM^0 &= \argmax_{\PiM \in \Dn} \langle \PiM,\mathbf F_\Yy \mathbf F_\Xx^\top \rangle.
\end{align}
\begin{figure}
 \includegraphics[width=0.8\linewidth]{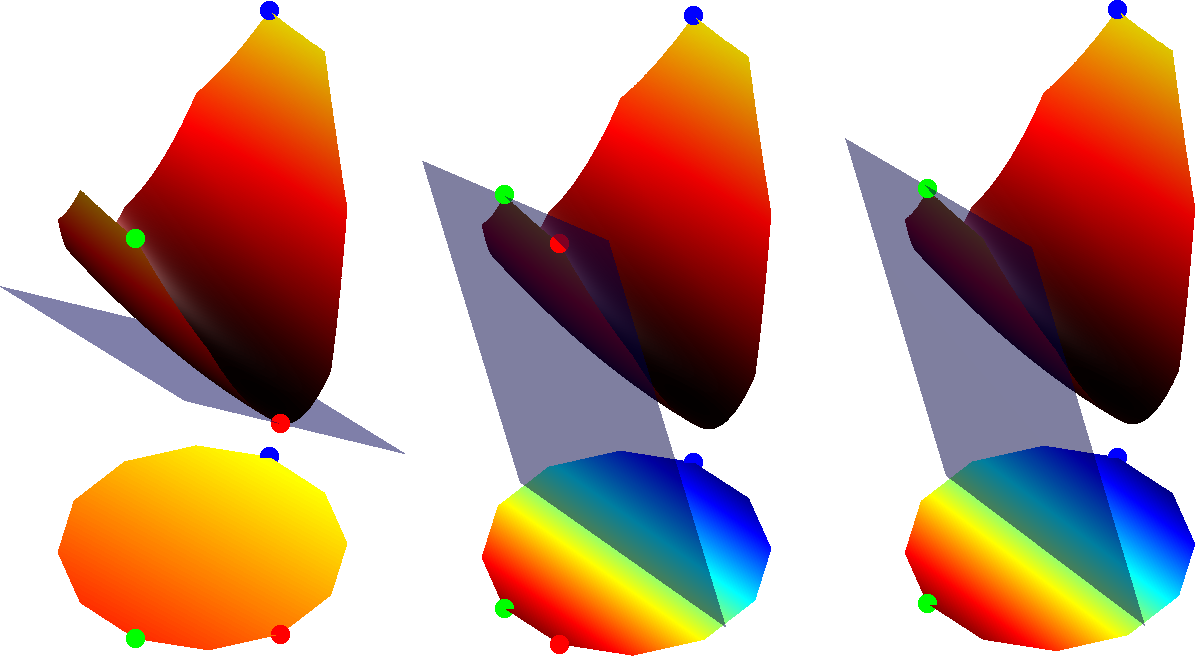}
 \centering
 \caption{Schematic illustration of the proposed algorithm for maximizing a convex quadratic objective over a convex polytope, by successively maximizing a linear sub-estimate of it. The \textcolor{red}{\textit{hot}} color map encodes the function values. The \textcolor{blue}{\textit{j}}\textcolor{green}{\textit{e}}\textcolor{red}{\textit{t}} color map encodes the values of the linear sub-estimate. The point around which the objective is linearized is depicted in red. The global maximum is depicted in blue. The maximum of the linear sub-estimate is depicted in green.  
 Notice that the algorithm travels between extreme but not necessarily adjacent points of the polytope, until it converges to a local maximum. \label{fig:FrankWolfe}}
\end{figure}

%--------------------------------------------------------------------------------------------------------------------

\subsection{Partial matching using slack variables}\label{sec:partialitySec}

\begin{figure*}
\centering
\begin{overpic}[width=\linewidth]{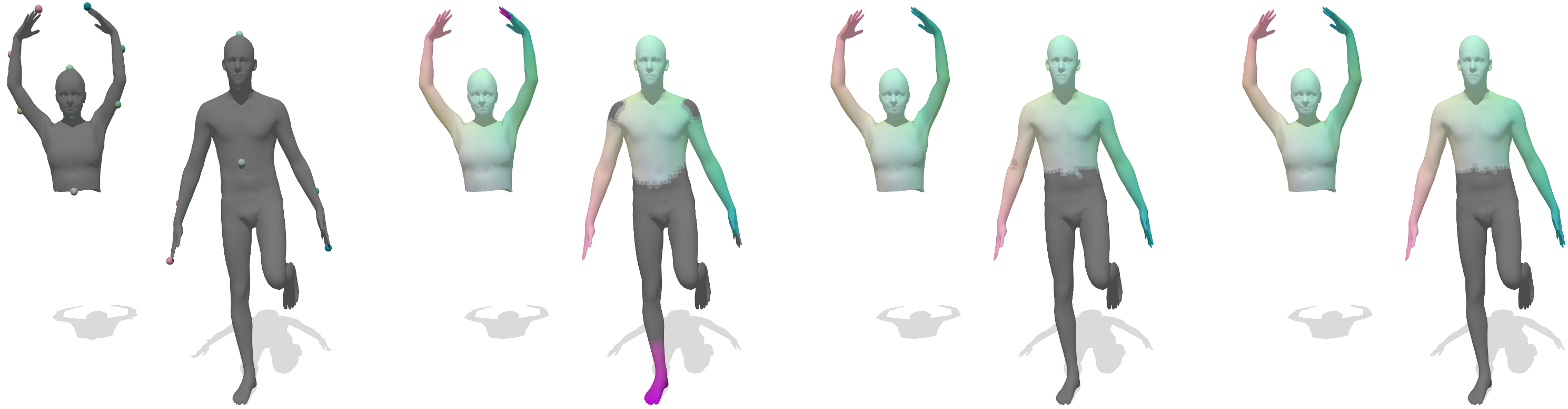}
\put(8,0){Input}
\put(34,0){Iter 1}
\put(60,0){Iter 2}
\put(87,0){Iter 3}
\end{overpic}
\caption{Our approach can tackle the challenging scenario of partial correspondences. As a proof of concept we initialized our method with sparse correspondences, indicated by spheres. We simulated noise by mapping a point on the left hand of the woman to the right foot of the man. At the first iteration all points spread their information, leading to a discontinuity of the mapping at the hand of the woman. After three iterations the method converged to the correct solution. This example was generated with Gaussian kernels. The proper choice of boundary conditions when using heat kernels will be discussed in future work.}
\label{fig:partiality}
\end{figure*}
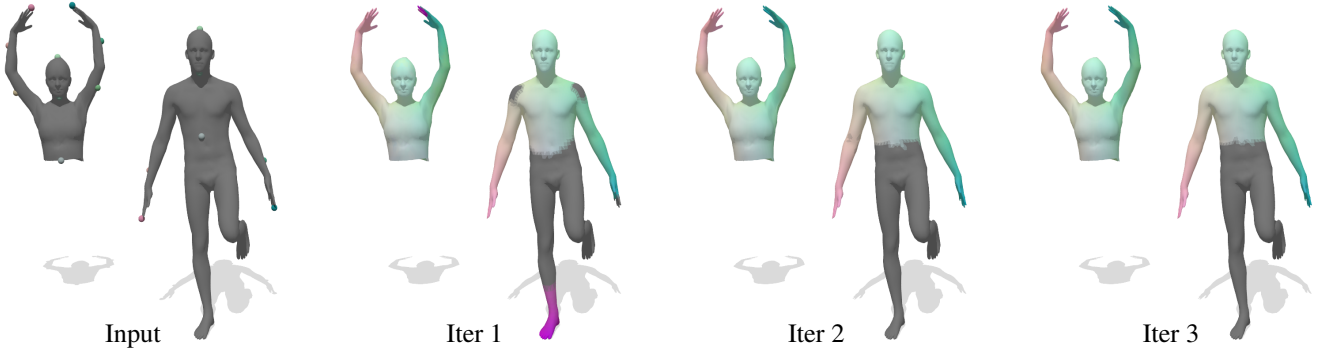

\noindent In a general setting, we will be dealing with shapes having different number of vertices. 
Let us denote by $n_\Xx$ the number of vertices on $\Xx$ and by $n_\Yy$ the number of vertices on $\Yy$, and assume w.l.o.g. $n_\Xx\geq n_\Yy$. We aim at optimizing
\begin{align}
 \argmax_{ \PiM\in \Pnxny} \langle { \PiM}, \alpha \mathbf F_\Yy \mathbf F_\Xx^\top  + \mathbf K_\Yy { \PiM}\mathbf  K_\Xx \rangle
\end{align}
where the space of \emph{rectangular permutation matrices} $\Pnxny$ is given by
$\Pnxny = \{ \PiM \in \{0,1\}^{n_\Yy\times n_\Xx}: \PiM\One \leq \One, \PiM^\top \One = \One \}$.
Analogously to the previously discussed case in which we had $n_\Xx=n_\Yy=n$, we iteratively solve 
\begin{align}
 \PiM^{k+1} &= \argmax_{ \PiM\in \Pnxny} \langle  \PiM, \alpha \mathbf F_\Yy \mathbf F_\Xx^\top  + \mathbf{K}_\Yy { \PiM^{k}}\mathbf  K_\Xx \rangle\,.
\label{eq:partial_step}
\end{align}

\noindent In order to solve these optimization problems we pad the rectangular matrix $ \alpha \mathbf F_\Yy \mathbf F_\Xx^\top  + \mathbf K_\Yy {\mathbf \Pi^{k}}\mathbf  K_\Xx$ with constant values $c$ (slack variables) such that it becomes square. After the correspondence is computed, we discard the ones belonging to the introduced slack variables. While such a treatment does not affect the value of the maximum, the constant $c$ has to be chosen appropriately to avoid ambiguity between the slacks and the actual vertices on $\Xx$. A drawback of this approach is that there are $(n_\Xx-n_\Yy)!$ solutions achieving the optimal score, leading to worse runtime in the presence of many slacks. See Fig.\ref{fig:partiality}. for a proof of concept of this approach.
%--------------------------------------------------------------------------------------------------------------------

\subsection{Multiscale acceleration}
\noindent Solving the LAP (\ref{eq:partial_step}) at each iteration of the DC algorithm has a super-quadratic complexity. As a consequence, the proposed method is only directly applicable for small $n$ (up to $15 \times 10^3$ in our experiments).
We therefore propose a multiscale approach that enables us to find correspondences between larger meshes.

\begin{figure}
  \centering
\begin{overpic}
[trim=0cm 0cm 0cm 0cm,clip,width=0.48\linewidth]{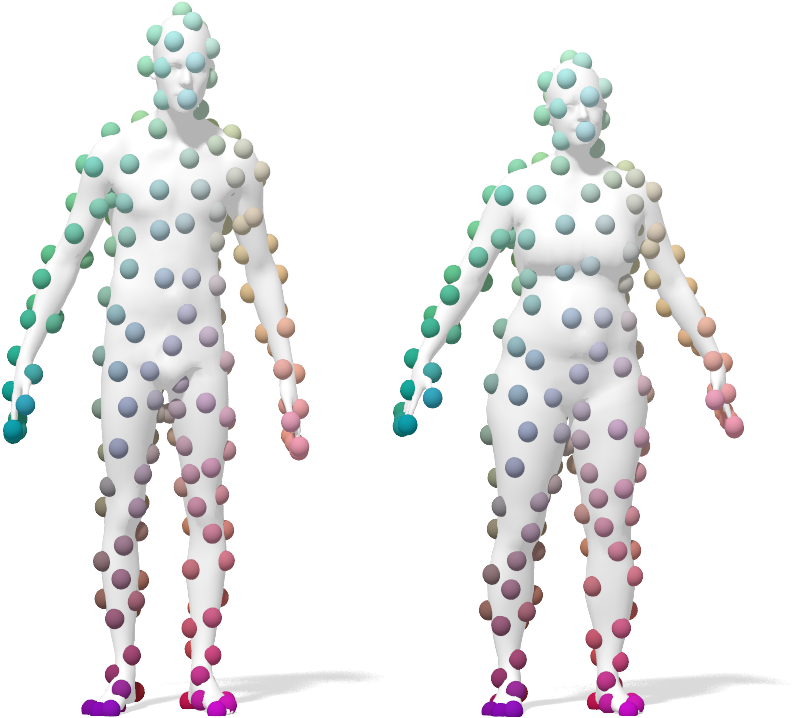}
\put(45,80){\footnotesize (a)}
\end{overpic}
\begin{overpic}
[trim=0cm 0cm 0cm 0cm,clip,width=0.48\linewidth]{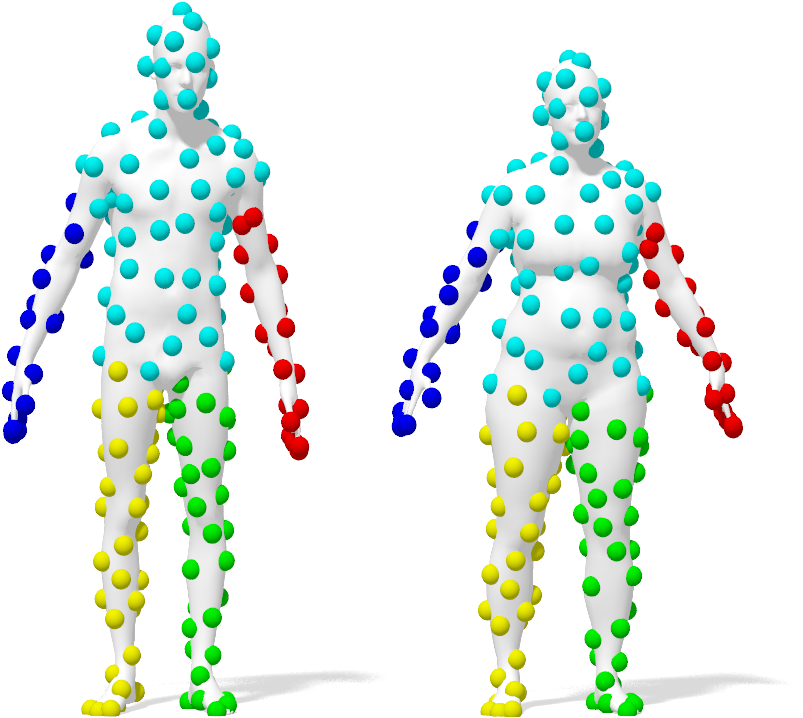}
\put(45,80){\footnotesize (b)}
\end{overpic}
\begin{overpic}
[trim=0cm 0cm 0cm 0cm,clip,width=0.48\linewidth]{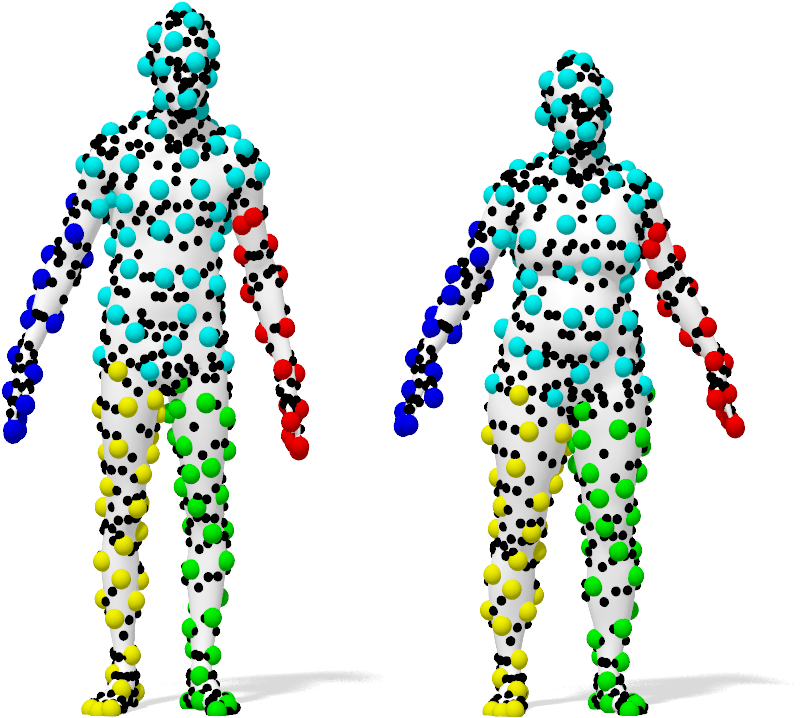}
\put(45,80){\footnotesize (c)}
\end{overpic}
\begin{overpic}
[trim=0cm 0cm 0cm 0cm,clip,width=0.48\linewidth]{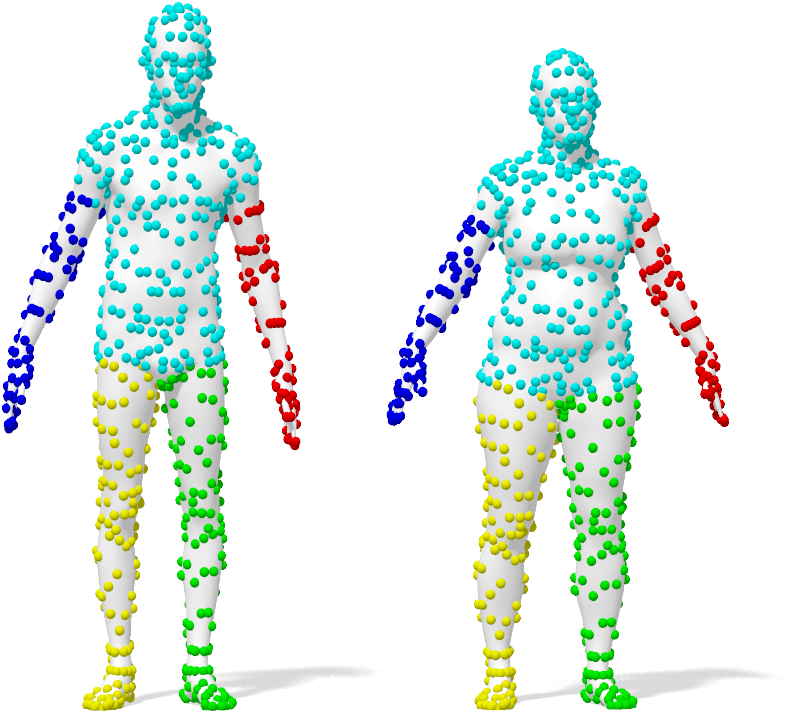}
\put(45,80){\footnotesize (d)}
\end{overpic}
    \caption{\textbf{Conceptual illustration of our multiscale approach:} (a) correspondence at a coarse scale is given; (b) vertices on the source shape are grouped into sets (left), and the known correspondence is used to group vertices on the target shape (right); (c) vertices at a finer scale are added and (d) included in the group they reside in; finally, a correspondence is calculated for each group separately.}
\label{fig:multiscale}
\end{figure}

\noindent We start by resampling both shapes to a number of vertices we can handle and solving for a bijection $\pi_0:\mathcal{X}_0\to \mathcal{Y}_0$. This set of initial vertices is called \emph{seeds}. The seeds on $\mathcal{X}$ are clustered into $k$ Voronoi cells and these cells are transfered to $\mathcal{Y}$ via $\pi_0$. More points are added iteratively and assigned to the same Voronoi cell as their closest seed. Next, we solve for $\pi_i: \mathcal{X}_i\to \mathcal{Y}_i$ where $i$ refers to the $i$-th Voronoi cell. This proceeds until all points are sampled (see Figure \ref{fig:multiscale} for a visualization). To keep the correspondence consistent at the boundary of the Voronoi cells, we choose $1000$ correspondences from $\pi_0$ and use them to orient each Voronoi cell correctly over all iterations. Additional details are provided in the supplementary material.

%% file: 04_interpretation.tex
\section{Interpretation}
\noindent In what follows we provide different, yet complementary interpretations of the proposed method, shedding light on its effectiveness. 
%--------------------------------------------------------------------------------------------------------------------

\subsection{Alternating diffusion}

%\ronslos{We must cite \cite{lederman2015learning} in this section and discuss the relation to our method. Additionally, we should point out that this works based on the assumption that one has some noisy estimation of the correct mapping (such as NN in descriptor space). }

\noindent To intuitively understand the efficacy of kernel alignment for the purpose of finding correspondences, consider the $k$-th iteration (without data term):
\begin{align}
\max_{\PiM\in \Pn}\langle \PiM, \bb{K}_{\mathcal{Y}} \PiM^k \bb{K}_{\mathcal{X}}\rangle\,.
\end{align}

\noindent Let us denote by $\boldsymbol{\delta}^j$ the discrete indicator function of vertex $j$ on shape $\Xx$, representing initial heat distribution concentrated at vertex $j$.
\noindent This heat is propagated via the application of the heat kernel $\bb{K}_{\mathcal{X}}$ to the rest of the vertices, resulting in the new heat distribution on $\Xx$ given by $\bb{k}_{\mathcal{X}}^j = \bb{K}_{\mathcal{X}}{\boldsymbol{\delta}}^j$. This heat distribution, whose spread depends on the time parameter $t$, is mapped via $\PiM^k$ onto the shape $\Yy$, where it is propagated via the heat kernel $\bb{K}_{\mathcal{Y}}$. The $ij$-th element of the matrix $\bb{K}_{\Yy}\PiM^k \bb{K}_{\mathcal{X}}$,
\begin{align}
(\bb{K}_{\Yy}\PiM^k \bb{K}_{\mathcal{X}})_{ij} &=  (\bb{k}_{\mathcal{Y}}^i)^\top \PiM^k \bb{k}_{\mathcal{X}}^j \nonumber\\
&= \sum_{m} (\bb{K}_{\mathcal{Y}})_{i,\pi^k(m)} (\bb{K}_{\mathcal{X}})_{jm},
\end{align}
represents the probability of a point $i$ on $\Yy$ being in correspondence with the point $j$ on $\Xx$. 
This is affected by both the distance between $i$ and $\pi^k(m)$ on $\Yy$ for every $m$ on $\Xx$, encoded in the entries of $(\bb{K}_{\mathcal{Y}})_{i,\pi^k(m)}$, and by the distance between $m$ and $j$ on $\Xx$, encoded in the entries of $(\bb{K}_\Xx)_{jm}$.

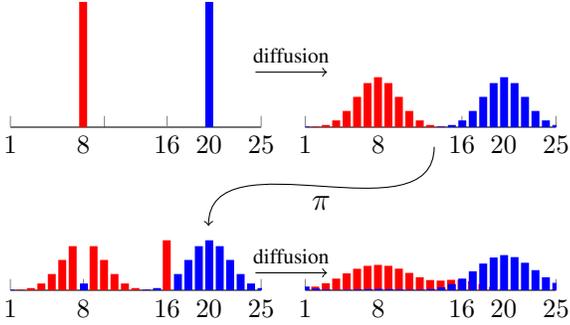
\begin{figure}
\input{figures/diffusion1.tikz}
\input{figures/diffusion2.tikz} 
\input{figures/arrow1.tikz}\\
\input{figures/diffusion3.tikz}
\input{figures/diffusion4.tikz}
\input{figures/arrow2.tikz}
 \caption{\textbf{Illustration of the alternating diffusion process} initialized with a noisy correspondence that wrongly maps $\pi(8)=16$ and $\pi(16)=8$ but correctly 
 maps $\pi(x)=x$ elsewhere. Top left: Indicator functions on the source shape, one on a point with a wrong correspondence (red) and one with a correct correspondence (blue). Top right: Both indicator functions are diffused. Bottom left: The diffused functions are transported to the target shape via $\pi$.}
 %Notice how the red maximum is in the wrong position.
Bottom right: Diffusion on the target shape.
%Now the maximum of the previously wrong matched point (red) is at the correct position.
 \label{fig:AD}
\end{figure}

\noindent This process, as illustrated in Figure \ref{fig:AD}, resembles the alternating diffusion process described in \cite{lederman2015learning}. Its success in uncovering the latent correspondence is based on the following statistical assumptions on the distribution of correspondences in the initial assignment: we tacitly assume that a sufficiently large number of (uniformly distributed) points are initially mapped correctly while the rest are mapped randomly, such that when averaging over their ``votes'' they do not bias towards any particular candidate. These concepts will be presented more rigorously in a longer version of this paper.

\noindent There is an inherent trade-off between the stability of the process and its accuracy, controlled by the time parameter $t$. Smaller $t$ enables more accurate correspondence, but limits the ability of far away points to compensate for local inaccuracies in the initial correspondence, while larger $t$ allows information to propagate from farther away, but introduces ambiguity at the fine scale. Examining the extremities, when $t\rightarrow 0$ each point is discouraged to change its initial match, while as $t\rightarrow \infty$ every point becomes a likely candidate for a match. In practice, we approximately solve a series of problems parametrized by a decreasing sequence of $t$ values, as explained in the experimental section.
%\ref{sec:Implementation}.

% If we bring this statement, then in the heat kernel section and not in the interpretation

%It should be highlighted that two manifolds $\Xx$ and $\Yy$ are isometric if and only if there exists a surjective map $\varphi:\Xx \rightarrow \Yy$ such that ${k_{\Xx}}_t(x,x')={k_{\Yy}}_t(\varphi(x), \varphi(x'))\;\; \forall x,x' \in X, t\in \mathbb{R}_+.$ \cite{memoli2011spectral}.

%--------------------------------------------------------------------------------------------------------------------

\subsection{Iterated blurring and sharpening}
\noindent An alternative point of view is to recall that
a diffusion process corresponds to a smoothing operation, or low-pass filtering in the spectral domain. To that end we view each iteration \eqref{eq:PMF} as an application of a series of low-pass filters (smoothing) followed by a projection operation (deblurring/sharpening). To see that, we use the spectral decomposition of the heat kernels to rewrite the payoff matrix in \eqref{eq:PMF}
\begin{align}\label{eq:KMfilter}
\bb{K}_\Yy\PiM\bb{K}_\Xx & = \bb{\Psi}e^{t\bb{\Lambda}_\Yy}\bb{\Psi}^\top\PiM\bb{\Phi}e^{t\bb{\Lambda}_\Xx}\bb{\Phi}^\top\nonumber\\ 
&= \bb{\Psi}e^{t\bb{\Lambda}_\Yy}\bb{C}e^{t\bb{\Lambda}_\Xx}\bb{\Phi}^\top.
\end{align}
where the functional map $\mathbf{C}$ is seen as a low-pass approximation of the permutation matrix in the truncated Laplacian eigenbasis, $\boldsymbol{\Pi} \approx \bb{\Psi}\mathbf{C}\bb{\Phi}^\top$. 
Equation \eqref{eq:KMfilter} can thus be interpreted as applying a low-pass filter to the functional map matrix $\bb{C}$.
The second step in \eqref{eq:PMF} can be regarded as a projection of the smoothed correspondence on the set of permutations \eqref{eq:proj}, producing a point-wise bijection.

\subsection{Kernel density estimation in the product space}

\noindent Similar to the interpretation in \cite{PMF}, our approach can be seen as estimating the graph
$\Pi = \{(x,\pi(x)):x\in \Xx \}$ of the latent correspondence  $\pi: \mathcal{X} \to \mathcal{Y}$ on the product manifold $\mathcal{X} \times \mathcal{Y}$. In case of a bijective, continuous $\pi$, the graph $\Pi$ is a submanifold without a boundary of same dimension as $\mathcal{X}$ (2 in the discussed case).
In each iteration of the process a probability distribution $P:\Xx\times \Yy\rightarrow [0,1]$ is constructed by placing kernels (geodesic Gaussian kernels in \cite{PMF}, and heat kernels in our case) on the graph of the previous iterate and maximizing
\begin{align}
 \hat{\pi} = \mathrm{arg} \max_{ \hat{\pi} : \Xx \overset{1:1}{\rightarrow} \Yy  } \int_{\Xx}  P(x,\hat{\pi}(x)) dx
\end{align}
over the set of bijective but not necessarily continuous correspondences.

%% file: figures/diffusion1.tikz
\definecolor{mycolor1}{rgb}{0.24220,0.15040,0.66030}%
\definecolor{mycolor2}{rgb}{0.97690,0.98390,0.08050}%
\begin{tikzpicture}

\begin{axis}[%
width=.4\linewidth,
height=.2\linewidth,
at={(1.236in,0.481in)},
scale only axis,
xmin=1,
xmax=25,
ymin=0,
ymax=0.7,
axis background/.style={fill=white},
hide y axis,
axis x line*=bottom,
xticklabels={},
extra x ticks={1,8,16,20,25},
extra x tick style={xticklabel=\pgfmathprintnumber{\tick}}
]
\addplot[ybar,bar width=3,bar shift=-0.143,draw=none,fill=red,area legend] plot table[row sep=crcr] {%
1	0\\
2	0\\
3	0\\
4	0\\
5	0\\
6	0\\
7	0\\
8	0.7\\
9	0\\
10	0\\
11	0\\
12	0\\
13	0\\
14	0\\
15	0\\
16	0\\
17	0\\
18	0\\
19	0\\
20	0\\
21	0\\
22	0\\
23	0\\
24	0\\
25	0\\
};
\addplot[ybar,bar width=3,bar shift=0.143,draw=none,fill=blue,area legend] plot table[row sep=crcr] {%
1	0\\
2	0\\
3	0\\
4	0\\
5	0\\
6	0\\
7	0\\
8	0\\
9	0\\
10	0\\
11	0\\
12	0\\
13	0\\
14	0\\
15	0\\
16	0\\
17	0\\
18	0\\
19	0\\
20	0.7\\
21	0\\
22	0\\
23	0\\
24	0\\
25	0\\
};
\end{axis}
\end{tikzpicture}%

%% file: figures/diffusion2.tikz
\definecolor{mycolor1}{rgb}{0.24220,0.15040,0.66030}%
\definecolor{mycolor2}{rgb}{0.97690,0.98390,0.08050}%
\begin{tikzpicture}

\begin{axis}[%
width=.4\linewidth,
height=.2\linewidth,
at={(1.236in,0.481in)},
scale only axis,
xmin=1,
xmax=25,
ymin=0,
ymax=0.5,
axis background/.style={fill=white},
hide y axis,
axis x line*=bottom,
xticklabels={},
extra x ticks={1,8,16,20,25},
extra x tick style={xticklabel=\pgfmathprintnumber{\tick}}
]
\addplot[ybar,bar width=3,bar shift=-0.143,draw=none,fill=red,area legend] plot table[row sep=crcr] {%
1	0.000436341347643458\\
2	0.00221592420658135\\
3	0.00876415024920614\\
4	0.0269954832640539\\
5	0.0647587978508412\\
6	0.120985362293005\\
7	0.176032663430794\\
8	0.199471140255838\\
9	0.176032663430794\\
10	0.120985362293005\\
11	0.0647587978508412\\
12	0.0269954832640539\\
13	0.00876415024920614\\
14	0.00221592420658135\\
15	0.000436341347643458\\
16	6.69151129009339e-05\\
17	7.9918705556612e-06\\
18	7.43359757572568e-07\\
19	5.38488002275969e-08\\
20	3.03794142575114e-09\\
21	3.03794142575114e-09\\
22	5.38488002275969e-08\\
23	7.43359757572568e-07\\
24	7.9918705556612e-06\\
25	6.69151129009339e-05\\
};
\addplot[ybar,bar width=3,bar shift=0.143,draw=none,fill=blue,area legend] plot table[row sep=crcr] {%
1	0.00221592420658135\\
2	0.000436341347643458\\
3	6.69151129009339e-05\\
4	7.9918705556612e-06\\
5	7.43359757572568e-07\\
6	5.38488002275969e-08\\
7	3.03794142575114e-09\\
8	3.03794142575114e-09\\
9	5.38488002275969e-08\\
10	7.43359757572568e-07\\
11	7.9918705556612e-06\\
12	6.69151129009339e-05\\
13	0.000436341347643458\\
14	0.00221592420658135\\
15	0.00876415024920614\\
16	0.0269954832640539\\
17	0.0647587978508412\\
18	0.120985362293005\\
19	0.176032663430794\\
20	0.199471140255838\\
21	0.176032663430794\\
22	0.120985362293005\\
23	0.0647587978508412\\
24	0.0269954832640539\\
25	0.00876415024920614\\
};
\end{axis}
\end{tikzpicture}%

%% file: figures/arrow1.tikz
\begin{tikzpicture}[remember picture, overlay]
% nodes
\node (A) at (-4.5,1.2) {};
\node (B) at (-3.3,1.2) {};
% arrows
\draw[->]%, to path={-| (\tikztotarget)}]
  (A) edge node[above,midway,above]  {\footnotesize diffusion} (B);

\end{tikzpicture}

%% file: figures/diffusion3.tikz
\definecolor{mycolor1}{rgb}{0.24220,0.15040,0.66030}%
\definecolor{mycolor2}{rgb}{0.97690,0.98390,0.08050}%
\begin{tikzpicture}

\begin{axis}[%
width=.4\linewidth,
height=.2\linewidth,
at={(1.236in,0.481in)},
scale only axis,
xmin=1,
xmax=25,
ymin=0,
ymax=0.5,
axis background/.style={fill=white},
hide y axis,
axis x line*=bottom,
xticklabels={},
extra x ticks={1,8,16,20,25},
extra x tick style={xticklabel=\pgfmathprintnumber{\tick}}
]
\addplot[ybar,bar width=3,bar shift=-0.143,draw=none,fill=red,area legend] plot table[row sep=crcr] {%
1	0.000436341347643458\\
2	0.00221592420658135\\
3	0.00876415024920614\\
4	0.0269954832640539\\
5	0.0647587978508412\\
6	0.120985362293005\\
7	0.176032663430794\\
8	6.69151129009339e-05\\
9	0.176032663430794\\
10	0.120985362293005\\
11	0.0647587978508412\\
12	0.0269954832640539\\
13	0.00876415024920614\\
14	0.00221592420658135\\
15	0.000436341347643458\\
16	0.199471140255838\\
17	7.9918705556612e-06\\
18	7.43359757572568e-07\\
19	5.38488002275969e-08\\
20	3.03794142575114e-09\\
21	3.03794142575114e-09\\
22	5.38488002275969e-08\\
23	7.43359757572568e-07\\
24	7.9918705556612e-06\\
25	6.69151129009339e-05\\
};
\addplot[ybar,bar width=3,bar shift=0.143,draw=none,fill=blue,area legend] plot table[row sep=crcr] {%
1	0.00221592420658135\\
2	0.000436341347643458\\
3	6.69151129009339e-05\\
4	7.9918705556612e-06\\
5	7.43359757572568e-07\\
6	5.38488002275969e-08\\
7	3.03794142575114e-09\\
8	0.0269954832640539\\
9	5.38488002275969e-08\\
10	7.43359757572568e-07\\
11	7.9918705556612e-06\\
12	6.69151129009339e-05\\
13	0.000436341347643458\\
14	0.00221592420658135\\
15	0.00876415024920614\\
16	3.03794142575114e-09\\
17	0.0647587978508412\\
18	0.120985362293005\\
19	0.176032663430794\\
20	0.199471140255838\\
21	0.176032663430794\\
22	0.120985362293005\\
23	0.0647587978508412\\
24	0.0269954832640539\\
25	0.00876415024920614\\
};
\end{axis}
\end{tikzpicture}%

%% file: figures/diffusion4.tikz
\definecolor{mycolor1}{rgb}{0.24220,0.15040,0.66030}%
\definecolor{mycolor2}{rgb}{0.97690,0.98390,0.08050}%
\begin{tikzpicture}

\begin{axis}[%
width=.4\linewidth,
height=.2\linewidth,
at={(1.236in,0.481in)},
scale only axis,
xmin=1,
xmax=25,
ymin=0,
ymax=0.5,
axis background/.style={fill=white},
hide y axis,
axis x line*=bottom,
xticklabels={},
extra x ticks={1,8,16,20,25},
extra x tick style={xticklabel=\pgfmathprintnumber{\tick}}
]
\addplot[ybar,bar width=3,bar shift=-0.143,draw=none,fill=red,area legend] plot table[row sep=crcr] {%
1	0.00651001463888159\\
2	0.0144244322710817\\
3	0.0278175323403524\\
4	0.046505424389529\\
5	0.067453216524868\\
6	0.0857229782350437\\
7	0.0974017030067357\\
8	0.101285350960038\\
9	0.0974871177023047\\
10	0.0861646946553561\\
11	0.069200814376669\\
12	0.0518884372063784\\
13	0.0407307096412091\\
14	0.0385494139550775\\
15	0.0416115232610701\\
16	0.0423454201833551\\
17	0.0359928689638244\\
18	0.0243972398490745\\
19	0.0129871266714772\\
20	0.00540406797352564\\
21	0.00176866314037708\\
22	0.000515813414230925\\
23	0.000359255735725085\\
24	0.000904555268797824\\
25	0.00257162563501723\\
};
\addplot[ybar,bar width=3,bar shift=0.143,draw=none,fill=blue,area legend] plot table[row sep=crcr] {%
1	0.0148780453596608\\
2	0.00665669320255691\\
3	0.00281996754124044\\
4	0.00162156167795349\\
5	0.00202058904965186\\
6	0.00333999739059661\\
7	0.00477292569708858\\
8	0.00540406797352564\\
9	0.00481426654309416\\
10	0.00347863367380541\\
11	0.00240440812741313\\
12	0.0025833751785\\
13	0.00508527231603172\\
14	0.0116600481654637\\
15	0.0248248332834795\\
16	0.046505424389529\\
17	0.0756145131534125\\
18	0.106581784539565\\
19	0.13075357285207\\
20	0.140318640012254\\
21	0.132265173882505\\
22	0.109788003943214\\
23	0.0803546245169563\\
24	0.0518868465451548\\
25	0.0295667309852769\\
};
\end{axis}
\end{tikzpicture}%

%% file: figures/arrow2.tikz
\begin{tikzpicture}[remember picture, overlay]
% nodes
\node (A) at (-4.5,0.7) {};
\node (B) at (-3.3,0.7) {};
\node (C) at (-2,2.5) {};
\node (D) at (-5,1.2) {};
% arrows
\draw[->]%, to path={-| (\tikztotarget)}]
  (A) edge node[above,midway,above]  {\footnotesize diffusion} (B);
\draw[->]%, to path={-| (\tikztotarget)}]
  (C) edge[out=-90,in=90,->] node[above,midway,below]  {\large $\pi$} (D);

\end{tikzpicture}

%% file: 05_experiments.tex
\section{Experiments}

%\noindent We performed an extensive quantitative evaluation of the proposed method on the TOSCA~\cite{bronstein2008numerical}, SCAPE~\cite{scape2005}, FAUST~\cite{Bogo:CVPR:2014} and SHREC'16 Topology~\cite{SHREC16-topology} benchmarks.
\noindent We performed an extensive quantitative evaluation of the proposed method on four different benchmarks.
All datasets include several classes of (nearly) isometric shapes, with the last one additionally introducing strong topological noise (i.e., mesh `gluing' in areas of contact). In our experiments we used the SHOT~\cite{shot} and heat kernel signature (HKS)~\cite{sun2009concise} descriptors with default parameters. For the computation of heat kernels we used $500$ Laplacian eigenfunctions. % To preclude any use of groundtruth information (such as consistent ordering of the vertices) we shuffled all shapes before running the experiments.
We provide comparisons with complete matching pipelines as well as with learning-based approaches, where we show how using our method as a post-processing step leads to a significant boost in performance. In addition Figure \ref{fig:runtimes} provides runtime comparison against \cite{PMF} which uses a similar method with geodesic Gaussian kernels. 
Code of our method is available at \url{https://github.com/zorah/KernelMatching}.

\vspace{1ex}\noindent\textbf{Error measure.}
We measure correspondence quality according to the Princeton benchmark protocol \cite{kim11}. Assume
to be given a match $(x, y) \in \mathcal{X} \times \mathcal{Y}$, whereas the
ground-truth correspondence is $(x, y^*)$. Then, we measure
the geodesic error
$\epsilon(x) = d_{\mathcal{Y}} (y, y^*)/\text{diam} (\mathcal{Y})$
normalized by the geodesic diameter of $\mathcal{Y}$. Ideal correspondence should produce $\epsilon = 0$. We plot cumulative curves showing the percentage of matches that have error smaller than a variable threshold.

\begin{figure}[!h]
  \input{figures/curve_scape.tikz}
 \caption{Correspondence accuracy on the SCAPE dataset.}% (top) and learning-based methods that were trained on the FAUST models (bottom).}
 \label{fig:curve_scape}
\end{figure}
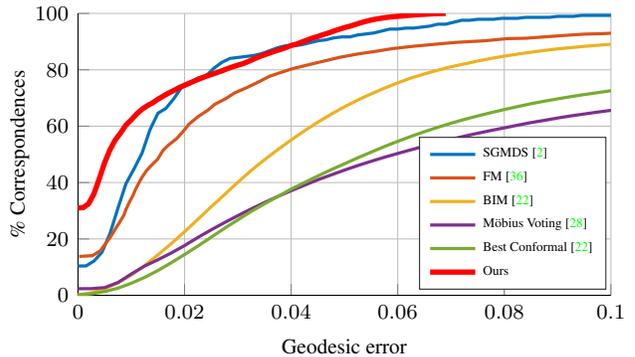

\vspace{1ex}\noindent\textbf{Parameters.}
The optimal choice of parameters does not only depend on properties of the considered shapes (such as diameterand density of the sampling) but also on the noise of the input correspondence. The exact dependencies in particular on the latter will be investigated in follow up works.
% For each dataset we fixed the parameters and report them in the respective paragraphs.

\begin{figure}[t]
 \input{figures/curve_tosca.tikz}
 \caption{Correspondence accuracy on the TOSCA dataset. }
 \label{fig:curve_tosca}
\end{figure}
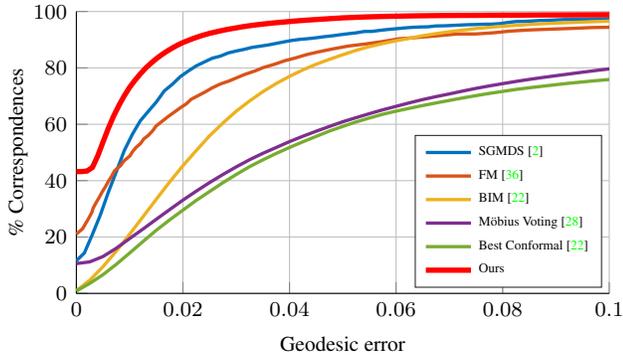

\vspace{1ex}\noindent\textbf{TOSCA.}
The TOSCA dataset \cite{bronstein2008numerical} contains 76 shapes divided into 8 classes (humans and animals) of varying resolution (3K to 50K vertices). We match each shape with one instance of the same class. For shapes having more than 10K vertices we use our multiscale acceleration with an initial problem size of 10K and a maximum problem size of 3K for all further iterations. The parameters were set to $\alpha = 10^{-10}$ and $t = [300\ 100\ 50\ 10]$, with 5 iterations per diffusion time. Figure \ref{fig:curve_tosca} shows a quantitative evaluation.

\vspace{1ex}\noindent\textbf{SCAPE.}
The SCAPE dataset \cite{scape2005} contains 72 clean shapes of scanned humans in different poses. For this test we set $\alpha = 10^{-7}$, $t=[0.1\ 0.05\ 0.009\ 0.001\ 0.0001]$, and 5 iterations per diffusion time. We used multiscale acceleration with initial size equal to 10K vertices, and equal to 1K for subsequent iterations. Quantitative and qualitative results are given in Figure~\ref{fig:curve_scape} and~\ref{fig:renderings} (right) respectively.

\begin{figure}
 \input{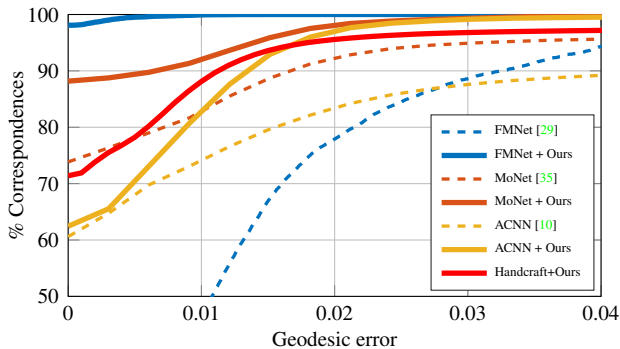}
 \vspace{-0.5cm}
 \caption{%Correspondence accuracy on FAUST (template subset). The dashed curves indicate the raw performance of recent deep learning methods, while solid curves are the results obtained after using our method as post-processing. Our method based on handcrafted descriptors (SHOT) is denoted as `Handcrafted+Ours'.
 Correspondence accuracy on FAUST. Dashed curves indicate the performance of recent deep learning methods, solid curves are obtained using our method as post-processing. Our method based on handcrafted descriptors (SHOT) is denoted as `Handcrafted+Ours'.
 }
 \label{fig:curve_faust}
\end{figure}

\vspace{1ex}\noindent\textbf{FAUST.}
The FAUST dataset \cite{Bogo:CVPR:2014} contains 100 human scans belonging to 10 different individuals; for these tests we used the template subset of FAUST, consisting of shapes with around 7K vertices each. This allowed us to run our algorithm without multiscale acceleration. We set $\alpha = 10^{-7}$ and $t=[500\ 323\ 209\ 135\ 87\ 36\ 23\ 15\ 10]$.
Differently from the previous experiments, here we employ our method as a refinement step for several deep learning-based methods, demonstrating significant improvements (up to $50\%$) upon the `raw' output of such approaches. The results are reported in Figure~\ref{fig:curve_faust}.
Our results contain a few shapes in which body parts were swapped, preventing us from reaching 100\%. An example is presented in the supp. material.

\begin{figure}[!h]
 \input{figures/curve_topkids.tikz}
 \caption{Correspondence accuracy on SHREC'16 Topology.}
 \label{fig:curve_topo}
\end{figure}
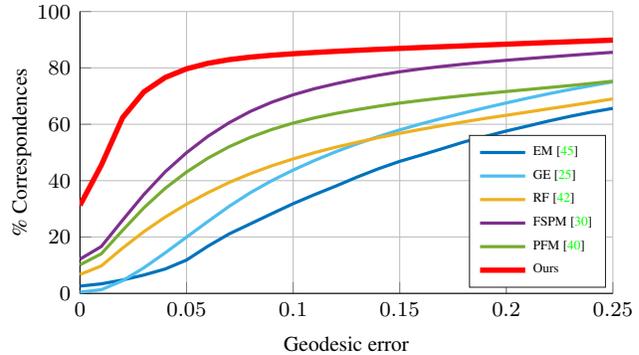

% \begin{figure}
% \centering
% 	\includegraphics[width=.25\linewidth]{figures/faust_fail1.png}
%     \includegraphics[width=.33\linewidth]{figures/faust_fail2.png}
%     \caption{A failure case of our method. Left and right are switched on the upper body, causing a non-continuous correspondence. We observed eight such failure cases in the entire FAUST dataset, preventing the red curve from attaining 100\% accuracy in Figure~\ref{fig:curve_faust}.}
%     \label{fig:faust_fail}
% \end{figure}

%--------------------------------------------------------------------------------------------------------------------

\vspace{1ex}\noindent\textbf{SHREC'16 Topology.} 
This dataset \cite{SHREC16-topology} contains 25 shapes of the same class with around 12K vertices, undergoing near-isometric deformations in addition to large topological shortcuts (see Figure~\ref{fig:renderings} middle). Here we use only SHOT as a descriptor, since HKS is not robust against topological changes. We used $\alpha = 10^{-6}$ and $t=[2.7\ 2.44\ 2.1\ 1.95\ 1.7]$, using multiscale with an initial problem of size 12k and the following problems with maximum size 1k.
Quantitative results are reported in Figure \ref{fig:curve_topo}.
% Although there is room for improvement, we clearly outperform all previous methods on this challenging benchmark.

%% file: figures/curve_scape.tikz
\definecolor{mycolor1}{rgb}{0.00000,0.44700,0.74100}%
\definecolor{mycolor2}{rgb}{0.85000,0.32500,0.09800}%
\definecolor{mycolor3}{rgb}{0.92900,0.69400,0.12500}%
\definecolor{mycolor4}{rgb}{0.49400,0.18400,0.55600}%
\definecolor{mycolor5}{rgb}{0.46600,0.67400,0.18800}%
\definecolor{mycolor6}{rgb}{0.30100,0.74500,0.93300}%
\begin{tikzpicture}

\begin{axis}[%
width=0.85\linewidth,
height=0.45\linewidth,
at={(0.8in,0.609in)},
scale only axis,
xmin=0,
xmax=0.1,
ymin=0,
ymax=100,
xmajorgrids,
ymajorgrids,
every x tick label/.append style={font=\color{black}, font=\footnotesize},
every y tick label/.append style={font=\color{black}, font=\footnotesize},
axis background/.style={fill=white},
axis x line*=bottom,
axis y line*=left,
legend style={at={(0.64,0.02)},anchor=south west,legend cell align=left,align=left,draw=white!15!black,font=\tiny},
x label style={at={(axis description cs:0.5,0.02)},anchor=north},
y label style={at={(axis description cs:0.1,.5)},rotate=0,anchor=south},
xlabel={\footnotesize Geodesic error},
ylabel={\footnotesize \% Correspondences},
x tick label style={/pgf/number format/fixed}
]

\addplot [color=mycolor1,solid,very thick]
  table[row sep=crcr]{%
0	10.4166666666667\\
0.0015	10.4166666666667\\
0.003	12.1527777777778\\
0.0045	15.2777777777778\\
0.006	21.875\\
0.0075	30.9027777777778\\
0.009	39.5833333333333\\
0.0105	44.7916666666667\\
0.012	50.3472222222222\\
0.0135	58.6805555555556\\
0.015	64.5833333333333\\
0.0165	66.3194444444445\\
0.018	69.7916666666667\\
0.0195	73.9583333333333\\
0.021	75\\
0.0225	77.0833333333333\\
0.024	77.7777777777778\\
0.0255	80.5555555555556\\
0.027	82.6388888888889\\
0.0285	84.0277777777778\\
0.03	84.375\\
0.0315	84.7222222222222\\
0.033	85.0694444444444\\
0.0345	85.7638888888889\\
0.036	86.8055555555556\\
0.0375	87.8472222222222\\
0.039	88.5416666666667\\
0.0405	88.5416666666667\\
0.042	88.8888888888889\\
0.0435	89.5833333333333\\
0.045	90.2777777777778\\
0.0465	90.625\\
0.048	91.3194444444444\\
0.0495	91.6666666666667\\
0.051	91.6666666666667\\
0.0525	92.3611111111111\\
0.054	92.7083333333333\\
0.0555	93.0555555555555\\
0.057	93.75\\
0.0585	94.4444444444444\\
0.06	94.4444444444444\\
0.0615	94.7916666666667\\
0.063	94.7916666666667\\
0.0645	95.1388888888889\\
0.066	95.4861111111111\\
0.0675	96.1805555555556\\
0.069	96.1805555555556\\
0.0705	96.875\\
0.072	97.5694444444444\\
0.0735	97.5694444444444\\
0.075	97.9166666666667\\
0.0765	97.9166666666667\\
0.078	97.9166666666667\\
0.0795	98.2638888888889\\
0.081	98.2638888888889\\
0.0825	98.2638888888889\\
0.084	98.6111111111111\\
0.0855	98.6111111111111\\
0.087	98.6111111111111\\
0.0885	98.6111111111111\\
0.09	98.9583333333333\\
0.0915	98.9583333333333\\
0.093	98.9583333333333\\
0.0945	99.3055555555556\\
0.096	99.3055555555556\\
0.0975	99.3055555555556\\
0.099	99.3055555555556\\
0.1005	99.3055555555556\\
0.102	99.3055555555556\\
0.1035	99.3055555555556\\
0.105	99.3055555555556\\
0.1065	99.3055555555556\\
0.108	99.3055555555556\\
0.1095	99.3055555555556\\
0.111	99.3055555555556\\
0.1125	99.3055555555556\\
0.114	99.3055555555556\\
0.1155	99.3055555555556\\
0.117	99.3055555555556\\
0.1185	99.3055555555556\\
0.12	99.3055555555556\\
0.1215	99.3055555555556\\
0.123	99.3055555555556\\
0.1245	99.3055555555556\\
0.126	99.3055555555556\\
0.1275	99.3055555555556\\
0.129	99.6527777777778\\
0.1305	99.6527777777778\\
0.132	99.6527777777778\\
0.1335	99.6527777777778\\
0.135	99.6527777777778\\
0.1365	99.6527777777778\\
0.138	99.6527777777778\\
0.1395	99.6527777777778\\
0.141	99.6527777777778\\
0.1425	99.6527777777778\\
0.144	99.6527777777778\\
0.1455	99.6527777777778\\
0.147	99.6527777777778\\
0.1485	99.6527777777778\\
0.15	99.6527777777778\\
0.1515	99.6527777777778\\
0.153	99.6527777777778\\
0.1545	99.6527777777778\\
0.156	99.6527777777778\\
0.1575	99.6527777777778\\
0.159	99.6527777777778\\
0.1605	99.6527777777778\\
0.162	99.6527777777778\\
0.1635	99.6527777777778\\
0.165	99.6527777777778\\
0.1665	99.6527777777778\\
0.168	99.6527777777778\\
0.1695	99.6527777777778\\
0.171	99.6527777777778\\
0.1725	99.6527777777778\\
0.174	99.6527777777778\\
0.1755	99.6527777777778\\
0.177	99.6527777777778\\
0.1785	99.6527777777778\\
0.18	99.6527777777778\\
0.1815	99.6527777777778\\
0.183	99.6527777777778\\
0.1845	99.6527777777778\\
0.186	99.6527777777778\\
0.1875	99.6527777777778\\
0.189	99.6527777777778\\
0.1905	99.6527777777778\\
0.192	99.6527777777778\\
0.1935	99.6527777777778\\
0.195	99.6527777777778\\
0.1965	99.6527777777778\\
0.198	99.6527777777778\\
0.1995	99.6527777777778\\
0.201	99.6527777777778\\
0.2025	99.6527777777778\\
0.204	99.6527777777778\\
0.2055	99.6527777777778\\
0.207	99.6527777777778\\
0.2085	99.6527777777778\\
0.21	99.6527777777778\\
0.2115	99.6527777777778\\
0.213	99.6527777777778\\
0.2145	99.6527777777778\\
0.216	99.6527777777778\\
0.2175	99.6527777777778\\
0.219	99.6527777777778\\
0.2205	99.6527777777778\\
0.222	99.6527777777778\\
0.2235	99.6527777777778\\
0.225	99.6527777777778\\
0.2265	99.6527777777778\\
0.228	99.6527777777778\\
0.2295	99.6527777777778\\
0.231	100\\
};
\addlegendentry{SGMDS \cite{aflalo2016spectral}};

\addplot [color=mycolor2,solid,very thick]
  table[row sep=crcr]{%
0.000210260723296888	13.782991202346\\
0.00273338940285955	14.0762463343109\\
0.00420521446593776	15.8357771260997\\
0.00504625735912532	18.0351906158358\\
0.00609756097560976	20.6744868035191\\
0.00693860386879731	23.3137829912023\\
0.00777964676198486	25.8064516129032\\
0.00862068965517241	28.4457478005865\\
0.00904121110176619	30.6451612903226\\
0.00967199327165686	33.1378299120235\\
0.0103027754415475	35.3372434017595\\
0.0107232968881413	37.2434017595308\\
0.011354079058032	39.5894428152493\\
0.0119848612279226	41.7888563049853\\
0.0128259041211102	44.1348973607038\\
0.0136669470142977	45.8944281524927\\
0.0149285113540791	48.2404692082111\\
0.0159798149705635	51.4662756598241\\
0.0170311185870479	53.3724340175953\\
0.0185029436501262	55.5718475073314\\
0.0197645079899075	58.2111436950147\\
0.0208158116063919	60.7038123167155\\
0.022497897392767	63.3431085043988\\
0.024390243902439	65.6891495601173\\
0.0256518082422204	67.741935483871\\
0.0277544154751892	69.7947214076246\\
0.0294365012615643	71.8475073313783\\
0.0317493692178301	73.7536656891496\\
0.033851976450799	75.6598240469208\\
0.0359545836837679	77.7126099706745\\
0.0382674516400336	79.1788856304985\\
0.0401597981497056	80.3519061583578\\
0.043313708999159	81.8181818181818\\
0.0462573591253154	83.1378299120235\\
0.0492010092514718	84.4574780058651\\
0.052565180824222	85.6304985337243\\
0.0557190916736754	86.5102639296188\\
0.0590832632464256	87.5366568914956\\
0.0632884777123633	88.41642228739\\
0.0674936921783011	89.149560117302\\
0.0710681244743482	89.7360703812317\\
0.0765349032800673	90.3225806451613\\
0.0805298570227082	91.0557184750733\\
0.0841042893187553	91.2023460410557\\
0.088309503784693	91.7888563049853\\
0.0927249789739277	92.375366568915\\
0.0952481076534903	92.6686217008798\\
0.0984020185029437	92.8152492668622\\
0.103027754415475	93.2551319648094\\
0.107022708158116	93.4017595307918\\
0.110597140454163	93.6950146627566\\
0.115222876366695	94.1348973607038\\
0.118797308662742	94.1348973607038\\
0.123212783851976	94.574780058651\\
0.126787216148024	94.7214076246334\\
0.130992430613961	94.8680351906158\\
0.135618166526493	95.0146627565982\\
0.13919259882254	95.3079178885631\\
0.144028595458368	95.4545454545455\\
0.148023549201009	95.4545454545455\\
0.151597981497056	95.4545454545455\\
0.156433978132885	95.7478005865103\\
0.160218671152229	95.7478005865103\\
0.164634146341463	95.8944281524927\\
0.169680403700589	95.8944281524927\\
0.174095878889823	96.0410557184751\\
0.178511354079058	96.1876832844575\\
0.182926829268293	96.3343108504399\\
0.187552565180824	96.3343108504399\\
0.191757779646762	96.4809384164223\\
0.197224558452481	96.6275659824047\\
0.20206055508831	96.6275659824047\\
0.205845248107653	96.6275659824047\\
0.210050462573591	96.7741935483871\\
0.214255677039529	96.7741935483871\\
0.219091673675357	96.7741935483871\\
0.223717409587889	96.7741935483871\\
0.228132884777124	96.9208211143695\\
0.232127838519765	97.0674486803519\\
0.236543313708999	97.0674486803519\\
0.24053826745164	97.2140762463343\\
0.244112699747687	97.2140762463343\\
0.247476871320437	97.2140762463343\\
0.25	97.2140762463343\\
};
\addlegendentry{FM \cite{ovsjanikov2012functional}};

\addplot [color=mycolor3,solid,very thick]
  table[row sep=crcr]{%
0	0.198063\\
0.0025	0.891285\\
0.005	2.30661\\
0.0075	4.4523\\
0.01	7.17017\\
0.0125	10.5579\\
0.015	14.35\\
0.0175	18.3071\\
0.02	22.5435\\
0.0225	26.9614\\
0.025	31.4219\\
0.0275	35.7408\\
0.03	40.1051\\
0.0325	44.2864\\
0.035	48.095\\
0.0375	51.6863\\
0.04	55.107\\
0.0425	58.3517\\
0.045	61.4065\\
0.0475	64.2372\\
0.05	66.867\\
0.0525	69.2699\\
0.055	71.45\\
0.0575	73.4691\\
0.06	75.3617\\
0.0625	77.0975\\
0.065	78.5239\\
0.0675	79.8471\\
0.07	81.0726\\
0.0725	82.0808\\
0.075	83.111\\
0.0775	84.016\\
0.08	84.844\\
0.0825	85.562\\
0.085	86.2387\\
0.0875	86.8522\\
0.09	87.4147\\
0.0925	87.8631\\
0.095	88.3156\\
0.0975	88.6925\\
0.1	89.0584\\
0.1025	89.35\\
0.105	89.6443\\
0.1075	89.8616\\
0.11	90.0762\\
0.1125	90.2578\\
0.115	90.4256\\
0.1175	90.5356\\
0.12	90.6511\\
0.1225	90.7268\\
0.125	90.8231\\
0.1275	90.8863\\
0.13	90.9537\\
0.1325	91.0101\\
0.135	91.0665\\
0.1375	91.1229\\
0.14	91.1752\\
0.1425	91.2178\\
0.145	91.2577\\
0.1475	91.2962\\
0.15	91.3485\\
0.1525	91.3829\\
0.155	91.4338\\
0.1575	91.475\\
0.16	91.5149\\
0.1625	91.5617\\
0.165	91.5851\\
0.1675	91.6236\\
0.17	91.6552\\
0.1725	91.6868\\
0.175	91.7102\\
0.1775	91.7432\\
0.18	91.7817\\
0.1825	91.812\\
0.185	91.8588\\
0.1875	91.8932\\
0.19	91.9275\\
0.1925	91.9564\\
0.195	91.9853\\
0.1975	92.0183\\
0.2	92.0431\\
0.2025	92.0706\\
0.205	92.1146\\
0.2075	92.1421\\
0.21	92.1627\\
0.2125	92.193\\
0.215	92.226\\
0.2175	92.2521\\
0.22	92.2728\\
0.2225	92.2852\\
0.225	92.3113\\
0.2275	92.3319\\
0.23	92.3539\\
0.2325	92.3814\\
0.235	92.3952\\
0.2375	92.4158\\
0.24	92.4392\\
0.2425	92.4543\\
0.245	92.4805\\
0.2475	92.5025\\
};
\addlegendentry{BIM \cite{kim11}};

\addplot [color=mycolor4,solid,very thick]
  table[row sep=crcr]{%
0	2.33687\\
0.0025	2.36576\\
0.005	2.74538\\
0.0075	4.54033\\
0.01	7.55942\\
0.0125	10.3763\\
0.015	12.6238\\
0.0175	15.0171\\
0.02	17.6414\\
0.0225	20.5257\\
0.025	23.2326\\
0.0275	25.806\\
0.03	28.2598\\
0.0325	30.6531\\
0.035	32.9074\\
0.0375	35.0792\\
0.04	37.097\\
0.0425	39.0515\\
0.045	40.9689\\
0.0475	42.6992\\
0.05	44.4088\\
0.0525	46.0209\\
0.055	47.5325\\
0.0575	48.9753\\
0.06	50.3191\\
0.0625	51.6271\\
0.065	52.9269\\
0.0675	54.1167\\
0.07	55.3312\\
0.0725	56.4536\\
0.075	57.4824\\
0.0775	58.4686\\
0.08	59.4025\\
0.0825	60.3488\\
0.085	61.2649\\
0.0875	62.0599\\
0.09	62.8948\\
0.0925	63.6485\\
0.095	64.3197\\
0.0975	64.9923\\
0.1	65.6071\\
0.1025	66.2646\\
0.105	66.8381\\
0.1075	67.3704\\
0.11	67.9124\\
0.1125	68.3442\\
0.115	68.8146\\
0.1175	69.2493\\
0.12	69.6743\\
0.1225	70.0814\\
0.125	70.4569\\
0.1275	70.8641\\
0.13	71.1859\\
0.1325	71.5146\\
0.135	71.8447\\
0.1375	72.1281\\
0.14	72.4238\\
0.1425	72.7003\\
0.145	73.0194\\
0.1475	73.2862\\
0.15	73.5737\\
0.1525	73.8075\\
0.155	74.0152\\
0.1575	74.2573\\
0.16	74.4636\\
0.1625	74.6768\\
0.165	74.8996\\
0.1675	75.0756\\
0.17	75.2655\\
0.1725	75.4484\\
0.175	75.5997\\
0.1775	75.7771\\
0.18	75.9601\\
0.1825	76.1031\\
0.185	76.2888\\
0.1875	76.4332\\
0.19	76.5928\\
0.1925	76.7551\\
0.195	76.9036\\
0.1975	77.0467\\
0.2	77.2048\\
0.2025	77.3561\\
0.205	77.4964\\
0.2075	77.6257\\
0.21	77.7564\\
0.2125	77.9146\\
0.215	78.0535\\
0.2175	78.1704\\
0.22	78.2942\\
0.2225	78.4166\\
0.225	78.5486\\
0.2275	78.6614\\
0.23	78.7825\\
0.2325	78.9062\\
0.235	79.0204\\
0.2375	79.1456\\
0.24	79.2515\\
0.2425	79.378\\
0.245	79.5004\\
0.2475	79.6352\\
};
\addlegendentry{M\"obius Voting \cite{lipman2009mobius}};

\addplot [color=mycolor5,solid,very thick]
  table[row sep=crcr]{%
0	0.148548\\
0.0025	0.532295\\
0.005	1.33968\\
0.0075	2.53081\\
0.01	4.27899\\
0.0125	6.34628\\
0.015	8.79732\\
0.0175	11.5564\\
0.02	14.46\\
0.0225	17.4736\\
0.025	20.622\\
0.0275	23.769\\
0.03	26.7867\\
0.0325	29.7398\\
0.035	32.4329\\
0.0375	35.0682\\
0.04	37.6678\\
0.0425	39.9923\\
0.045	42.3209\\
0.0475	44.5546\\
0.05	46.6398\\
0.0525	48.7731\\
0.055	50.7716\\
0.0575	52.7729\\
0.06	54.627\\
0.0625	56.3889\\
0.065	58.0367\\
0.0675	59.5992\\
0.07	61.0448\\
0.0725	62.4023\\
0.075	63.6251\\
0.0775	64.7997\\
0.08	65.8987\\
0.0825	66.944\\
0.085	67.9688\\
0.0875	68.8559\\
0.09	69.6881\\
0.0925	70.4707\\
0.095	71.2107\\
0.0975	71.8929\\
0.1	72.5792\\
0.1025	73.194\\
0.105	73.7525\\
0.1075	74.2683\\
0.11	74.7607\\
0.1125	75.2256\\
0.115	75.6368\\
0.1175	75.9903\\
0.12	76.3631\\
0.1225	76.7331\\
0.125	77.0398\\
0.1275	77.3823\\
0.13	77.6835\\
0.1325	78.0163\\
0.135	78.3024\\
0.1375	78.5871\\
0.14	78.8911\\
0.1425	79.1593\\
0.145	79.411\\
0.1475	79.6903\\
0.15	79.9461\\
0.1525	80.1909\\
0.155	80.4715\\
0.1575	80.6971\\
0.16	80.9378\\
0.1625	81.129\\
0.165	81.3353\\
0.1675	81.5347\\
0.17	81.7424\\
0.1725	81.9212\\
0.175	82.1165\\
0.1775	82.305\\
0.18	82.5017\\
0.1825	82.6901\\
0.185	82.8441\\
0.1875	83.0532\\
0.19	83.2196\\
0.1925	83.4204\\
0.195	83.5827\\
0.1975	83.734\\
0.2	83.9018\\
0.2025	84.0586\\
0.205	84.2044\\
0.2075	84.3667\\
0.21	84.5387\\
0.2125	84.6996\\
0.215	84.8715\\
0.2175	85.0242\\
0.22	85.1879\\
0.2225	85.3529\\
0.225	85.4864\\
0.2275	85.617\\
0.23	85.7573\\
0.2325	85.91\\
0.235	86.0475\\
0.2375	86.1823\\
0.24	86.3047\\
0.2425	86.4382\\
0.245	86.5743\\
0.2475	86.7311\\
};
\addlegendentry{Best Conformal \cite{kim11}};

\addplot [color=red,solid, line width=2.0pt]
  table[row sep=crcr]{%
0	30.975\\
0.001	31.1166666666667\\
0.002	32.5083333333333\\
0.003	36.025\\
0.004	41.3583333333333\\
0.005	47.0333333333333\\
0.006	51.6333333333333\\
0.007	55.1333333333333\\
0.008	57.6083333333333\\
0.009	60.4\\
0.01	62.25\\
0.011	64.2916666666667\\
0.012	65.9833333333333\\
0.013	67.25\\
0.014	68.2833333333333\\
0.015	69.55\\
0.016	70.6583333333333\\
0.017	71.75\\
0.018	72.6833333333333\\
0.019	73.6833333333333\\
0.02	74.5\\
0.021	75.3583333333333\\
0.022	76.1666666666666\\
0.023	76.9416666666667\\
0.024	77.675\\
0.025	78.3166666666666\\
0.026	78.9583333333333\\
0.027	79.6166666666667\\
0.028	80.375\\
0.029	81.0416666666667\\
0.03	81.6416666666667\\
0.031	82.2\\
0.032	82.9833333333333\\
0.033	83.8333333333333\\
0.034	84.6333333333333\\
0.035	85.3166666666667\\
0.036	85.9833333333333\\
0.037	86.6666666666667\\
0.038	87.2166666666667\\
0.039	87.875\\
0.04	88.6333333333333\\
0.041	89.0916666666667\\
0.042	89.8333333333333\\
0.043	90.55\\
0.044	91.2583333333333\\
0.045	91.8166666666667\\
0.046	92.4666666666667\\
0.047	93.0333333333333\\
0.048	93.7416666666667\\
0.049	94.35\\
0.05	94.9583333333333\\
0.051	95.45\\
0.052	96.0416666666667\\
0.053	96.6416666666667\\
0.054	97.1666666666667\\
0.055	97.65\\
0.056	98.0166666666667\\
0.057	98.3166666666667\\
0.058	98.6\\
0.059	98.8166666666667\\
0.06	98.975\\
0.061	99.1\\
0.062	99.3\\
0.063	99.4416666666667\\
0.064	99.6\\
0.065	99.6916666666667\\
0.066	99.8\\
0.067	99.9083333333333\\
0.068	99.95\\
0.069	100\\
};
\addlegendentry{Ours};

\end{axis}
\end{tikzpicture}%

%% file: figures/curve_tosca.tikz
\definecolor{mycolor1}{rgb}{0.00000,0.44700,0.74100}%
\definecolor{mycolor2}{rgb}{0.85000,0.32500,0.09800}%
\definecolor{mycolor3}{rgb}{0.92900,0.69400,0.12500}%
\definecolor{mycolor4}{rgb}{0.49400,0.18400,0.55600}%
\definecolor{mycolor5}{rgb}{0.46600,0.67400,0.18800}%
\definecolor{mycolor6}{rgb}{0.30100,0.74500,0.93300}%
\begin{tikzpicture}

\begin{axis}[%
width=.85\linewidth,
height=.45\linewidth,
at={(0.797in,0.617in)},
scale only axis,
xmin=0,
xmax=0.1,
ymin=0,
ymax=100,
xmajorgrids,
ymajorgrids,
every x tick label/.append style={font=\color{black}, font=\footnotesize},
every y tick label/.append style={font=\color{black}, font=\footnotesize},
axis background/.style={fill=white},
axis x line*=bottom,
axis y line*=left,
x label style={at={(axis description cs:0.5,0.02)},anchor=north},
y label style={at={(axis description cs:0.1,.5)},rotate=0,anchor=south},
xlabel={\footnotesize Geodesic error},
ylabel={\footnotesize \% Correspondences},
x tick label style={/pgf/number format/fixed},
legend style={at={(0.635,0.02)},anchor=south west,legend cell align=left,align=left,draw=white!15!black,font=\tiny},
]

\addplot [color=mycolor1,solid,line width=1.25pt]
  table[row sep=crcr]{%
0	11.5676759834369\\
0.0015	14.3400621118012\\
0.003	20.8993271221532\\
0.0045	27.8416149068323\\
0.006	35.9042874396135\\
0.0075	43.2116977225673\\
0.009	50.7093253968254\\
0.0105	56.4141218081435\\
0.012	61.2409420289855\\
0.0135	64.5876466528641\\
0.015	67.8847481021394\\
0.0165	71.8249654934438\\
0.018	74.1681763285024\\
0.0195	76.8855676328502\\
0.021	78.8860852311939\\
0.0225	80.7153640441684\\
0.024	82.0867839889579\\
0.0255	83.4331866804693\\
0.027	84.1597653554176\\
0.0285	85.3101276742581\\
0.03	85.9845151828847\\
0.0315	86.5821256038647\\
0.033	87.2888630089717\\
0.0345	87.70294168392\\
0.036	88.1344893029676\\
0.0375	88.6824965493444\\
0.039	89.2503450655625\\
0.0405	89.7586697722567\\
0.042	90.1356538992409\\
0.0435	90.3687888198758\\
0.045	90.688621463078\\
0.0465	91.0656055900621\\
0.048	91.367969289165\\
0.0495	91.7669513457557\\
0.051	92.1215062111802\\
0.0525	92.4389665286405\\
0.054	92.9149413388544\\
0.0555	92.9496635610766\\
0.057	93.3266476880607\\
0.0585	93.6389320220842\\
0.06	93.8869478951001\\
0.0615	94.1597653554175\\
0.063	94.3853519668737\\
0.0645	94.514320220842\\
0.066	94.5738440303658\\
0.0675	94.8913043478261\\
0.069	94.9260265700483\\
0.0705	95.0649154589372\\
0.072	95.1938837129055\\
0.0735	95.3228519668737\\
0.075	95.3922964113182\\
0.0765	95.4889147688061\\
0.078	95.5236369910283\\
0.0795	95.7145013802623\\
0.081	96.1262077294687\\
0.0825	96.443668046929\\
0.084	96.4783902691512\\
0.0855	96.632160110421\\
0.087	96.8206521739131\\
0.0885	96.8900966183575\\
0.09	97.0091442374051\\
0.0915	97.162914078675\\
0.093	97.2323585231194\\
0.0945	97.2670807453416\\
0.096	97.3266045548654\\
0.0975	97.3861283643893\\
0.099	97.3861283643893\\
0.1005	97.4803743961353\\
0.102	97.4803743961353\\
0.1035	97.5150966183575\\
0.105	97.5150966183575\\
0.1065	97.5150966183575\\
0.108	97.5150966183575\\
0.1095	97.6117149758454\\
0.111	97.6464371980676\\
0.1125	97.6464371980676\\
0.114	97.6464371980676\\
0.1155	97.7406832298137\\
0.117	97.7406832298137\\
0.1185	97.7406832298137\\
0.12	97.8002070393375\\
0.1215	97.8002070393375\\
0.123	97.9192546583851\\
0.1245	97.9539768806073\\
0.126	97.9539768806073\\
0.1275	97.9539768806073\\
0.129	98.0135006901311\\
0.1305	98.0135006901311\\
0.132	98.0135006901311\\
0.1335	98.1077467218772\\
0.135	98.1077467218772\\
0.1365	98.1424689440994\\
0.138	98.2367149758454\\
0.1395	98.2367149758454\\
0.141	98.2367149758454\\
0.1425	98.3309610075915\\
0.144	98.4252070393375\\
0.1455	98.4599292615597\\
0.147	98.4599292615597\\
0.1485	98.4946514837819\\
0.15	98.4946514837819\\
0.1515	98.4946514837819\\
0.153	98.5293737060042\\
0.1545	98.588897515528\\
0.156	98.588897515528\\
0.1575	98.588897515528\\
0.159	98.6484213250518\\
0.1605	98.6484213250518\\
0.162	98.6484213250518\\
0.1635	98.683143547274\\
0.165	98.683143547274\\
0.1665	98.683143547274\\
0.168	98.7178657694962\\
0.1695	98.7178657694962\\
0.171	98.7525879917184\\
0.1725	98.7873102139407\\
0.174	98.7873102139407\\
0.1755	98.7873102139407\\
0.177	98.7873102139407\\
0.1785	98.7873102139407\\
0.18	98.7873102139407\\
0.1815	98.7873102139407\\
0.183	98.7873102139407\\
0.1845	98.9063578329883\\
0.186	98.9063578329883\\
0.1875	98.9658816425121\\
0.189	98.9658816425121\\
0.1905	98.9658816425121\\
0.192	98.9658816425121\\
0.1935	98.9658816425121\\
0.195	99.0254054520359\\
0.1965	99.0601276742581\\
0.198	99.0601276742581\\
0.1995	99.0601276742581\\
0.201	99.0601276742581\\
0.2025	99.0601276742581\\
0.204	99.1196514837819\\
0.2055	99.1196514837819\\
0.207	99.1543737060042\\
0.2085	99.1543737060042\\
0.21	99.1543737060042\\
0.2115	99.1543737060042\\
0.213	99.1543737060042\\
0.2145	99.1543737060042\\
0.216	99.1543737060042\\
0.2175	99.1543737060042\\
0.219	99.1543737060042\\
0.2205	99.1543737060042\\
0.222	99.1543737060042\\
0.2235	99.1890959282264\\
0.225	99.1890959282264\\
0.2265	99.1890959282264\\
0.228	99.1890959282264\\
0.2295	99.1890959282264\\
0.231	99.1890959282264\\
0.2325	99.1890959282264\\
0.234	99.1890959282264\\
0.2355	99.1890959282264\\
0.237	99.2486197377502\\
0.2385	99.2486197377502\\
0.24	99.2486197377502\\
0.2415	99.2486197377502\\
0.243	99.2486197377502\\
0.2445	99.2486197377502\\
0.246	99.2486197377502\\
0.2475	99.308143547274\\
0.249	99.308143547274\\
};
\addlegendentry{SGMDS \cite{aflalo2016spectral}};

\addplot [color=mycolor2,solid,line width=1.25pt]
  table[row sep=crcr]{%
0	21.0526315789474\\
0.00114810562571757	22.9018492176387\\
0.00172215843857635	24.75106685633\\
0.00229621125143513	26.6002844950213\\
0.00287026406429392	28.5917496443812\\
0.00315729047072331	30.298719772404\\
0.00373134328358209	32.2901849217639\\
0.00430539609644087	34.1394025604552\\
0.00487944890929966	36.1308677098151\\
0.00545350172215844	37.8378378378378\\
0.00602755453501722	39.8293029871977\\
0.00660160734787601	41.5362731152205\\
0.00717566016073479	43.5277382645804\\
0.00832376578645235	45.3769559032717\\
0.00889781859931114	46.9416785206259\\
0.0100459242250287	48.7908961593172\\
0.0106199770378875	50.49786628734\\
0.0111940298507463	51.778093883357\\
0.0120551090700344	53.2005689900427\\
0.0126291618828932	54.7652916073969\\
0.0134902411021814	56.0455192034139\\
0.0143513203214696	57.7524893314367\\
0.0149253731343284	59.3172119487909\\
0.0160734787600459	61.0241820768137\\
0.0172215843857635	62.7311522048364\\
0.0183696900114811	64.2958748221906\\
0.0195177956371986	65.8605974395448\\
0.0206659012629162	67.2830725462304\\
0.0215269804822044	68.8477951635846\\
0.0229621125143513	70.2702702702703\\
0.0241102181400689	71.4082503556188\\
0.0255453501722158	72.972972972973\\
0.0269804822043628	74.25320056899\\
0.0287026406429392	75.3911806543386\\
0.0301377726750861	76.5291607396871\\
0.0315729047072331	77.6671408250356\\
0.0335820895522388	78.9473684210526\\
0.0350172215843858	80.0853485064011\\
0.0373134328358209	81.3655761024182\\
0.0390355912743972	82.5035561877667\\
0.041044776119403	83.4992887624467\\
0.0433409873708381	84.7795163584637\\
0.0456371986222733	85.7752489331437\\
0.0482204362801378	86.7709815078236\\
0.0510907003444317	87.7667140825036\\
0.054247990815155	88.4779516358464\\
0.0574052812858783	89.3314366998578\\
0.0605625717566016	90.3271692745377\\
0.0637198622273249	90.7539118065434\\
0.067451205510907	91.3229018492176\\
0.0714695752009185	92.0341394025605\\
0.0749138920780712	92.0341394025605\\
0.0783582089552239	92.4608819345662\\
0.0812284730195178	93.0298719772404\\
0.0849598163030999	93.4566145092461\\
0.0889781859931114	93.7411095305832\\
0.0932835820895522	94.0256045519203\\
0.0967278989667049	94.3100995732575\\
0.100746268656716	94.452347083926\\
0.104477611940299	94.5945945945946\\
0.107634902411022	94.8790896159317\\
0.111653272101033	95.3058321479374\\
0.115384615384615	95.5903271692745\\
0.119402985074627	95.8748221906117\\
0.12284730195178	96.0170697012802\\
0.12743972445465	96.3015647226173\\
0.131458094144661	96.5860597439545\\
0.134615384615385	96.5860597439545\\
0.138059701492537	96.7283072546231\\
0.140929965556831	96.5860597439545\\
0.144087256027555	96.7283072546231\\
0.147531572904707	96.7283072546231\\
0.151549942594719	96.7283072546231\\
0.154994259471871	96.8705547652916\\
0.159299655568312	97.0128022759602\\
0.163030998851894	97.0128022759602\\
0.167336394948335	96.8705547652916\\
0.171354764638347	96.8705547652916\\
0.175373134328358	97.0128022759602\\
0.179678530424799	97.0128022759602\\
0.18398392652124	97.1550497866287\\
0.188289322617681	97.1550497866287\\
0.192020665901263	97.1550497866287\\
0.196900114810563	97.2972972972973\\
0.200918484500574	97.2972972972973\\
0.205223880597015	97.4395448079659\\
0.209242250287026	97.4395448079659\\
0.213260619977038	97.2972972972973\\
0.21699196326062	97.4395448079659\\
0.221297359357061	97.4395448079659\\
0.224741676234214	97.5817923186344\\
0.228473019517796	97.5817923186344\\
0.232204362801378	97.5817923186344\\
0.23593570608496	97.724039829303\\
0.239380022962113	97.724039829303\\
0.243398392652124	97.8662873399716\\
0.247416762342135	97.724039829303\\
0.25	97.8662873399716\\
};
\addlegendentry{FM \cite{ovsjanikov2012functional}};

\addplot [color=mycolor3,solid,line width=1.25pt]
  table[row sep=crcr]{%
0	0.950928\\
0.0025	4.64478\\
0.005	9.49219\\
0.0075	15.0183\\
0.01	21.0095\\
0.0125	27.1948\\
0.015	33.3997\\
0.0175	39.5203\\
0.02	45.3137\\
0.0225	50.6812\\
0.025	55.9033\\
0.0275	60.4321\\
0.03	64.6362\\
0.0325	68.252\\
0.035	71.571\\
0.0375	74.4446\\
0.04	77.002\\
0.0425	79.2944\\
0.045	81.2939\\
0.0475	83.0432\\
0.05	84.6729\\
0.0525	86.2305\\
0.055	87.5488\\
0.0575	88.6487\\
0.06	89.6265\\
0.0625	90.4895\\
0.065	91.4148\\
0.0675	92.146\\
0.07	92.8247\\
0.0725	93.4058\\
0.075	93.9404\\
0.0775	94.3921\\
0.08	94.7803\\
0.0825	95.0964\\
0.085	95.3918\\
0.0875	95.6604\\
0.09	95.8862\\
0.0925	96.084\\
0.095	96.2659\\
0.0975	96.4355\\
0.1	96.593\\
0.1025	96.7273\\
0.105	96.8311\\
0.1075	96.9165\\
0.11	97.0288\\
0.1125	97.1191\\
0.115	97.1826\\
0.1175	97.2656\\
0.12	97.3376\\
0.1225	97.417\\
0.125	97.4915\\
0.1275	97.55\\
0.13	97.6172\\
0.1325	97.6758\\
0.135	97.738\\
0.1375	97.7954\\
0.14	97.8394\\
0.1425	97.8943\\
0.145	97.9468\\
0.1475	97.9944\\
0.15	98.0322\\
0.1525	98.0688\\
0.155	98.1079\\
0.1575	98.158\\
0.16	98.2104\\
0.1625	98.2422\\
0.165	98.291\\
0.1675	98.3374\\
0.17	98.3765\\
0.1725	98.4143\\
0.175	98.4497\\
0.1775	98.4863\\
0.18	98.5205\\
0.1825	98.5706\\
0.185	98.6084\\
0.1875	98.6438\\
0.19	98.6853\\
0.1925	98.7097\\
0.195	98.7549\\
0.1975	98.7817\\
0.2	98.8123\\
0.2025	98.8416\\
0.205	98.8745\\
0.2075	98.8989\\
0.21	98.9197\\
0.2125	98.9417\\
0.215	98.9709\\
0.2175	98.988\\
0.22	99.0088\\
0.2225	99.0417\\
0.225	99.0674\\
0.2275	99.082\\
0.23	99.0979\\
0.2325	99.1138\\
0.235	99.1235\\
0.2375	99.1382\\
0.24	99.1516\\
0.2425	99.1638\\
0.245	99.1797\\
0.2475	99.187\\
};
\addlegendentry{BIM \cite{kim11}};

\addplot [color=mycolor4,solid,line width=1.25pt]
  table[row sep=crcr]{%
0	10.5835\\
0.0025	11.1829\\
0.005	13.0859\\
0.0075	16.0339\\
0.01	19.4641\\
0.0125	22.8308\\
0.015	26.3\\
0.0175	29.718\\
0.02	33.1006\\
0.0225	36.2756\\
0.025	39.3176\\
0.0275	42.1777\\
0.03	44.8804\\
0.0325	47.4158\\
0.035	49.696\\
0.0375	51.7615\\
0.04	53.855\\
0.0425	55.7483\\
0.045	57.5366\\
0.0475	59.2407\\
0.05	60.9021\\
0.0525	62.3523\\
0.055	63.7659\\
0.0575	65.1318\\
0.06	66.3611\\
0.0625	67.6086\\
0.065	68.7646\\
0.0675	69.8096\\
0.07	70.7654\\
0.0725	71.7786\\
0.075	72.7368\\
0.0775	73.6047\\
0.08	74.4104\\
0.0825	75.199\\
0.085	75.9192\\
0.0875	76.5991\\
0.09	77.2339\\
0.0925	77.8552\\
0.095	78.479\\
0.0975	79.0784\\
0.1	79.6338\\
0.1025	80.1062\\
0.105	80.5847\\
0.1075	81.0669\\
0.11	81.4832\\
0.1125	81.9104\\
0.115	82.2632\\
0.1175	82.6721\\
0.12	83.0481\\
0.1225	83.3862\\
0.125	83.7183\\
0.1275	84.0161\\
0.13	84.292\\
0.1325	84.5654\\
0.135	84.8511\\
0.1375	85.0989\\
0.14	85.3503\\
0.1425	85.614\\
0.145	85.8276\\
0.1475	86.0425\\
0.15	86.2451\\
0.1525	86.4612\\
0.155	86.6479\\
0.1575	86.8127\\
0.16	86.9666\\
0.1625	87.124\\
0.165	87.2839\\
0.1675	87.4536\\
0.17	87.6062\\
0.1725	87.7844\\
0.175	87.9614\\
0.1775	88.1152\\
0.18	88.2751\\
0.1825	88.4058\\
0.185	88.5486\\
0.1875	88.6926\\
0.19	88.811\\
0.1925	88.9453\\
0.195	89.0991\\
0.1975	89.2322\\
0.2	89.3652\\
0.2025	89.5032\\
0.205	89.6216\\
0.2075	89.7131\\
0.21	89.8047\\
0.2125	89.9109\\
0.215	90.0159\\
0.2175	90.1196\\
0.22	90.2271\\
0.2225	90.3003\\
0.225	90.4004\\
0.2275	90.4687\\
0.23	90.564\\
0.2325	90.6445\\
0.235	90.7397\\
0.2375	90.8435\\
0.24	90.9241\\
0.2425	91.0229\\
0.245	91.1267\\
0.2475	91.2146\\
};
\addlegendentry{M\"obius Voting \cite{lipman2009mobius}};

\addplot [color=mycolor5,solid,line width=1.25pt]
  table[row sep=crcr]{%
0	0.848389\\
0.0025	3.53516\\
0.005	6.64307\\
0.0075	10.3369\\
0.01	14.3311\\
0.0125	18.4082\\
0.015	22.4023\\
0.0175	26.0754\\
0.02	29.5435\\
0.0225	32.9993\\
0.025	36.145\\
0.0275	39.2737\\
0.03	42.1802\\
0.0325	44.9548\\
0.035	47.3743\\
0.0375	49.6497\\
0.04	51.7737\\
0.0425	53.7781\\
0.045	55.6995\\
0.0475	57.5964\\
0.05	59.2493\\
0.0525	60.835\\
0.055	62.2278\\
0.0575	63.551\\
0.06	64.6619\\
0.0625	65.6543\\
0.065	66.6443\\
0.0675	67.5708\\
0.07	68.4631\\
0.0725	69.3323\\
0.075	70.1587\\
0.0775	70.9509\\
0.08	71.6516\\
0.0825	72.3596\\
0.085	72.9553\\
0.0875	73.5046\\
0.09	74.0466\\
0.0925	74.563\\
0.095	75.0195\\
0.0975	75.4736\\
0.1	75.8716\\
0.1025	76.272\\
0.105	76.6541\\
0.1075	77.0129\\
0.11	77.3706\\
0.1125	77.7307\\
0.115	78.092\\
0.1175	78.418\\
0.12	78.7598\\
0.1225	79.0674\\
0.125	79.3872\\
0.1275	79.7192\\
0.13	79.9976\\
0.1325	80.3125\\
0.135	80.5725\\
0.1375	80.9082\\
0.14	81.1963\\
0.1425	81.5051\\
0.145	81.8066\\
0.1475	82.1106\\
0.15	82.3694\\
0.1525	82.6807\\
0.155	82.9358\\
0.1575	83.2043\\
0.16	83.501\\
0.1625	83.7524\\
0.165	84.0015\\
0.1675	84.2639\\
0.17	84.5288\\
0.1725	84.7986\\
0.175	85.0513\\
0.1775	85.3198\\
0.18	85.5774\\
0.1825	85.8167\\
0.185	86.0876\\
0.1875	86.3037\\
0.19	86.5344\\
0.1925	86.7688\\
0.195	87.0239\\
0.1975	87.2681\\
0.2	87.4658\\
0.2025	87.6978\\
0.205	87.9443\\
0.2075	88.1384\\
0.21	88.3496\\
0.2125	88.5242\\
0.215	88.6926\\
0.2175	88.9063\\
0.22	89.0979\\
0.2225	89.3066\\
0.225	89.5215\\
0.2275	89.6802\\
0.23	89.8694\\
0.2325	90.0452\\
0.235	90.2136\\
0.2375	90.3882\\
0.24	90.5615\\
0.2425	90.7056\\
0.245	90.8728\\
0.2475	91.0339\\
};
\addlegendentry{Best Conformal \cite{kim11}};

\addplot [color=red,solid,line width=2.0pt]
  table[row sep=crcr]{%
0	43.2027397260274\\
0.001	43.2205479452055\\
0.002	43.4205479452055\\
0.003	44.6698630136986\\
0.004	48.7123287671233\\
0.005	53.7109589041096\\
0.006	58.5808219178082\\
0.007	62.9561643835616\\
0.008	66.7397260273973\\
0.009	70.1397260273973\\
0.01	73.1315068493151\\
0.011	75.6205479452055\\
0.012	77.8643835616438\\
0.013	79.9164383561644\\
0.014	81.6657534246576\\
0.015	83.2520547945206\\
0.016	84.6630136986301\\
0.017	85.9123287671233\\
0.018	87.0493150684931\\
0.019	88.0917808219178\\
0.02	88.9986301369863\\
0.021	89.7479452054795\\
0.022	90.4794520547945\\
0.023	91.1301369863014\\
0.024	91.7232876712329\\
0.025	92.2506849315068\\
0.026	92.6972602739726\\
0.027	93.127397260274\\
0.028	93.5301369863013\\
0.029	93.8780821917808\\
0.03	94.1904109589041\\
0.031	94.5383561643836\\
0.032	94.8219178082191\\
0.033	95.086301369863\\
0.034	95.331506849315\\
0.035	95.5438356164384\\
0.036	95.7493150684932\\
0.037	95.9520547945206\\
0.038	96.1205479452055\\
0.039	96.2794520547945\\
0.04	96.458904109589\\
0.041	96.6191780821918\\
0.042	96.772602739726\\
0.043	96.9027397260274\\
0.044	97.0342465753425\\
0.045	97.1589041095891\\
0.046	97.2739726027397\\
0.047	97.3958904109589\\
0.048	97.5287671232876\\
0.049	97.6260273972602\\
0.05	97.7506849315068\\
0.051	97.827397260274\\
0.052	97.913698630137\\
0.053	97.986301369863\\
0.054	98.058904109589\\
0.055	98.1191780821917\\
0.056	98.1671232876712\\
0.057	98.213698630137\\
0.058	98.2684931506849\\
0.059	98.3041095890411\\
0.06	98.3369863013698\\
0.061	98.3767123287671\\
0.062	98.4109589041095\\
0.063	98.4438356164383\\
0.064	98.4712328767123\\
0.065	98.4876712328767\\
0.066	98.5123287671233\\
0.067	98.5397260273973\\
0.068	98.5561643835616\\
0.069	98.5794520547945\\
0.07	98.5931506849315\\
0.071	98.6041095890411\\
0.072	98.6164383561644\\
0.073	98.6246575342466\\
0.074	98.6342465753425\\
0.075	98.6397260273973\\
0.076	98.6534246575342\\
0.077	98.6671232876713\\
0.078	98.6808219178082\\
0.079	98.6904109589041\\
0.08	98.7\\
0.081	98.7068493150685\\
0.082	98.7164383561644\\
0.083	98.7246575342466\\
0.084	98.7315068493151\\
0.085	98.7465753424657\\
0.086	98.7547945205479\\
0.087	98.7616438356164\\
0.088	98.7753424657534\\
0.089	98.7794520547945\\
0.09	98.7835616438356\\
0.091	98.7876712328767\\
0.092	98.7945205479452\\
0.093	98.8\\
0.094	98.8123287671233\\
0.095	98.8205479452055\\
0.096	98.827397260274\\
0.097	98.8369863013699\\
0.098	98.8424657534247\\
0.099	98.8534246575342\\
0.1	98.8561643835616\\
0.101	98.8657534246575\\
0.102	98.8684931506849\\
0.103	98.8698630136986\\
0.104	98.8767123287671\\
0.105	98.8849315068493\\
0.106	98.8958904109589\\
0.107	98.9082191780822\\
0.108	98.9164383561644\\
0.109	98.9191780821918\\
0.11	98.9205479452055\\
0.111	98.9260273972603\\
0.112	98.9315068493151\\
0.113	98.9328767123288\\
0.114	98.941095890411\\
0.115	98.9465753424657\\
0.116	98.9534246575342\\
0.117	98.958904109589\\
0.118	98.9657534246575\\
0.119	98.9739726027397\\
0.12	98.9835616438356\\
0.121	98.9890410958904\\
0.122	98.9958904109589\\
0.123	99\\
0.124	99.0013698630137\\
0.125	99.0082191780822\\
0.126	99.0164383561644\\
0.127	99.0232876712329\\
0.128	99.027397260274\\
0.129	99.0342465753425\\
0.13	99.0424657534247\\
0.131	99.0479452054795\\
0.132	99.0547945205479\\
0.133	99.0602739726027\\
0.134	99.0616438356164\\
0.135	99.0712328767123\\
0.136	99.072602739726\\
0.137	99.0739726027397\\
0.138	99.0821917808219\\
0.139	99.0849315068493\\
0.14	99.0876712328767\\
0.141	99.0904109589041\\
0.142	99.0945205479452\\
0.143	99.1\\
0.144	99.1013698630137\\
0.145	99.1109589041096\\
0.146	99.113698630137\\
0.147	99.1219178082192\\
0.148	99.1301369863014\\
0.149	99.1356164383562\\
0.15	99.1438356164383\\
0.151	99.1561643835616\\
0.152	99.158904109589\\
0.153	99.1630136986301\\
0.154	99.1712328767123\\
0.155	99.1753424657534\\
0.156	99.1780821917808\\
0.157	99.1849315068493\\
0.158	99.1890410958904\\
0.159	99.1890410958904\\
0.16	99.1945205479452\\
0.161	99.1958904109589\\
0.162	99.2\\
0.163	99.2027397260274\\
0.164	99.2068493150685\\
0.165	99.2123287671233\\
0.166	99.2123287671233\\
0.167	99.2164383561644\\
0.168	99.2219178082192\\
0.169	99.2246575342466\\
0.17	99.2301369863014\\
0.171	99.2369863013699\\
0.172	99.2424657534247\\
0.173	99.2479452054794\\
0.174	99.2534246575342\\
0.175	99.2561643835616\\
0.176	99.2616438356164\\
0.177	99.2643835616438\\
0.178	99.2657534246575\\
0.179	99.2753424657534\\
0.18	99.2808219178082\\
0.181	99.286301369863\\
0.182	99.2904109589041\\
0.183	99.2972602739726\\
0.184	99.3041095890411\\
0.185	99.3095890410959\\
0.186	99.3109589041096\\
0.187	99.313698630137\\
0.188	99.3150684931507\\
0.189	99.3178082191781\\
0.19	99.3191780821918\\
0.191	99.3191780821918\\
0.192	99.3232876712329\\
0.193	99.3301369863014\\
0.194	99.3342465753425\\
0.195	99.3342465753425\\
0.196	99.3369863013699\\
0.197	99.3383561643836\\
0.198	99.3397260273973\\
0.199	99.341095890411\\
0.2	99.3465753424658\\
0.201	99.3465753424658\\
0.202	99.3520547945205\\
0.203	99.3534246575342\\
0.204	99.358904109589\\
0.205	99.3602739726027\\
0.206	99.3616438356164\\
0.207	99.3630136986301\\
0.208	99.3643835616438\\
0.209	99.3657534246575\\
0.21	99.3657534246575\\
0.211	99.3684931506849\\
0.212	99.3684931506849\\
0.213	99.372602739726\\
0.214	99.3767123287671\\
0.215	99.3780821917808\\
0.216	99.3835616438356\\
0.217	99.386301369863\\
0.218	99.3917808219178\\
0.219	99.3931506849315\\
0.22	99.3986301369863\\
0.221	99.4\\
0.222	99.4013698630137\\
0.223	99.4041095890411\\
0.224	99.4054794520548\\
0.225	99.4095890410959\\
0.226	99.4109589041096\\
0.227	99.4123287671233\\
0.228	99.4123287671233\\
0.229	99.413698630137\\
0.23	99.413698630137\\
0.231	99.4164383561644\\
0.232	99.4178082191781\\
0.233	99.4219178082192\\
0.234	99.4219178082192\\
0.235	99.4232876712329\\
0.236	99.4260273972603\\
0.237	99.4287671232877\\
0.238	99.4301369863014\\
0.239	99.4315068493151\\
0.24	99.4328767123288\\
0.241	99.4342465753425\\
0.242	99.4356164383562\\
0.243	99.441095890411\\
0.244	99.4438356164383\\
0.245	99.4493150684931\\
0.246	99.4493150684931\\
0.247	99.4520547945205\\
0.248	99.4547945205479\\
0.249	99.4575342465753\\
0.25	99.4602739726027\\
};
\addlegendentry{Ours};

\end{axis}
\end{tikzpicture}%

%% file: figures/curve_topkids.tikz
\definecolor{mycolor1}{rgb}{0.00000,0.44700,0.74100}%
\definecolor{mycolor2}{rgb}{0.85000,0.32500,0.09800}%
\definecolor{mycolor3}{rgb}{0.92900,0.69400,0.12500}%
\definecolor{mycolor4}{rgb}{0.49400,0.18400,0.55600}%
\definecolor{mycolor5}{rgb}{0.46600,0.67400,0.18800}%
\definecolor{mycolor6}{rgb}{0.30100,0.74500,0.93300}%
\begin{tikzpicture}

\begin{axis}[%
width=0.85\linewidth,
height=0.45\linewidth,
at={(0.772in,0.516in)},
scale only axis,
xmin=0,
xmax=0.25,
ymin=0,
ymax=100,
xmajorgrids,
ymajorgrids,
every x tick label/.append style={font=\color{black}, font=\footnotesize},
every y tick label/.append style={font=\color{black}, font=\footnotesize},
axis background/.style={fill=white},
axis x line*=bottom,
axis y line*=left,
x label style={at={(axis description cs:0.5,0.02)},anchor=north},
y label style={at={(axis description cs:0.1,.5)},rotate=0,anchor=south},
xlabel={\footnotesize Geodesic error},
ylabel={\footnotesize \% Correspondences},
x tick label style={/pgf/number format/fixed},
legend style={at={(0.73,0.02)},anchor=south west,legend cell align=left,align=left,draw=white!15!black,font=\tiny},
]
\addplot [color=mycolor1,solid,very thick]
  table[row sep=crcr]{%
0	2.59186656367258\\
0.01	3.38915028136611\\
0.02	4.76359272323923\\
0.03	6.52470939909646\\
0.04	8.66421081963232\\
0.05	11.8310706659159\\
0.06	16.7404270618203\\
0.07	21.1118311346637\\
0.08	24.6987633481227\\
0.09	28.2922523997457\\
0.1	31.8255472986559\\
0.11	35.0297464125853\\
0.12	38.062761187677\\
0.13	41.2638983515032\\
0.14	44.2139106471693\\
0.15	46.8414323947308\\
0.16	49.0553276500261\\
0.17	51.3558860281895\\
0.18	53.692981145922\\
0.19	55.65570627831\\
0.2	57.6074265022789\\
0.21	59.3695124462739\\
0.22	61.1769616789454\\
0.23	62.8190978126739\\
0.24	64.3191684829309\\
0.25	65.6390002646242\\
};
\addlegendentry{EM \cite{sahi12}};

\addplot [color=mycolor6,solid,very thick]
  table[row sep=crcr]{%
0	0.445294555428997\\
0.01	1.31245446341963\\
0.02	4.51000462883511\\
0.03	9.08664672498074\\
0.04	14.3987000966751\\
0.05	19.9703394871348\\
0.06	25.5057257828931\\
0.07	30.9008892302496\\
0.08	35.7647952028367\\
0.09	40.0027260342255\\
0.1	43.7512922427831\\
0.11	47.1431314799796\\
0.12	50.2471758130442\\
0.13	53.06279522252\\
0.14	55.6360768869578\\
0.15	58.0082396103639\\
0.16	60.1506136802062\\
0.17	62.1230959585937\\
0.18	64.0208762391739\\
0.19	65.833213079631\\
0.2	67.5642360581011\\
0.21	69.2517508285218\\
0.22	70.8398579735419\\
0.23	72.3378073837109\\
0.24	73.7134272564856\\
0.25	75.0557555225489\\
};
\addlegendentry{GE \cite{SHREC16-topology}};

\addplot [color=mycolor3,solid,very thick]
  table[row sep=crcr]{%
0	6.69306308296569\\
0.01	9.75182598172478\\
0.02	16.0739754499995\\
0.03	21.9105202493059\\
0.04	27.0985167507045\\
0.05	31.6650974236389\\
0.06	35.7293417273631\\
0.07	39.3590888180227\\
0.08	42.5098297810401\\
0.09	45.2774398987196\\
0.1	47.7124888384332\\
0.11	49.9247027830963\\
0.12	51.8717109345963\\
0.13	53.5816911524929\\
0.14	55.2548294222688\\
0.15	56.7746062699835\\
0.16	58.2054164273659\\
0.17	59.547385848875\\
0.18	60.8408060035973\\
0.19	62.0718751224272\\
0.2	63.1909999291417\\
0.21	64.3272975452623\\
0.22	65.5001824431898\\
0.23	66.6245064393347\\
0.24	67.8089497477715\\
0.25	68.9986900611532\\
};
\addlegendentry{RF \cite{rodola2014dense}};

\addplot [color=mycolor4,solid,very thick]
  table[row sep=crcr]{%
0	12.1490007876752\\
0.01	16.6260384675421\\
0.02	25.9923280374089\\
0.03	35.0914755083157\\
0.04	43.1519798518944\\
0.05	49.918114386159\\
0.06	55.7261663354573\\
0.07	60.559244905773\\
0.08	64.5925689984792\\
0.09	67.8366672504121\\
0.1	70.4490294407048\\
0.11	72.6095039776298\\
0.12	74.3884020322704\\
0.13	76.000852161099\\
0.14	77.39950441352\\
0.15	78.623244011866\\
0.16	79.655938640508\\
0.17	80.5413459104384\\
0.18	81.3337657856216\\
0.19	82.0575800981134\\
0.2	82.7096111357128\\
0.21	83.3063080808233\\
0.22	83.882455902221\\
0.23	84.4490835905116\\
0.24	85.0035803053277\\
0.25	85.537218465754\\
};
\addlegendentry{FSPM \cite{litany2017fullyspectral}};

\addplot [color=mycolor5,solid,very thick]
  table[row sep=crcr]{%
0	10.1253362704985\\
0.01	14.0455175839875\\
0.02	22.3635355525081\\
0.03	30.3815749580386\\
0.04	37.2965102353228\\
0.05	43.0612099627044\\
0.06	47.9731524933252\\
0.07	52.0428249453636\\
0.08	55.392027532148\\
0.09	58.149352327709\\
0.1	60.4236849518351\\
0.11	62.2639907968202\\
0.12	63.8372234208708\\
0.13	65.1922017414349\\
0.14	66.4134371879144\\
0.15	67.5242317566194\\
0.16	68.4530869483707\\
0.17	69.3341994958915\\
0.18	70.1356317891122\\
0.19	70.8964769162009\\
0.2	71.6249301381613\\
0.21	72.3274623421685\\
0.22	73.0321758973901\\
0.23	73.7372853135066\\
0.24	74.4914854752655\\
0.25	75.2754485097991\\
};
\addlegendentry{PFM \cite{rodola16-partial}};

\addplot [color=red,solid,line width=2.0pt]
  table[row sep=crcr]{%
0	31.2591531285357\\
0.01	45.4756338590483\\
0.02	62.343542771077\\
0.03	71.5036978862531\\
0.04	76.6024171894662\\
0.05	79.6623105863754\\
0.06	81.6241010041431\\
0.07	82.9318910750074\\
0.08	83.8300034176848\\
0.09	84.4867969989308\\
0.1	85.0171715726564\\
0.11	85.4640619704382\\
0.12	85.8790169041412\\
0.13	86.2374738246235\\
0.14	86.5873349228953\\
0.15	86.910540845052\\
0.16	87.2198401453331\\
0.17	87.5276471314084\\
0.18	87.8182207596991\\
0.19	88.1233875703617\\
0.2	88.4184086864735\\
0.21	88.7116965592408\\
0.22	89.0114058891813\\
0.23	89.2790980082497\\
0.24	89.5614571822996\\
0.25	89.8492599814893\\
};
\addlegendentry{Ours};

\end{axis}
\end{tikzpicture}%

%% file: 06_conclusion.tex
\section{Conclusions}

\noindent We considered a formulation of the problem of finding a smooth, possibly partial, correspondence between two non-isometric shapes as a quadratic assignment problem matching between point-wise and pair-wise descriptors.
We showed that when choosing the pair-wise descriptors to be positive-definite kernel matrices (unlike the traditionally used distance matrices), the NP-hard QAP admits an exact relaxation over the space of  bistochastic matrices, which we proposed to solve using a projected descent procedure motivated by the DC algorithm. The resulting iterations take the form of LAPs, which are solved using a multi-scale version of the auction algorithm. We interpreted the proposed algorithm as an alternating diffusion process, as iterated blurring and sharpening, and as a kernel density estimation procedure. 
%
%Despite the relatively high complexity of the involved LAPs
%The algorithm scales very well to many tens and even hundreds of thousands of vertices, and produces surprisingly good results. Experimental evaluation on various datasets shows that our method significantly improves the output obtained by the best existing correspondence methods.
The algorithm scales very well to even hundreds of thousands of vertices, and produces surprisingly good results. Experimental evaluation on various datasets shows that our method significantly improves the output obtained by the best existing correspondence methods.

%One potential way to accelerate our code and reduce its memory complexity is the use of truncated heat kernels in factorized form. We intend to explore this direction in follow-up studies.

\paragraph*{Acknowledgements}

This work has been supported by the ERC grants 307047 (COMET), 335491 (RAPID), 649323 (3D Reloaded) and 724228 (LEMAN).
%This work has been supported by the ERC grants COMET, RAPID, 3D Reloaded and LEMAN.
%This work has been supported by the ERC grants 307047, 335491, 649323 and 724228.

%% file: 07_supplMaterial.tex
\section*{Supplementary Material}
\appendix
\section{Details about the DC algorithm}

In Section 3 of the submission we propose to use the DC algorithm to optimize
\begin{align}\label{energy}
 \argmin_{\mathbf P\in \Rr^{n\times n}} B(\mathbf P)-E(\mathbf P).
\end{align}
where $B$ is the convex indicator function of the set of bistochastic matrices $\Dn$ and $E$ is strictly convex and differentiable.
We will now prove that the two steps
\begin{itemize}
\item Select $\mathbf Q^k \in \partial E(\mathbf P^k)$
\item Select $\mathbf P^{k+1} \in \partial B^*(\mathbf Q^k)$.
\end{itemize}
of the DC algorithm are equivalent to
\begin{align}
\mathbf P^{k+1} = \argmax_{\mathbf P\in \Dn} \langle \mathbf P, \nabla  E(\mathbf P^k) \rangle \,,
\end{align}
that each iterate $\mathbf P^k$ can be chosen to be a permutation matrix, and that $E(\mathbf P^k)$ is a strictly increasing.

We assume that the reader is familiar with the concepts of convex conjugates and sub-gradients and just recall the following Lemma

\begin{lemma}\label{lemma:equiv}
Let $X$ be a Banach space and $f:X\rightarrow (-\infty,\infty]$ with $\partial f\neq \emptyset$. Then $f^{**}(x) = f(x)$ and
\begin{align}
 x^*\in \partial f(x) &\Leftrightarrow x\in \partial f^*(x^*)
\end{align}
\end{lemma}
% Reference:
% http://www-m6.ma.tum.de/~brokate/con_ss09.pdf
% Standard textbook, Rockafellar?
% Is this Fenchel duality theorem?

Moreover for convex functions $f$, $0\in \partial f(x)$ is equivalent to
\begin{align}
 x &= \argmin_{x} f(x)
\end{align}

Let now $E$ be convex differentiable and $B$ the (convex) indicator function of a convex set $C$. We will derive equivalent expressions for the two steps in the DC algorithm for solving \eqref{energy}. Since $E$ is differentiable, its subdifferential at any point has one element, namely the gradient at that point:
\begin{align}
 \mathbf Q^k\in \partial E(\mathbf P^k) &\Leftrightarrow \mathbf Q^k=\nabla E(\mathbf P^k)
\end{align}

The second step $\mathbf P^{k+1} \in \partial B^*(\mathbf Q^k)$ can be rewritten using Lemma \ref{lemma:equiv}:
\begin{align}
  \mathbf P^{k+1} \in \partial B^*(\mathbf Q^k) & \Leftrightarrow \mathbf Q^k \in \partial B(\mathbf P^{k+1})\nonumber \\
   & \Leftrightarrow 0 \in - \mathbf Q^k + \partial B(\mathbf P^{k+1})\nonumber \\
  & \Leftrightarrow \mathbf P^{k+1} = \argmin_{\mathbf P} - \langle \mathbf Q^k,\mathbf P \rangle  + B(x)\nonumber \\
   & \Leftrightarrow \mathbf P^{k+1} = \argmax_{\mathbf P\in C} \langle \mathbf Q^k,\mathbf P \rangle
\end{align}

Thus the DC algorithm in this special case reads
\begin{align}\label{eq:update}
\mathbf P^{k+1} = \argmax_{\mathbf P\in C} \langle \mathbf P, \nabla  E(\mathbf P^k) \rangle \,.
\end{align}

In our case the convex set $C$ is the polyhedron $\Dn$ of bistochastic matrices.
Since linear functions defined on a polyhedron attain their extrema at the vertices of the polyhedron, we can choose the maximizer to be a permutation matrix. %Can we say more because E is strictly convex?

Due to the strict convexity of $E$ we further see:
\begin{align}
 E(\mathbf P^{k+1}) &> E(\mathbf P^{k}) + \langle\mathbf P^{k+1}-\mathbf P^{k},\nabla E(\mathbf P^{k}) \rangle\nonumber \\
 &\geq E(\mathbf P^{k}) + \langle\mathbf P^{k}-\mathbf P^{k},\nabla E(\mathbf P^{k}) \rangle\nonumber \\
 &=  E(\mathbf P^{k}) 
\end{align}
where the strong inequality holds until convergence and the weak inequality follows directly from \eqref{eq:update}.

%----------------------------------------------------------------------------------------------------------------

% \section{Additional details about heat kernels}
% The following lemma states the connection between geodesic distances and heat kernels. 
% \begin{lemma}
% \label{lemma:Varadahan}(Varadahan's lemma)
% \cite{varadhan1967behavior} 
% \begin{equation}\label{eq:VaradahanLemma}
% \forall x,x'\in \mathcal{X}\;\;
% \lim_{t\rightarrow 0}-4t \ln\left( {K_{\Xx}}_t(x,x')\right)=d_\Xx^2(x,x')
% \end{equation}
% \end{lemma}
% \begin{proposition}\label{prop:heatKernelIsometry}
% Two manifolds $\Xx$ and $\Yy$ are isometric if and only if there exists a surjective map $\varphi:\Xx \rightarrow \Yy$ such that ${k_{\Xx}}_t(x,x')={k_{\Yy}}_t(\varphi(x), \varphi(x'))\;\; \forall x,x' \in X, t\in \mathbb{R}_+.$ In that case $
% \varphi$ is the isometry.
% \end{proposition}
% \begin{proof}
% The if part is a direct consequence of Lemma 
% \ref{lemma:Varadahan}. The only if part is a consequence of the fact that the (negative-definite) Laplace-Beltrami operator $\Delta$ is invariant to isometric deformations, and that $k = e^{t\Delta}$.
% \end{proof}

% Corollary \ref{prop:heatKernelIsometry} shows that in principle, there is no compromise in matching heat kernels rather than distances for the case that the shapes are isometric.

% \paragraph{Relation to spectral Gromov-Hausdorff distance}

%----------------------------------------------------------------------------------------------------------------

\begin{figure}[b]
\centering
	\includegraphics[width=.4\linewidth]{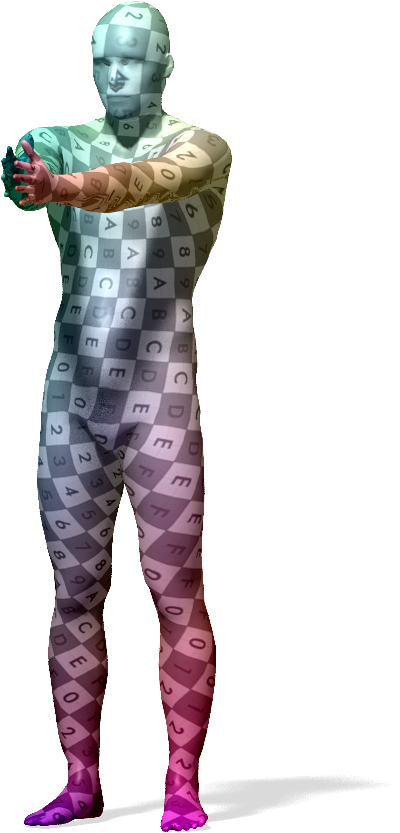}
    \includegraphics[width=.533\linewidth]{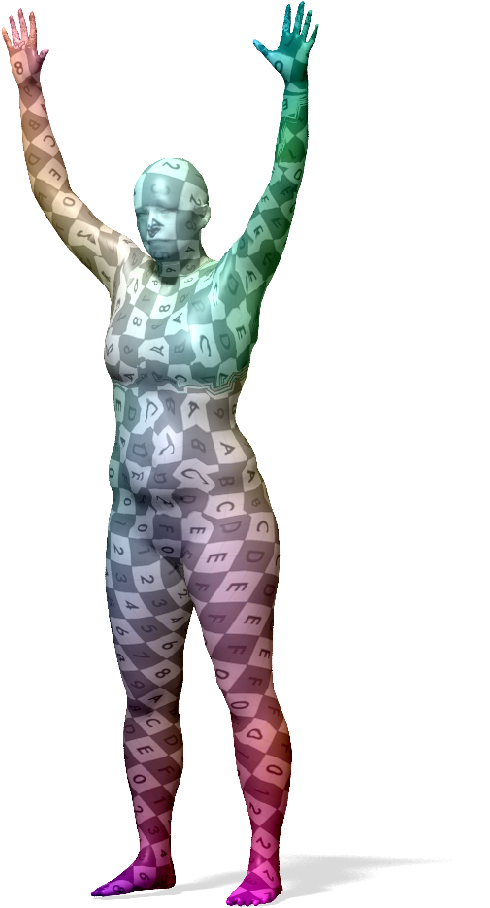}
    \caption{A failure case of our method. Left and right are switched on the upper body, causing a non-continuous correspondence. We observed eight such failure cases in the entire FAUST dataset.}
    \label{fig:faust_fail}
\end{figure}

\section{Details on multiscale acceleration}

The multiscale algorithm begins by solving for an initial sparse bijection $\pi_0:\mathcal{X}_0\to \mathcal{Y}_0$ between $n_0$ samples $s_\mathcal{X}, s_\mathcal{Y}$ (also called {\em seeds}), obtained with farthest point sampling (Euclidean in our experiments). $n_0$ can either be the maximum amount of vertices that can be handled (around 15k in our experiments) or smaller if runtime is crucial. Then $s_\mathcal{X}$ is divided into $n_0/(k \cdot maxP)$ Voronoi cells $V_{\mathcal{X},0}$ and these Voronoi cells are transferred to $\mathcal{Y}$ using $\pi_0$ to create $V_{\mathcal{Y},0}$. The parameter $maxP$ is the maximum problem size allowed in later iterations and normally much smaller than $n_0$. $k$ determines how many new samples are added in each iteration. A small $maxP$ makes the method faster but less robust, and a small $k$ slower but more robust. In our experiments, we always choose $maxP = 1500$ and $k = 3$. 
At the first iteration ($i = 1$) and any following iteration $i$, $n_i = k \times n_{i-1}$ new points are sampled in a farthest point manner on both shapes to create $\mathcal{X}_i, \mathcal{Y}_i$. Each new point is assigned to the same Voronoi cell as its nearest neighbor in $s_\mathcal{X}, s_\mathcal{Y}$ resulting in the new cells $V_{\mathcal{X},i}, V_{\mathcal{Y},i}$. If any cell has more than $maxP$ vertices, the number of cells is increased until this is not the case anymore. Next we solve for $\pi_i: \mathcal{X}_i \to \mathcal{Y}_i$ by solving for a mapping from the $m$-th cell of $V_{\mathcal{X},i}$ to the $m$-th of $V_{\mathcal{X},i}$ using the proposed method from this paper and combining them into a global permutation. Notice that the $m$-th cells of both shapes correspond to roughly the same areas as long as the previous matching $\pi_{i-1}$ that was used for its construction is reasonable. Nevertheless, the cells could include a different amount of points due to discretization errors, so we need to apply the partial matching scheme for each cell and some points may stay unmatched (in this iteration). All matched points are added to the sets $s_\mathcal{X}, s_\mathcal{Y}$ for the next iteration. Again, $\mathcal{X}$ is divided into $n_i/(k \cdot maxP)$ Voronoi cells and these are transfered to $\mathcal{Y}$ via $\pi_i$. The Voronoi cells of previous iterations are discarded to allow exchange of points between cells. This proceeds until all points have been sampled.

We use Euclidean FPS in all cases and build approximate Voronoi cells on remeshed versions of the shape to keep the runtime small. Each $\pi_i$ is solved for by using descriptors and initial matches from the previous iteration in the same cell. Additionally, we add $1000$ equally distributed matches from $\pi_0$ to every problem which aligns the solution along the boundaries of the cells with each other. Notice that even if the shapes have the same number of vertices at the beginning due to the sampling and decoupling of each cell not all vertices might be matched.

If the matched shapes are partial versions of each other, this information needs to be propagated from the first iteration on since all later cells are solved independently and can therefore not see partiality. In this case, $n_{0, \mathcal{X}}, n_{0, \mathcal{Y}}$ can be chosen dependently on the ratio of areas or number of vertices between $\mathcal{X}$ and $\mathcal{Y}$, either assuming the scale or the discretization is comparable. Then certain points of the initial sampling will stay unmatched and be marked \emph{forbidden}. They are handled exactly like any other seed but have their own Voronoi cell and any point that gets a assigned to the forbidden Voronoi cell is also marked forbidden such that the information spreads only to the neighborhood.

%----------------------------------------------------------------------------------------------------------------
\section{Run time comparison}

\begin{table*}[]
\centering
\begin{tabular}{|l|l|l|l|}
\hline
shapes in experiment & \#vertices & runtime with heat kernel in sec & runtime with Gaussian kernel in sec \\ \hline

Tosca: cat0 to cat2                       & 3400         & 29.25      & 97.77            \\
Tosca: dog0 to dog2                       & 3400         & 36         & 98.43            \\
Tosca: centaur0 to centaur1               & 3400         & 25.31      & 98.91            \\
Tosca: wolf0 to wolf1                     & 4344         & 60.7       & 192.72           \\
Faust models: 000 to 098 & 6890         & 109.477019 & 639.9            \\
Faust models: 001 to 031 & 6890         & 104.68     & 609.56           \\
Faust models: 002 to 039 & 6890         & 104.5      & 611.24           \\
Faust models: 003 to 021 & 6890         & 106.41     & 614.23           \\
Faust models: 004 to 033 & 6890         & 106.28     & 652.58           \\ \hline
\end{tabular}
\caption{Runtime comparison of matching between shapes with different number of vertices using heat kernels and Gaussian kernels.}
\label{table:run_times}
\end{table*}

The run time experiments, were conducted on a MacBook pro with a $2.5$ GHz Intel Core i7 processor and $16$ GB RAM running Matlab $2016b$. 
The experiments were conducted using $9$ pairs of shapes with a varying number of vertices from the TOSCA high and low resolution meshes as well as  FAUST registrations set.
The complete results are presented in Table \ref{table:run_times}.
We ran all our tests using SHOT descriptors, 10 iterations with $\alpha = 1/10^8$, $400$ eigenvectors to construct the heat kernels and a logarithmic scale of time parameters between $400$ and $10$.

\begin{figure}
\centering
\includegraphics[width=.46\linewidth]{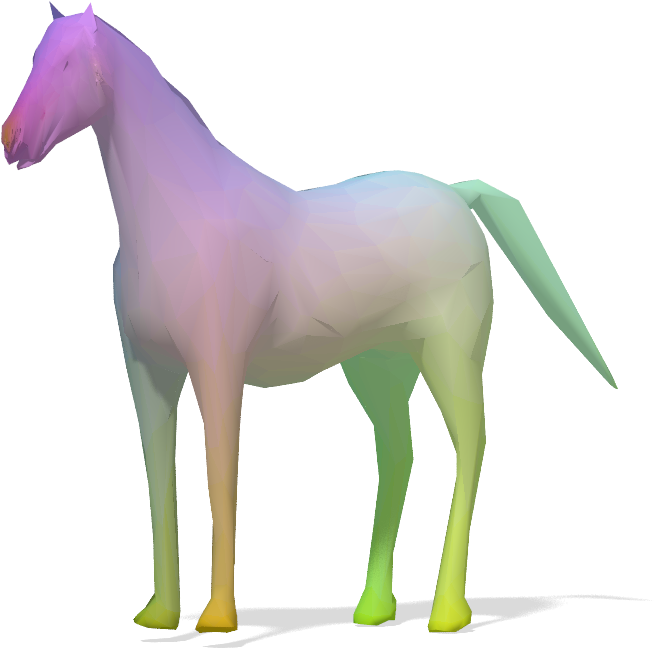}
\includegraphics[width=.5\linewidth]{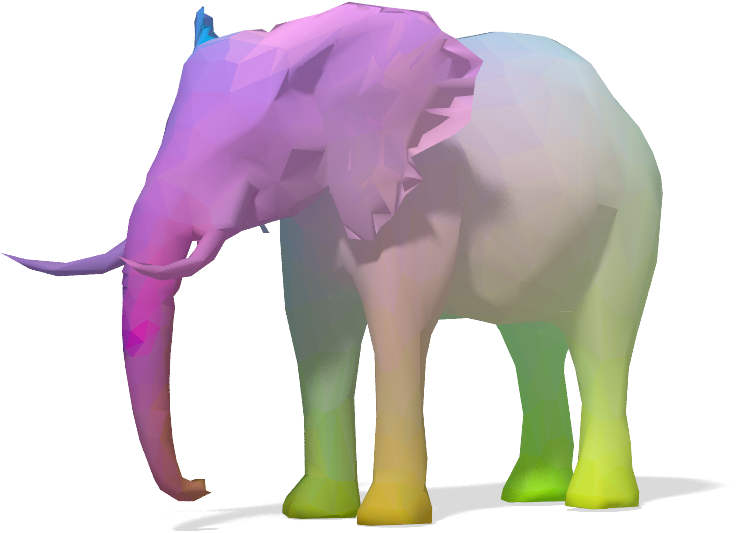}
\caption{Matching from a horse to an elephant using SHOT and HKS descriptors. The shapes are sampled in a way such that a bijective matching is possible.}
\label{fig:horse_elephant}
\end{figure}

\section{More results, Failure cases}

\begin{figure*}
\includegraphics[width=.2\linewidth]{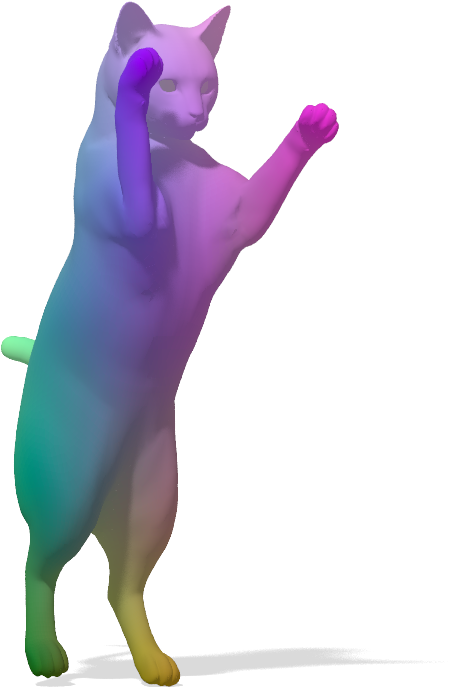}
\includegraphics[width=.2\linewidth]{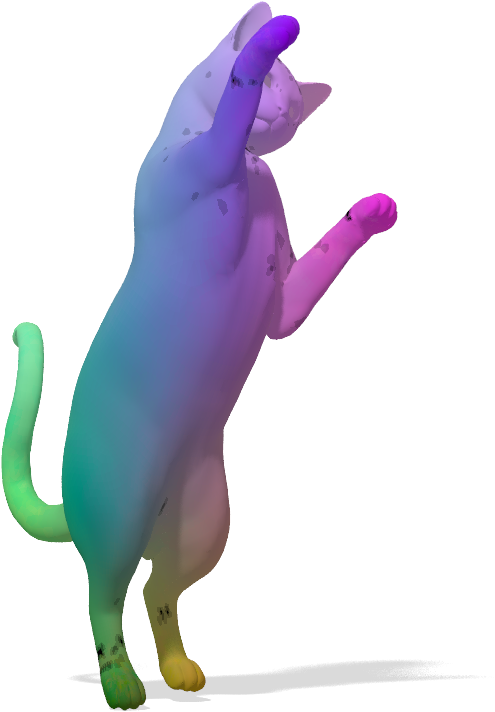}
\includegraphics[width=.3\linewidth]{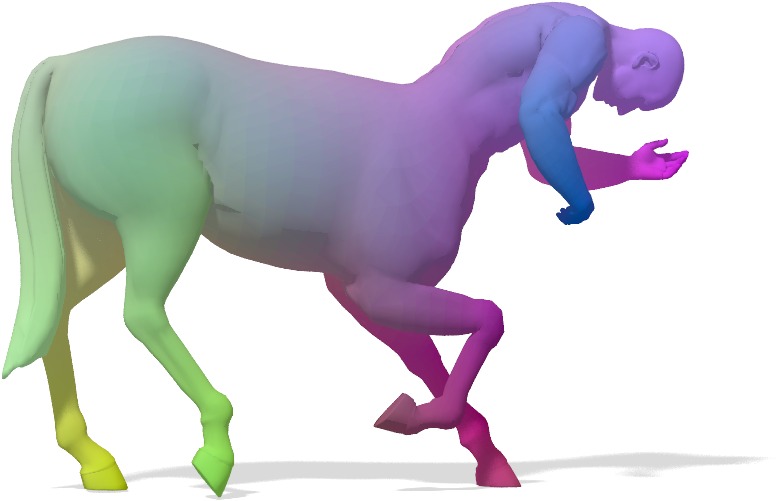}
\includegraphics[width=.3\linewidth]{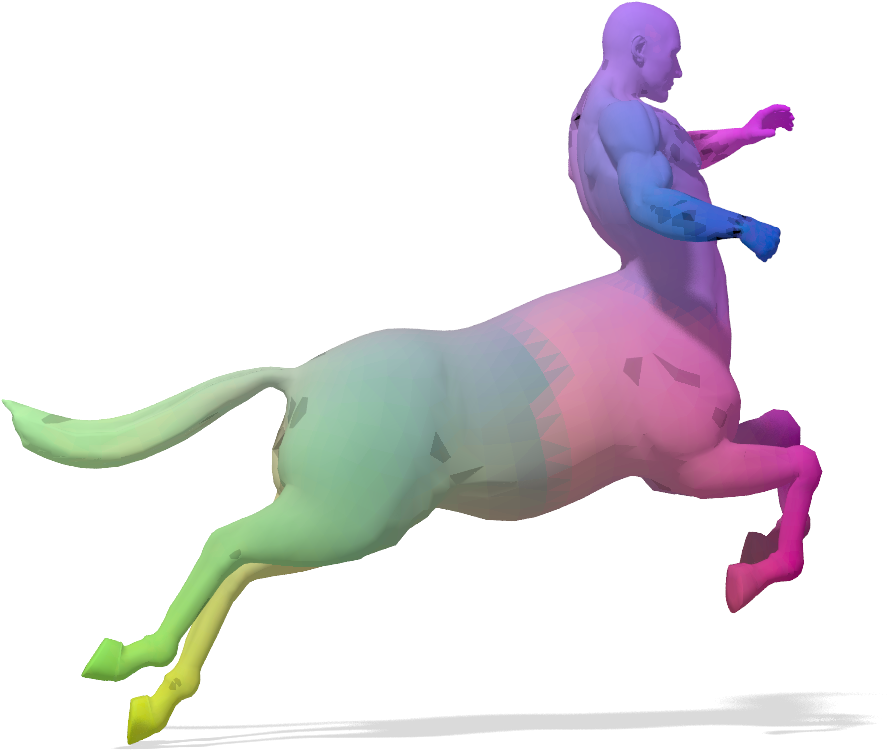}
\caption{(left) A matching between two cats from the Tosca dataset. The unmatched points resulting from the multiscale (black) are very sparse. (right) A failed matching on the centaurs from Tosca. The front legs are swapped but only few points are unmatched.}
\label{fig:tosca}
\end{figure*}

\begin{figure*}[]
\includegraphics[width=.19\linewidth]{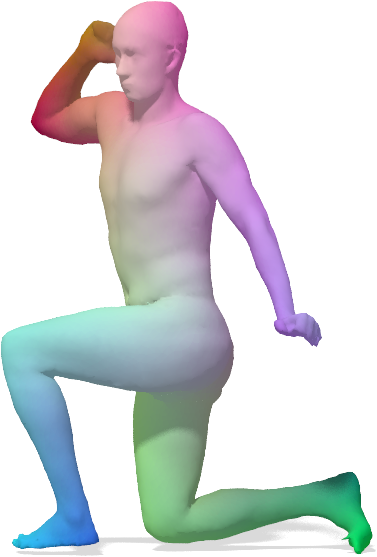}
\includegraphics[width=.22\linewidth]{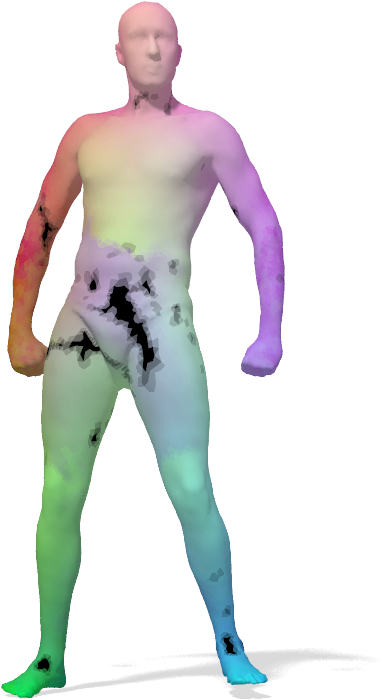}
\includegraphics[width=.35\linewidth]{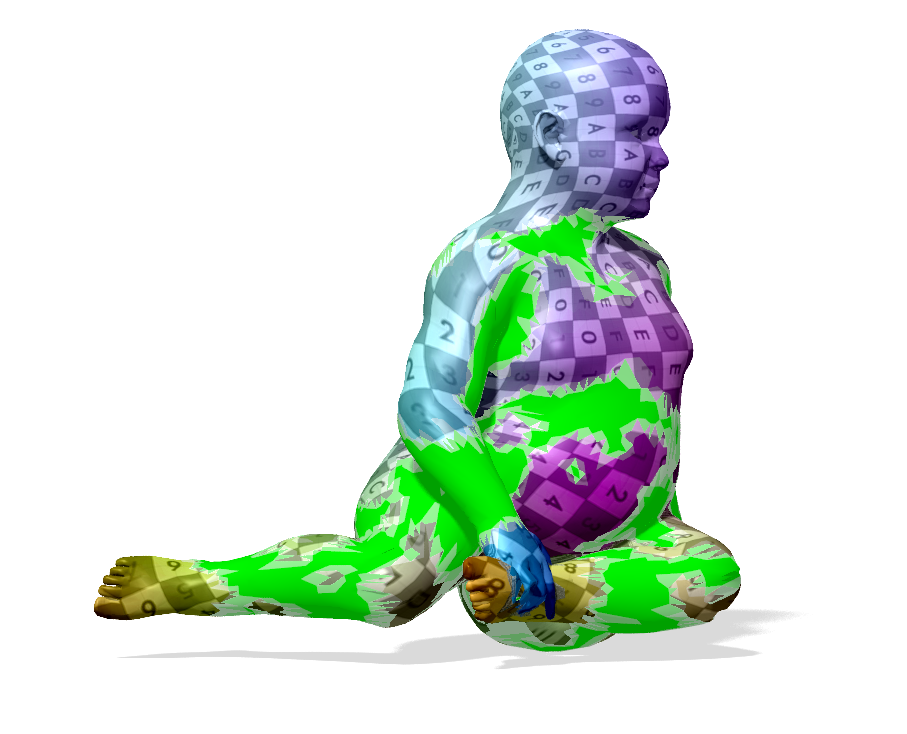}
\includegraphics[width=.22\linewidth]{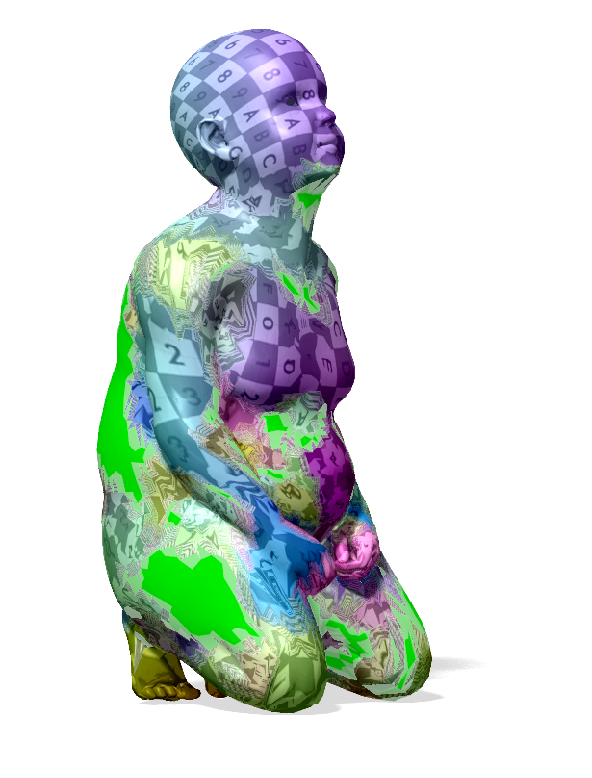}
\caption{(left) Failure case on the SCAPE dataset.The legs are mapped front to back causing a non continuous correspondence on the torso. Large unmatched areas due to the multiscale also appear there. Over the knees unaligned cell boundaries are visible. (right) Failure case on the SHREC'16 dataset. Many parts are missing or the texture is heavily distorted. These are really challenging shapes to match because the hands and feet are topologically merged to different parts of the body in both cases. }
\label{fig:failures}
\end{figure*}

In this section we show additional results for a pair of dramatically non-isometric shapes (Fig.\ref{fig:horse_elephant}), pairs from the Tosca dataset (Fig.\ref{fig:tosca}) and failure cases (Figs.\ref{fig:faust_fail},\ref{fig:failures}).

